\documentclass[table,dvipsnames]{arxiv}

\usepackage{tikz}
\usepackage{xr-hyper}
\usepackage{hyperref}
\usepackage{cleveref}

\definecolor{darkblue}{RGB}{46,25, 110}
\newcommand{\dssectionheader}[1]{%
   \noindent\framebox[\columnwidth]{%
      {\fontfamily{phv}\selectfont \textbf{\textcolor{darkblue}{#1}}}
   }
}
\newcommand{\dsquestion}[1]{%
    {\noindent \fontfamily{phv}\selectfont \textcolor{darkblue}{\textbf{#1}}}
}
\newcommand{\dsquestionex}[2]{%
    {\noindent \fontfamily{phv}\selectfont \textcolor{darkblue}{\textbf{#1} #2}}
}
\newcommand{\dsanswer}[1]{%
   {\noindent #1 \medskip}
}
\NewDocumentCommand{\codeword}{v}{%
\frenchspacing\texttt{\textcolor{gray}{#1}}%
}
\hypersetup{
    pdftitle={SuperAnimal pretrained pose estimation models for behavioral analysis},
    pdfauthor={Ye et al.},
    pdfsubject={SuperAnimal models pretrained for behavioral analysis},
    colorlinks=true, allcolors=violet
}

\usepackage{graphicx}
\usepackage{epstopdf}
\usepackage{booktabs}
\usepackage{siunitx}
\usepackage{enumitem}
\usepackage{verbatim}
\usepackage{dblfloatfix}
\usepackage{algpseudocode}
\usepackage{bbding}
\usepackage{balance}

\usepackage{amsmath}
\usepackage{amsfonts}
\usepackage{amssymb}
\usepackage{mathtools}
\usepackage[sectionbib]{bibunits}
\defaultbibliographystyle{unsrt} 
\defaultbibliography{refs}

\usepackage{changepage}
\usepackage[breakable]{tcolorbox}

\newenvironment{mcsection}[1]
    {%
        \textbf{#1}

        \begin{itemize}[leftmargin=*,topsep=0pt,itemsep=-1ex,partopsep=1ex,parsep=1ex,after=\vspace{\medskipamount}]
    }
    {%
        \end{itemize}
    }

\newcommand\blfootnote[1]{%
  \begingroup
  \renewcommand\thefootnote{}\footnote{#1}%
  \addtocounter{footnote}{-1}%
  \endgroup
}

\newcommand{\beginsupplement}{%
        \setcounter{table}{0}
        \renewcommand{\thetable}{S\arabic{table}}%
        \setcounter{figure}{0}
        \renewcommand{\thefigure}{S\arabic{figure}}%
     }

\usepackage{nimbusmononarrow}
\usepackage{orcidlink}
\usepackage{fontawesome}
\usepackage{multicol}

\usepackage{pdflscape}
\usepackage{longtable}

\usepackage{listings}
\usepackage{xcolor}

\definecolor{codegreen}{rgb}{0,0.6,0}
\definecolor{codegray}{rgb}{0.5,0.5,0.5}
\definecolor{codepurple}{rgb}{0.58,0,0.82}
\definecolor{backcolour}{rgb}{0.95,0.95,0.92}

\lstdefinestyle{mystyle}{
    backgroundcolor=\color{backcolour},   
    commentstyle=\color{codegreen},
    keywordstyle=\color{magenta},
    numberstyle=\tiny\color{codegray},
    stringstyle=\color{codepurple},
    basicstyle=\ttfamily\footnotesize,
    breakatwhitespace=false,         
    breaklines=true,                 
    captionpos=b,                    
    keepspaces=true,                 
    numbers=left,                    
    numbersep=5pt,                  
    showspaces=false,                
    showstringspaces=false,
    showtabs=false,                  
    tabsize=2
}

\title{SuperAnimal pretrained pose estimation models for behavioral analysis}

\shorttitle{SuperAnimal}

\author[1\orcidlink{0000-0003-4250-2220}]{Shaokai Ye}
\author[1\orcidlink{0009-0000-9488-4174}]{Anastasiia Filippova}
\author[1\orcidlink{0000-0002-3656-2449}]{Jessy Lauer}
\author[1\orcidlink{0000-0003-2327-6459}]{Steffen Schneider}
\author[1]{Maxime Vidal}
\author[1]{Tian Qiu}
\author[1\orcidlink{0000-0002-3777-2202}]{Alexander Mathis}
\author[1 \orcidlink{0000-0001-7368-4456}]{Mackenzie Weygandt Mathis* \Envelope}

\affil[1]{École Polytechnique Fédérale de Lausanne (EPFL), Brain Mind Institute \& Neuro-X Institute. Geneva, Switzerland} 
\leadauthor{Ye}

\begin{document}
\onecolumn
\maketitle

\begin{multicols}{2}
[
\textbf{Quantification of behavior is critical in applications ranging from neuroscience, veterinary medicine and animal conservation efforts. A common key step for behavioral analysis is first extracting relevant keypoints on animals, known as pose estimation. However, reliable inference of poses currently requires domain knowledge and manual labeling effort to build supervised models. We present a series of technical innovations that enable a new method, collectively called SuperAnimal, to develop unified foundation models that can be used on over 45 species, without additional human labels. Concretely, we introduce a method to unify the keypoint space across differently labeled datasets (via our generalized data converter) and for training these diverse datasets in a manner such that they don't catastrophically forget keypoints given the unbalanced inputs (via our keypoint gradient masking and memory replay approaches). These models show excellent performance across six pose benchmarks. Then, to ensure maximal usability for end-users, we demonstrate how to fine-tune the models on differently labeled data and provide tooling for unsupervised video adaptation to boost performance and decrease jitter across frames. If the models are fine-tuned, we show SuperAnimal models are 10-100$\times$ more data efficient than prior transfer-learning-based approaches. We illustrate the utility of our models in behavioral classification in mice and gait analysis in horses. Collectively, this presents a data-efficient solution for animal pose estimation.
\blfootnote{\Envelope  mackenzie.mathis@epfl.ch}
\vspace{6pt}
}
]

\section*{Introduction}

Measuring and modeling behavior is an important step in many clinical, biotechnological, and scientific quests~\cite{datta2019computational, mathis2020deep, pereira2020quantifying, von2021big,hausmann2021measuring,Tuia2021PerspectivesIM}. A key part of many behavioral analysis pipelines is animal pose estimation, yet this requires domain knowledge and labeling efforts to obtain reliable pose models~\cite{Mathis2020APO,Anderson2014TowardAS,pereira2020quantifying,mathis2020deep}. Open-source pose estimation software, such as DeepLabCut~\cite{mathis2018deeplabcut,Lauer2022multi} and other tools~\cite{graving2019deepposekit,gunel2019deepfly3d,pereira2022sleap,Bala2020AutomatedMP} also reviewed in~\cite{Mathis2020APO}, have gained popularity in the research community interested in understanding animal behavior. Compared to commercial solutions constrained to fixed cage and camera settings~\cite{sturman2020deep}, DeepLabCut offers flexibility to train customized pose models of various animals in diverse settings. Notably, it requires few human-labeled images (around 100–800) to train a typical lab animal pose estimator that matches human level accuracy~\cite{mathis2018deeplabcut,Lauer2022multi} due to its transfer learning abilities~\cite{mathis2018deeplabcut, mathis2021pretraining}.
\medskip 

However, regardless of the data efficiency of current solutions, their flexibility still comes with the cost of requiring users to label if they want to define keypoints themselves (note, some unsupervised approaches are available, but lack the ability of users to customize to keypoints of scientific interest~\cite{sun2023bkind,sosa2023self}). Then, they train deep neural networks, an effort that is often duplicated across labs given that often users study similar model organisms. 
\medskip 

A solution is to build generalized, foundation models~\cite{Bommasani2021OnTO}, for common model organisms across labs and in-the-wild settings (proposed and discussed in~\cite{Mathis2020APO}). Such models could be used across labs and settings without further training and/or requiring little fine-tuning. Yet, there are several key challenges to build these models. Firstly, data on the same species is rarely labeled the same way or even with the same names (for example consider simply naming the nose on a mouse: ``nose'', ``snout'', ``mouse1\_nose'', etc -- all found in the literature~\cite{mathis2018deeplabcut,sturman2020deep}), which brings semantic and annotation bias challenges: how do we merge such data? Secondly, even if we unify the naming, how do we train across datasets that don't have the full super-set of keypoints? Missing data would confuse a network without any interventions.
\medskip 

\begin{figure*}
\centering
\includegraphics[width=.83\textwidth]{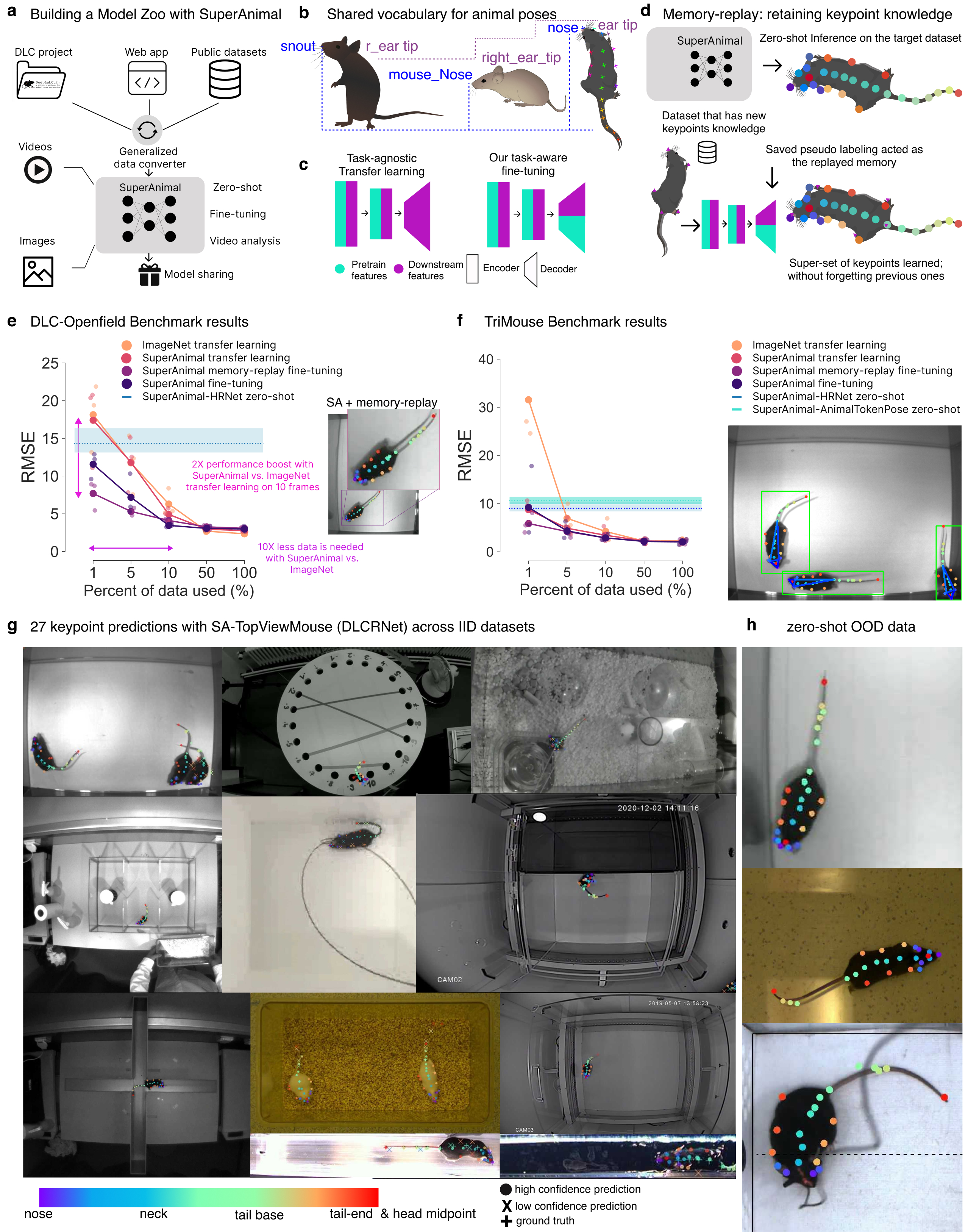}
\caption{\textbf{The DeepLabCut Model Zoo, the SuperAnimal method, and SuperAnimal-TopViewMouse model performance.}
\textbf{a}: The website can collect data shared by the research community; SuperAnimal models are trained, and can be used for inference on novel images and videos with or without further training (fine-tuning).
\textbf{b}: The panoptic animal pose estimation approach unifies the vocabulary of pose data across labs, such that each individual dataset is a subset of a super-set keypoint space, independently of its naming. 
\textbf{c}: For canonical task-agnostic transfer learning, the encoder learns universal visual features from ImageNet, and a randomly initialized decoder is used to learn the pose from the downstream dataset. For task-aware fine-tuning, both encoder and decoder learn task-related visual-pose features in the pre-training datasets and the decoder is fine-tuned to update pose priors in downstream datasets. Crucially, the network has pose-estimation specific weights. 
\textbf{d}: Memory replay combines the strengths of SuperAnimal models' zero-shot inference, data combination strategy, and leveraging labeled data for fine-tuning (if needed).
\textbf{e}: Data efficiency of baseline (ImageNet) and various SuperAnimal fine-tuning methods using bottom-up DLCRNet on the DLC-Openfield OOD dataset. 1-100\% of the train data is 10, 50, 101, 506, 1012 frames respectively. Blue shadow represents minimum, maximum and blue dash is the mean for zero-shot performance across three shuffles. Large, connected dots represent mean results across three shuffles and smaller dots represent results for individual shuffles. Inset: Using memory replay avoids catastrophic forgetting.
\textbf{f}: SuperAnimal vs. baseline results on the TriMouse benchmark, showing zero-shot performance with top-down HRNet and AnimalTokenPose, and fine-tuning results with HRNet.  1-100\% of the train data is 1, 7, 15, 76, 152 frames respectively Inset: example image of results.
\textbf{g}: SuperAnimal-TopViewMouse (DLCRNet) qualitative results on the within distribution test images (IID). They were randomly selected based on the visibility of the keypoints within the figure (but not on performance). Full keypoint color and mapping is available in Extended Data Figure~\ref{fig:Extended_data_fig1}).
\textbf{h}: Visualization of model performance on OOD images using DLCRNet.}
\label{fig:Fig1overviewASTVM}
\end{figure*}

To provide the research community with an easy method to build such high performance models we present a new panoptic paradigm -- which we call the SuperAnimal method -- for building unified pre-trained pose-aware models, and the ability to perform fine-tuning and video adaptation across many species, environments and animal or video sizes (Figure~\ref{fig:Fig1overviewASTVM}a).
\medskip

In brief, our new approach allows for merging and training diverse, differently labeled datasets. We developed an optimal keypoint matching algorithm to automatically align out-of-distribution datasets with our models. Then, at inference time, to minimize domain shifts, we developed a spatial-pyramid search method to account for changes in animal size or leverage a top-down detector. We also provide a rapid, unsupervised video-adaptation method that uses pseudo-labeling to boost performance and minimize temporal jitter in videos and allows users to fine-tune videos without access to source data or requiring any target labeling on that video.
\medskip

We developed models based on state-of-the-art (SOTA) convolutional neural networks (CNNs) such as HRNet~\cite{wang2020hrnet} or DLCRNet~\cite{Lauer2022multi}, and transformers~\cite{Dosovitskiy2021ViT,Yang_2021_ICCV,xu2022vitpose}. We show that the resulting models have excellent zero-shot performance (i.e., with no additional training, tested on new data), and our approach outperforms ImageNet-pretraining on six benchmarks. If used for fine-tuning, we show they are 10 to nearly 100$\times$ more data efficient in the low data regime (and can still improve performance in the high-data setting), and our video adaptation method allows for smooth, refined videos that can be used in behavioral analysis pipelines.

\vspace{-5pt}
\section*{Results}

The SuperAnimal method comprises generalized data conversion, training with keypoint gradient masking and memory replay, a keypoint matching algorithm, and the ability to fine-tune models plus video adaptation that pseudo-labels using unlabeled video data (Figure~\ref{fig:Fig1overviewASTVM}a), which will be explained below. Firstly, we describe the data we used to train our two exemplary models with our SuperAnimal approach.

\subsection*{Animal Pose Data}

In order to demonstrate the strength of our SuperAnimal method, we present two datasets that cover over 45 species: TopViewMouse-5K and Quadruped-80K, which are built from over 85,000 images sourced from diverse laboratory settings and in-the-wild data (Extended Data Figure~\ref{fig:Extended_data_fig1}a, b), yet critically they are not labeled in the same manner.
First, we used a new generalized data converter (see Methods) to unify the annotation space of those datasets and named the first dataset TopViewMouse-5K (as it contains approximately 5k images). Specifically, we merged 13 overhead-camera view-point lab mice datasets from across the research community~\cite{mathis2018deeplabcut,Lauer2022multi,sturman2020deep,isaac_chang_2020_3955216,Nilsson2020.04.19.049452} (see Methods) and from our own experiments (Figure~\ref{fig:Fig1overviewASTVM}e, h). Similarly, we collected side-view quadruped datasets~\cite{Joska2021AcinoSetA3,mathis2021pretraining,Khosla2012NovelDF,Cao2019CrossDomainAF,banik2021novel,biggs2020left,yu2021ap, yang2022apt}, including a new annotated ``iRodent'' dataset with images sourced from iNaturalist (see Methods), to form Quadruped-80K (Extended Data Figure~\ref{fig:Extended_data_fig1}b, and see Supplemental Datasheets).
We define six benchmark datasets of varying difficulty and always leave-one-out of the training in order to show performance of the model on unseen data. There are: DLC-Openfield~\cite{mathis2018deeplabcut}, TriMouse~\cite{Lauer2022multi}, iRodent (new), Horse-10~\cite{mathis2021pretraining}, AP-10K~\cite{yu2021ap} and AnimalPose~\cite{cao2019cross}. 
Note, that our released SuperAnimal-TopViewMouse and SuperAnimal-Quadruped weights (see Supplementary Information, Model Cards) are trained on all available data described above (Extended Data Figure~\ref{fig:Extended_data_fig1}b).

\subsection*{The SuperAnimal method}

Collectively, SuperAnimal is a formulation that treats diverse pose datasets as if they collectively formed one single super-set pose template, trains unified models for image and video analysis, and ultimately allows sharing of these models through standardized repositories (Figure~\ref{fig:Fig1overviewASTVM}a). This panoptic super-set approach effectively allowed us to overcome a major challenge with combining datasets that are not identically labeled across labs or datasets, as it is often the case even for the same species (Figure~\ref{fig:Fig1overviewASTVM}b and Extended Data Figure~\ref{fig:Extended_data_fig1}a). 
Multi-dataset training allows the model
%\footnote{Note, in this work we adopt the definition of Foundation Model from G. Hinton: \url{https://www.youtube.com/watch?v=E14IsFbAbpI&t=2700s}}
to receive richer learning signals (Figure~\ref{fig:Fig1overviewASTVM}c, d), resulting in the model having ``pose priors'' (whereas ImageNet pre-training, is common in animal pose~\cite{mathis2021pretraining,Lauer2022multi,cao2019cross} has no pose-specific features). 
For multi-dataset training we developed keypoint gradient masking (Extended Data Figure~\ref{fig:Extended_data_fig1}c, d) to train neural networks across disjoint datasets without penalizing ``missing'' ground truth data from the super-set of keypoints (Figure~\ref{fig:Fig1overviewASTVM}b). We also developed a keypoint matching algorithm (Methods and Extended Data Figure~\ref{fig:Extended_data_matching_ATP}a, b) to help minimize the mismatch caused by annotator bias in the human annotated datasets (see Suppl. Note).
\medskip

The neural networks we trained always consist of an encoder and a decoder. While transfer learning has been important for animal pose~\cite{mathis2018deeplabcut,mathis2021pretraining}, we hypothesized now that we have base encoders that have pose priors, that the trained decoder could be leveraged (Figure~\ref{fig:Fig1overviewASTVM}c, d).
Therefore, we tested two ways to train the architectures: one, via transfer learning, defined as fine-tuning only the pre-trained encoder but using a randomly initialized decoder in the downstream dataset; two, via fine-tuning, defined as fine-tuning both the encoder and decoder (see Methods). We also note that we used both bottom-up and top-down methods~\cite{Mathis2020APO}, meaning without or with an object detector, respectfully, as noted in figure captions. 
\medskip

\subsection*{Benchmarking}

We aim to show the performance of our approach in three important settings: (1) zero-shot inference: how performant is the base model on unseen data? (2) fine-tuning on a new dataset: how does the base model compare to using a base model trained on ImageNet (i.e., ImageNet transfer learning)? (3) If zero-shot and/or fine-tuning is efficient, how good is the base model performance on videos and for downstream tasks like behavior quantification?
\medskip

We report results on two models classes: SuperAnimal-TopViewMouse (SA-TVM) then SuperAnimal-Quadruped (SA-Q). For each class we consider several architectures for zero-shot and fine-tuning (Figures~\ref{fig:Fig1overviewASTVM}, \ref{fig:FigSAQ}, Extended Data Figure~\ref{fig:Extended_data_matching_ATP}c, d), and then consider performance in video and behavioral analysis (Figures~\ref{fig:FigVideoAdapt}, ~\ref{fig:action_segmentation}).
To evaluate each models performance we tested ``within distribution'' also known as ``independent and identically distributed" (IID), and on images considered ``out-of-distribution'' (OOD).
IID images are similar in appearance and from the same dataset, but not identical to those used in training. OOD data stems from images in datasets that were never included in training~\cite{koh2021wilds}, and they constitute the key benchmark results for showing the utility of the models in applied settings (Tables~\ref{tab:overall_mouse} and ~\ref{tab:quad}).

\subsection*{Zero-shot performance (SA-TVM)}

Using the panoptic SuperAnimal approach we first consider the performance in the OOD zero-shot setting and find that is has excellent performance across both top-view mouse benchmarks (Table~\ref{tab:overall_mouse}). We tested both a bottom-up DLCRNet (Figure~\ref{fig:Fig1overviewASTVM}e-h) and a top-down HRNet-w32~\cite{martinez2017simple,Lauer2022multi,pereira2022sleap} (Extended Data Figure~\ref{fig:Extended_data_TDmouse}), which was recently shown to be excellent in crowded animal scenes~\cite{zhou2023iccv}, and our transformer for testing. 
Specifically, to test performance, we built SA-TVM models that did not contain data from the DLC-Openfield~\cite{mathis2018deeplabcut} or TriMouse dataset~\cite{mathis2018deeplabcut,Lauer2022multi}.
\medskip

We found that the SuperAnimal methods were critical:
using gradient masking SA-TVM DLCRNet zero-shot performance was $14.31 \pm 1.00$ RMSE vs. $27.90 \pm 1.20$ without gradient masking (Extended Data Figure~\ref{fig:Extended_data_SATVMzeroshot}a). Memory replay was critical to avoid catastrophic forgetting qualitatively (see Suppl. Video 1) and quantitatively, measured with keypoint dropping (See Extended Data Figure~\ref{fig:Extended_data_SATVMzeroshot}c and Tables~\ref{memreplay_lme} and~\ref{contrasts_memreplay}).
\medskip

Collectively, they show excellent zero-shot performance on both benchmarks (Figure~\ref{fig:Fig1overviewASTVM}e, f, Extended Data Figure~\ref{fig:Extended_data_SATVMzeroshot}c, and Tables~\ref{TD_openfield_results} and ~\ref{TD_trimouse_result}). 
SA-TVM performed well within distribution (IID) and critically OOD data across diverse camera and cage settings (Figure~\ref{fig:Fig1overviewASTVM}g, h). Note that the performance of zero-shot inference is even likely under-estimated by annotator bias (see Suppl. Note).
Concretely, zero-shot SA-TVM DLCRNet bottom-up showed a RMSE error of $14.31$ pixels, $4.88$ pixels with the HRNet-w32-based top-down approach, and $4.57$ pixels with AnimalTokenPose on the DLC-Openfield dataset, where the average mouse's nose width is approximately 10 pixels~\cite{mathis2018deeplabcut} (Figure~\ref{fig:Fig1overviewASTVM}e, f). Thus, we found that without any labeling we could still outperform ImageNet-based transfer learning (Figure~\ref{fig:Fig1overviewASTVM}e; mixed-effect model; in the low-data regime (1\% training data ratio) for TriMouse: $d$=8.03 [5.27, 10.79] $p$<.0001; Tables~\ref{topview_trimouse_lme} and~\ref{contrasts_topview_trimouse}; for DLC-Openfield: $d$=3.86 [1.88, 5.84] $p$=.0002, Tables~\ref{BU_openfield_lme} and ~\ref{BU_contrasts_openfield}).

\subsection*{Fine-tuning performance (SA-TVM)}

For fine-tuning SuperAnimal models we consider two ways: one, naive fine-tuning (see Methods), and inspired by the excellent zero-shot inference of pre-trained models~\cite{radford2021learning} and continual learning \cite{van2020brain}, we developed a tailored fine-tuning approach that combines zero-shot inference and few-shot learning, which we call ``memory replay'' fine-tuning (Figure~\ref{fig:Fig1overviewASTVM}d).  We find that in the user-relevant low-data regime, fine-tuning significantly outperforms ImageNet transfer learning (Figure~\ref{fig:Fig1overviewASTVM}e, f; mixed effects model, DLC-Openfield: $d$=7.19 [4.61, 9.78]; $p$<.0001, TriMouse: $d$=8.06 [5.29, 10.82]; $p$<.0001; Tables~\ref{topview_trimouse_lme},~\ref{contrasts_topview_trimouse}~\ref{TD_topview_openfield_lme}, and~\ref{contrasts_topview_openfield}). This is approximately a 10$\times$ data efficiency factor and large margin of performance gain (Figure~\ref{fig:Fig1overviewASTVM}e). Note that effect sizes remain moderate to large even when training with 5\% of the data ($d$>.59).
\medskip

For example, if the model is memory replay fine-tuned with only 10 randomly selected images on DLC-Openfield, the SA-TVM pre-trained model obtained an RMSE of $7.68$ pixels, whereas ImageNet pre-training was $18.14$ pixels. The baseline ImageNet pre-trained model required 101 (randomly selected) images to reach a performance similar to SA-TVM ($6.28$ pixels; Figure~\ref{fig:Fig1overviewASTVM}e). Therefore, we outperformed DeepLabCut-DLCRNet (i.e., the ImageNet baseline) by over 2X in the low data regime (i.e., with 10 frames of labeling; $p$<.0001, Cohen's $d$=4.88), and we can achieve the same performance as DeepLabCut-ImageNet weights with 10$\times$ less data (i.e, using 10 frames with our SA-TVM memory replay gives the same RMSE as ImageNet transfer learning with 101 images).
\medskip

One important point is that the SA-TVM model is now imbued with a ``pose prior''. Historically, the transfer learning using ImageNet weights strategies assumed no ``animal pose task priors'' in the pretrained model, a paradigm adopted from previous task-agnostic transfer learning~\cite{Donahue2013DeCAFAD}. Yet, here we show that naively fine-tuning on datasets that do not have the full super-set of points might cause catastrophic forgetting (Suppl. Video 2 and Extended Data Figure \ref{fig:Extended_data_SATVMzeroshot}c). Namely, if we fine-tuned with the four keypoint dataset from DLC-Openfield, the model would forget the full 27 keypoints.

\begin{figure*}[t]
\begin{center}
\includegraphics[width=.81\textwidth]{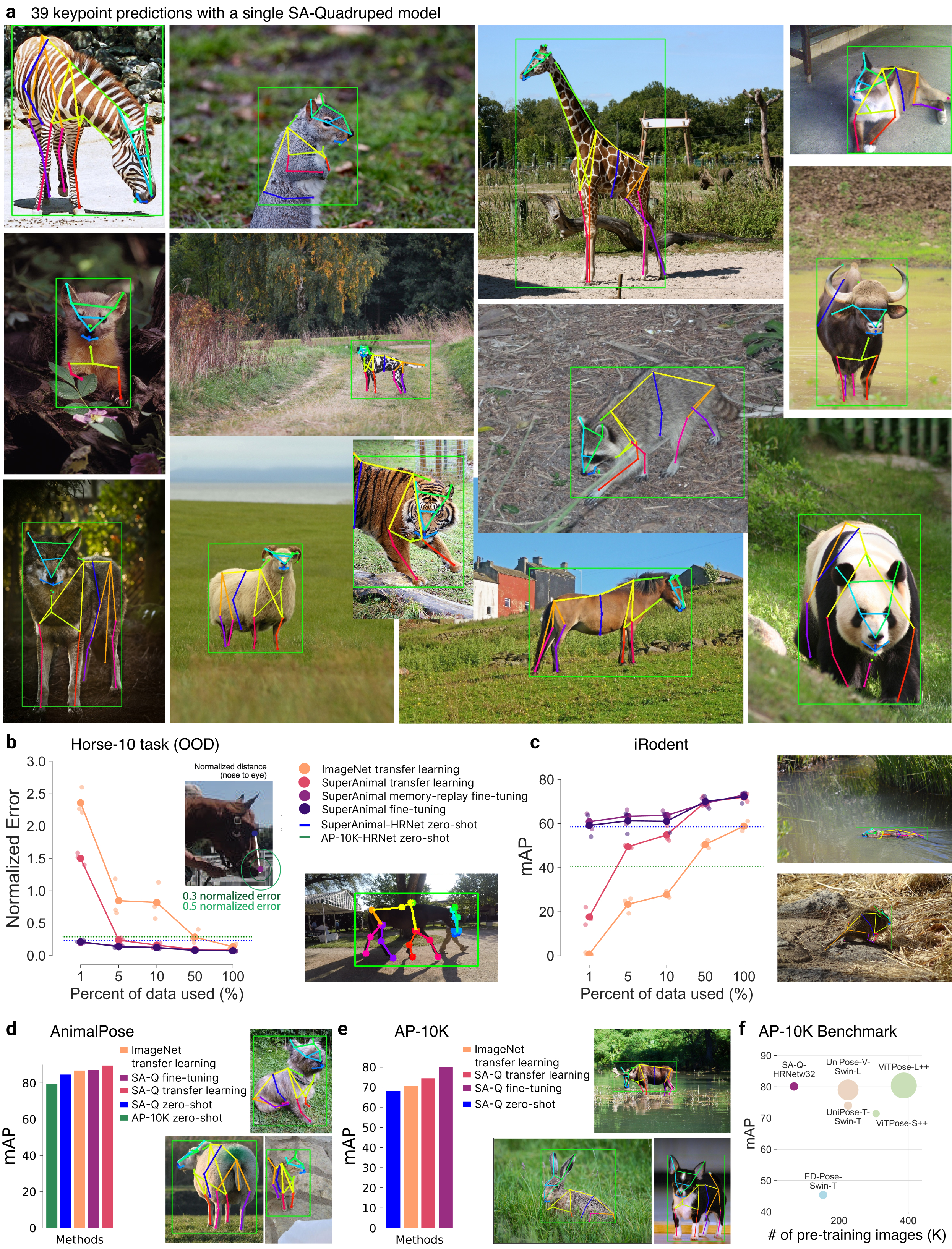}
\caption{\textbf{SuperAnimal-Quadruped}
\textbf{a}: Qualitative performance with SuperAnimal-Quadruped (HRNet-w32). Image randomly selected based on visibility of the keypoints within the figure (but not on performance). A likelihood cutoff of 0.6 was applied for keypoint visualization. Full keypoint color and mapping is available in Extended Data Figure~\ref{fig:Extended_data_fig1}). HRNet-w32 and same cutoff of 0.6 are used in other panels.
\textbf{b}: Performance on the official OOD Horse-10 test set, training with the official IID splits, reported as a normalized error from eye to nose, see inset adopted from~\cite{mathis2021pretraining} and qualitative zero-shot performance. HRNet-w32 is trained on AP-10K and Quadruped-80K, respectively, for zero-shot performance comparison. 1-100\% of the data is 14, 73, 146, 734, 1469 frames, respectively.
\textbf{c}: Performance on the OOD iRodent dataset, reported mAP. Colors and zero-shot baseline are as in \textbf{b}.  1-100\% of the data is 3, 17, 35, 177, 354 frames, respectively. See inset for qualitative zero-shot performance.
\textbf{d}: Performance on the OOD AnimalPose dataset, reported as mAP. HRNet-w32 trained on AP-10K is used as an additional zero-shot baseline. Qualitative zero-shot performance is shown.
\textbf{e}: Performance on the OOD AP-10K dataset, reported as mAP.
Qualitative zero-shot performance is also shown.
\textbf{f}: AP-10K benchmark with SA-Q and other pose data pre-trained models. The size of dots represents the parameter size of each model. The number of pre-training images represents the number of pose data models trained before being fine-tuned on AP-10K.
}
\label{fig:FigSAQ}
\end{center}
\end{figure*}

\subsection*{Zero-shot performance (SA-Q)}

Developing pre-trained animal pose models to work in the wild is a challenging task. There are two main reasons for its difficulty: (1) the lack of labeled data, and (2) the diversity of the data. Firstly, compared to the widely used COCO human keypoint benchmark ~\cite{cocodataset} that has 200K images, the biggest wild animal pose keypoint benchmarks have 10-36K images from AP-10K~\cite{yu2021ap} and APT-36K~\cite{yang2022apt}, respectively. Yet even with Quadruped-80K we generate, the number of images is still much less than that in COCO.
Secondly, the appearance size of the animals is a long tail distribution (discussed below), which can pose a challenge for models to learn.
\medskip

To tackle such challenges, we developed top-down HRNet-w32 based SA-Q models (Figure~\ref{fig:FigSAQ}a), test zero-shot our transformer (Extended Data Figure \ref{fig:Extended_data_SATVMzeroshot}d). 
We tested SA-Q performance on four OOD benchmarks that had various official metrics: Horse-10~\cite{mathis2021pretraining} reports the normalized error (NE, normalized by the animals size, see inset in Figure~\ref{fig:FigSAQ}b),  iRodent, AP-10K~\cite{yu2021ap} and AnimalPose~\cite{Cao2019CrossDomainAF} report the mAP. As a reminder, for every benchmark we took a leave-one-out strategy such that the benchmark data was not used for training (Extended Data Figure~\ref{fig:Extended_data_QuadCreate}a). Following the common practice in top-down animal pose works~\cite{yu2021ap, Xu2022ViTPoseVT}, we report results using ground-truth bounding boxes and flip test in the test time (see Methods).
\medskip

In addition to ImageNet pretrained weights as a baseline, in benchmarks other than on AP-10K we also used a HRNet-w32 pre-trained on AP-10K (green dash line or column, Figure~\ref{fig:FigSAQ}b-d) as an additional baseline comparison. 
For comparisons to the official benchmarks we report the official metrics in Figure~\ref{fig:FigSAQ}.
We also report mAP and RMSE for all benchmarks, which can be found in Tables~\ref{horse10_results},~\ref{rodent_results},~\ref{animalpose_results}, and ~\ref{ap10k_results}.
\medskip

Horse-10 is a benchmark challenge that tests OOD robustness of generalizing to held out individual horses. We evaluated on the official splits and show zero-shot SA-Q gives $0.228$ NE (Figure~\ref{fig:FigSAQ}b), which outperforms an AP-10K-trained model that achieves a $0.287$ NE. Note the current SOTA performance in Horse-10 benchmark is around $0.3$ NE with a bottom-up EfficientNet~\cite{mathis2021pretraining}. We thus used top-down HRNet-w32 as a stronger baseline that gives $0.135$ OOD NE. Importantly, we find that SA-Q has zero-shot that is as good as supervised training with 50\% training data (734 images) using the HRNet ImageNet baseline (Figure~\ref{fig:FigSAQ}b), and with SA-Q + fine-tuning we can achieve $0.146$ normalized error with 5\% (73 frames) of data, and with the full data a $0.07$ OOD NE, setting a new SOTA. 
\medskip

iRodent is a challenging new dataset comprising a diverse set of images of rodents that are often under heavy occlusion, have a complex background, and have very various appearance sizes (Figure~\ref{fig:FigSAQ}c)
, yet with SA-Q we can achieve excellent zero-shot performance $58.6$ mAP (Figure~\ref{fig:FigSAQ}c), which is on par with $58.9$ mAP obtained by fully trained (100\% training data) HRNet-w32 using ImageNet weights. In contrast, AP-10K's weights gives $40.4$ mAP, $18.2$ points lower than ours.
\medskip

On the AnimalPose, which is a benchmark dataset consisting of dogs, cats, cows, horses, and sheep with 20 keypoints~\cite{Cao2019CrossDomainAF}, our SA-Q zero-shot performance is $84.6$ mAP, which is almost on par with the $86.9$ mAP from fully supervised models that also use HRNet-w32 (Figure~\ref{fig:FigSAQ}d). Moreover, and we beat the $79.4$ mAP from zero-shot of HRNet-w32 that is pretrained on the AP-10K dataset.
\medskip

Lastly, on the AP-10K benchmark, we used the official train-val splits. The benchmark officially tests fine-tuning performance on the validation set (see below), but first we tested zero-shot and find with SA-Q a $68.0$ mAP (Figure~\ref{fig:FigSAQ}e), which is already close to the reported $71.4$ mAP that is obtained by a fully fine-tuned ViTPose-S++\cite{Xu2022ViTPoseVT}.

\subsection*{Fine-tuning performance (SA-Q)}

Thus, while the SA-Q shows generally strong zero-shot performance (matching or beating strong supervised baselines), we tested its fine-tuning capacity.
\medskip

On Horse-10 we show in (Figure~\ref{fig:FigSAQ}b) that using memory replay to fine-tune SA-Q, with only 5\% of the training data, we match the ImageNet-transfer-learning baseline with 100\% training data (which is a 20X data efficiency gain). In both the low and high data regime, we significantly outperform ImageNet-transfer-learning baseline (Tables~\ref{horse_ood_lme} and~\ref{contrasts_horse_ood}).
\medskip

On iRodent, SA-Q significantly outperforms ImageNet-transfer-learning baseline in both the low and high data regime (Figure~\ref{fig:FigSAQ}c, Tables~\ref{rodent_lme} and~\ref{contrasts_rodent}). In particular, using memory replay to fine-tune SA-Q gives $73$ mAP with full training data, outperforming ImageNet-transfer-learning by $14.1$ mAP, and can be nearly 100X more data efficient in the low data regime (meaning, one needs nearly 354 images to reach the same zero-shot performance and/or fine-tuning with 3 images).
\medskip

Next, on AnimalPose we show that our fine-tuning SA-Q gives $87$ mAP, slightly better than $86.9$ mAP from ImageNet-transfer-learning (Figure~\ref{fig:FigSAQ}d), and using transfer learning gives $89.6$ mAP, outperforming ImageNet weights by $2.7$ points.
\medskip

%AP-10K: 
AP-10K is one of the most challenging animal pose estimation benchmarks with many strong baselines (e.g., ViTPose++~\cite{Xu2022ViTPoseVT}, UniPose~\cite{yang2023unipose}) (Figure~\ref{fig:FigSAQ}e, f). SA-Q gives $80.1$ mAP after fine-tuning on AP-10K, which is very close to $80.4$ obtained by a top-down baseline ViTPose-L++, which uses a vision transformer that is 10X (307 M parameters) bigger than our architecture (HRNet-w32, that has 29M parameters) and it was pre-trained on 307K pose images, which is then 4.38X more than our 70K pose image dataset (Quadruped-80K excluding 10K AP-10K) (Figure~\ref{fig:FigSAQ}f). We also outperform the $79.0$ mAP obtained by UniPose-V-Swin-L, which is a bottom-up method that is pretrained on 226K pose images (plus previously pre-trained on 400M images for using CLIP weights)~\cite{radford2021learning}. Note, Swin-L~\cite{liu2021swin} has 197M parameters, making it 6X larger than the HRNet-w32 we used. Lastly, we are $34.7$ points higher than $45.4$ mAP reported by another strong  bottom-up method, ED-Pose~\cite{yang2023explicit}, which was pretrained on 154K pose images. Thus, our performance in AP-10K benchmark shows that our approach is not only data-efficient but also parameter-efficient.
Taken together, this means that our methods strong performance is not simply due to more data or bigger networks, its the algorithmic advancements and the animal pose prior from Quadruped-80K. 
\medskip

Collectively, the SuperAnimal method presents an efficient way to achieve strong zero-shot and few-shot performance and also provides better starting weights for fine-tuning (vs. ImageNet-based transfer learning). Of course, despite strong generalization, there can still be failures (Extended Data Figure~\ref{fig:Extended_data_QuadCreate}b). Note that both SuperAnimal models---TopViewMouse and Quadruped---learned to predict the union of all keypoints defined in multiple datasets even if no single dataset might have defined all of these keypoints (i.e., as in TopViewMouse-5k), and even if fine-tuned on data without the super-set they still retain the super-set.

\subsection*{Unsupervised Video Adaptation}

\begin{figure*}[ht!]
\begin{center}
\includegraphics[width=.855\textwidth]{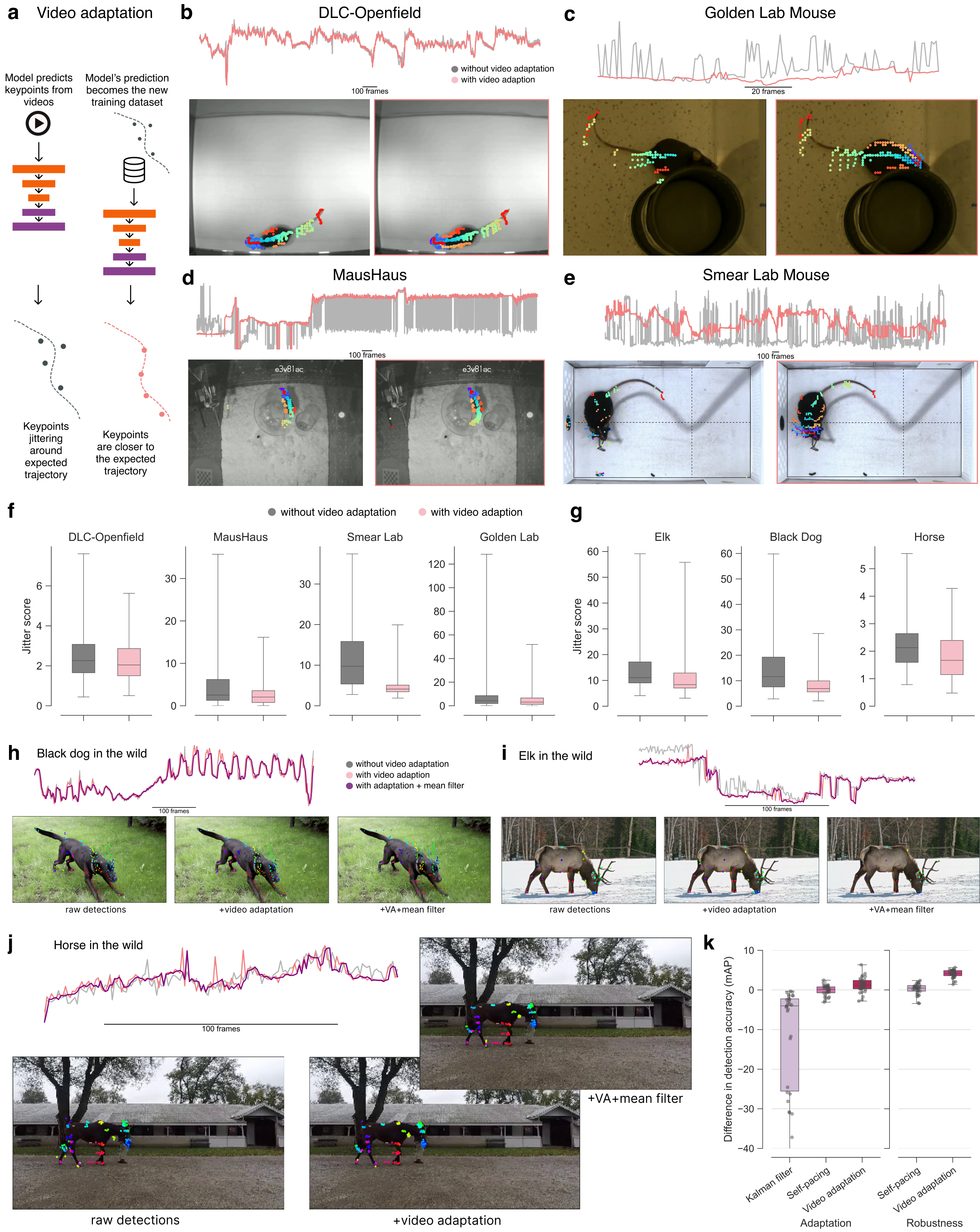}
\caption{\textbf{Unsupervised video adaptation methods.} \textbf{a}: Illustration of the unsupervised video adaptation algorithm.
\textbf{b-e}: Animal size described by convex hull of keypoints using the SA-TVM model. Frequent changes of the convex hull indicates non-smooth keypoint predictions, and below are example images with and without video adaptation showing the trailing keypoints for 10 past frames of data (to demonstrate the motion smoothness).
\textbf{f-g}: Change in jitter score before and after video adaptation. Overall, our method had a significant effect on reducing jitter ($F(1,23286)$=190.03, $p$<.0001; Table~\ref{jitter_lme}, in all but the dog ($p$=.36, $d$=-.03) and Golden lab ($p$=.62, $d$=-.06) videos; Table~\ref{contrasts_jitter}. 
\textbf{h-i}: Same analysis as in \textbf{b-e} using the SA-Q model. Note that an additional median filtering post-video adaptation examples (dark purple line) can be used if needed.
\textbf{k}: Video adaptation, self-pacing and Kalman filter's performance on the Horse-30 video dataset where \textbf{j} is an example of one of 30 videos from the dataset.}
\label{fig:FigVideoAdapt}
\end{center}
\end{figure*}

\justify Independent of the use case (i.e., zero-shot or few-shot fine-tuning), to optimize performance on unseen user data we also developed two unsupervised methods for video inference that help overcome differences in the data SuperAnimal models were trained on compared to what data users might have (Figure~\ref{fig:FigVideoAdapt}a, and Extended Data Figure~\ref{fig:Extended_data_Spatial_Size}a). These so-called distribution shifts can come in various forms (e.g., spatial or temporal; see Methods). 
For example, a bottom-up model can not perform well if the video resolution or animal appearance size is dramatically different those data which we trained on, and the animal datasets are particularly diverse in size, which can pose challenges (Figure~\ref{fig:Extended_data_Spatial_Size}b, c). 
Therefore, inspired by~\cite{lin2017feature}, we developed an unsupervised test-time augmentation called spatial-pyramid search that significantly boosted performance in three OOD videos (Extended Data Figure~\ref{fig:Extended_data_Spatial_Size}c-e, Suppl. Video 3, Table~\ref{ks}; and see Methods). This is unsupervised, as the user does not need to label any data, they simply give a range of video sizes. Note that in practice this does slow down inference time depending on the search parameter space, and this method is not needed with top-down pose models as top-down detection standardizes the size of the animal in both train and test time before the cropped image is seen by the pose models.
\medskip

Secondly, to improve temporal video performance we propose a new unsupervised domain adaptation method (Figure~\ref{fig:FigVideoAdapt}a). Others have considered pseudo-labeling for images but they always required access to the full underlying dataset, which is not practical for users~\cite{kumar2020understanding, rusak2021if, cao2019cross}. Our approach is tailored for pose video adaptation without the need for the ground-truth data. The method runs pose inference on the videos and treats the output predictions as the pseudo ground-truth labels and then fine-tunes the model. 
\medskip
First we used the animal's size (estimated by convex hull formed by animal keypoints, see more details in Methods) as an indicator to measure the improvement in smoothness of video pose predictions. Qualitative performance gain for SA-TVM is shown in Figure~\ref{fig:FigVideoAdapt}b-e. 
\medskip

We also use a jitter score (see Methods) as the indicator to measure whether video adaptation mitigates the jittering that can be seen in pose estimation outputs. Overall, our method had a significant effect on reducing jitter ($F(1,23286)$=190.03, $p$<.0001; Table~\ref{jitter_lme}, in all but the dog ($p$=.36, $d$=-.03) and Golden lab ($p$=.62, $d$=-.06) videos; Table~\ref{contrasts_jitter}, Figure~\ref{fig:FigVideoAdapt}f-j and Suppl. Video 4).
\medskip

To quantitatively measure the improvement of video adaptation, we define adaptation gain and robustness gain (see Methods) to evaluate the method's improvement to the adapted video (a subset of the video dataset) and to the target dataset (all videos in the video dataset). We used Horse-30~\cite{mathis2021pretraining} where 30 videos of horses are densely annotated to evaluate video adaptation (Figure~\ref{fig:FigVideoAdapt}k).
\medskip

We compare our method to Kalman filtering and so-named self-pacing~\cite{cao2019cross} (see Methods), and find that it significantly improves mAP in terms of video adaptation gain (\textit{p}<.003, Cohen's \textit{d}>0.785) and robustness gain (\textit{p}=.0001, Cohen's \textit{d}=3.124; Figure~\ref{fig:FigVideoAdapt}k; Tables~\ref{adapt_gain_aov},~\ref{adapt_gain_contrasts},~\ref{robust_gain_tt}).
\medskip 

Notably, video adaptation outperforms self-pacing by $4$ mAP in terms of robustness gain, demonstrating that it not only adapts to one single video, but to all 30 videos in the dataset. This is important because our method demonstrates successful domain adaptation to the whole video dataset rather than to a single video.
\medskip 

Our method does not take extensive additional time, and practically speaking, can be run during video analysis. For example, if a video (of a given size) can be run at 40 FPS, our video adaptation would slow down processing to approx. 12 FPS, while self-pacing would be closer to 4 FPS (thus slower and less accurate).

\begin{figure*}[ht]
\begin{center}
\includegraphics[width=.98\textwidth]{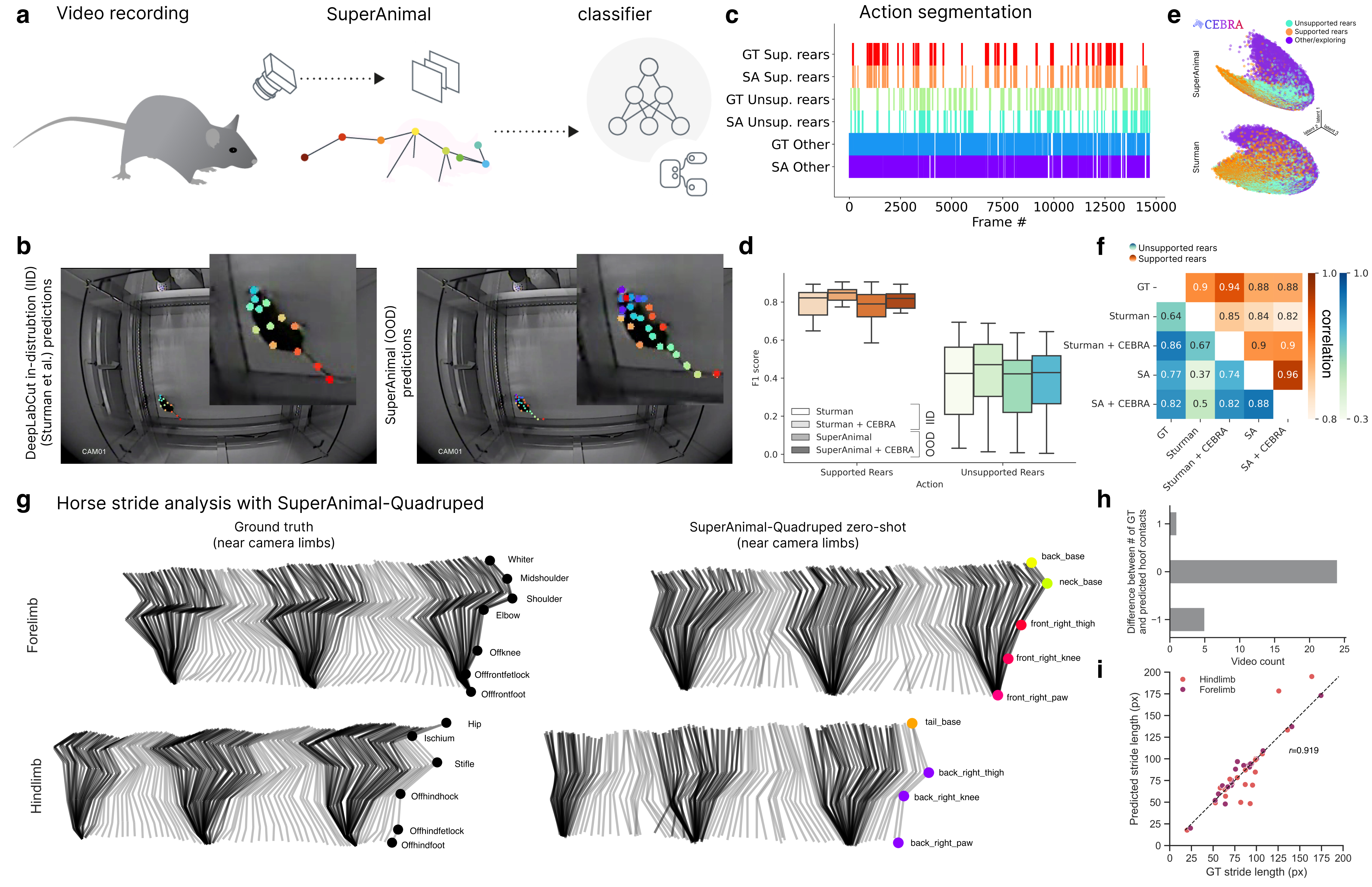}
\caption{\textbf{Zero-shot behavioral quantification with SuperAnimal.}
\textbf{a}: Workflow overview for behavioral analysis with SuperAnimal.
\textbf{b}: Images of the open-source dataset from Sturman et al.~\cite{sturman2020deep} with their DeepLabCut ``in distribution" model and our SuperAnimal zero-shot, out-of-distribution, results.
\textbf{c}: Ethogram comparing ground truth annotations vs. zero-shot predictions from SuperAnimal-TopViewMouse.
\textbf{d}: F1 score computed across IID (Sturman) and SuperAnimal with, or without CEBRA on the two behavioral classes.
\textbf{e}: CEBRA~\cite{schneider2022cebra} embedding on Sturman keypoints and SuperAnimal-based keypoints in 3D, transformed with FastICA for visualization.
\textbf{f}: Correlation matrix that demonstrates the correlation between SuperAnimal-TopViewMouse and ground-truth annotations averaged across 3 annotators and across the model and keypoint configurations. 
\textbf{g}: We analyzed 30 horse videos where every frame had a ground truth (GT) annotation of keypoints~\cite{mathis2021pretraining} (left) vs. our SuperAnimal-Quadruped model (right). The right limbs (closest to the camera) from one representative gait trial are shown. Swing and stance phases are colored in light grey and black zones, respectively.
\textbf{h}: Histogram delineating the number of videos where the ground contact by the hoof were identical to the GT vs. over or under counted by 1 stride (no error larger than 1 was found).
\textbf{i}: We computed the error between the GT stride length vs. model prediction for the hoofs (i.e, right\_back\_paw vs. Offhindfoot, etc). Each dot represents a stride, color denotes hindlimb vs. forelimb, near legs only.
}
\label{fig:action_segmentation}
\end{center}
\end{figure*}

\vspace{-4pt}
\subsection*{Unsupervised behavioral analysis}

\justify To illustrate the value of our zero-shot predictions for behavioral quantification (Figure~\ref{fig:action_segmentation}a), we first turned to an open-source dataset that was used to benchmark the performance of open-source machine learning tools vs. some commercially available solutions~\cite{sturman2020deep}. Specifically, we used the open-field test (OFT) dataset presented in Sturman et al.~\cite{sturman2020deep}. We evaluated the performance of SuperAnimal weights in an action segmentation task. To make OFT out-of-distribution, we made a variant of the SA-TVM model by excluding the OFT dataset during training from full SA-TVM. 
\medskip

As a strong baseline, we used the DeepLabCut keypoints trained by Sturman et al., who trained in a supervised way on each video specifically, thus making it in-domain (Figure~\ref{fig:action_segmentation}a, b).
We asked if the SuperAnimal model variant, which has never been trained on the 20 videos they present, is sufficient to classify two critical kinematic-based postures: unsupported rearing in the open field, and supported rearing against the box wall (Figure~\ref{fig:action_segmentation}a, b, see also Suppl. Video 5). If the keypoints were too noisy, this task would be very challenging. 
\medskip

In order to transform keypoints into behavioral actions via segmentation, we used skeleton-based features to convert keypoints to feature vectors (see Methods). 
We then either used only a MLP-based classifier as in Sturman et al., or we used a newly described non-linear clustering algorithm called CEBRA~\cite{schneider2022cebra} to further improve the feature space, followed by the same classifier (see Methods and Figure~\ref{fig:action_segmentation}c-e). 
\medskip

We found that SA-TVM zero-shot could be as good as the supervised keypoint model in predicting both behaviors (Figure~\ref{fig:action_segmentation}d-g; linear mixed effect model, fixed effect of `method': $F=0.999$, $p=0.393$; Table~\ref{contrasts_lme_oft}). Moreover, using CEBRA slightly improved upon the behavior classification, independent of which keypoints were used (Figure~\ref{fig:action_segmentation}e, f). We also compared the correlation of our result based on SuperAnimal or Sturman keypoint data against the three annotators per video and find that our model is well correlated to the ground truth annotations, particularly when using CEBRA (Figure~\ref{fig:action_segmentation}f).
\medskip

As a further test, we compared the performance of using our keypoints vs. the officially provided keypoints on the MABe benchmark~\cite{sun2023mabe22}. In brief, we used the top-down SA-TVM model and ran inference over three million frames over 1830 videos (without video adaptation to test baseline performance). We then used the outputs and ran PointNet~\cite{pointnet_mabe}, which was provided as a baseline method in MABe. Here, we find near identical performance on the behavioral classification tasks as the fully-supervised pose estimation data they provide (Table~\ref{mabe_results}, Extended Data Figure~\ref{fig:Extended_data_behavior}a). Further suggesting that SA-TVM can be combined with other approaches for mouse behavioral classification.
\medskip

Moreover, since the time of pre-printing this work~\cite{ye2023superanimal}, SA-TVM has been used zero-shot with with post-hoc unsupervised analysis of mouse behavior with Keypoint-MoSeq~\cite{kptmoseq2023} and (both SA-TVM and SA-Q) within AmadeusGPT~\cite{ye2023amadeusgpt}.
Therefore, collectively this demonstrates that without any training, the SA-TVM model can be used for downstream behavioral analysis on out-of-distribution data.
\medskip

Lastly, to show the utility of the SA-Q model in video analysis we performed gait analysis in horses. Here, we turn to a ground truth video dataset where every frame of the video was annotated by an equine expert~\cite{mathis2021pretraining}. We computed the stride and swing phase of the gait and show that the SA-Q model with video adaptation can match ground truth (Figure~\ref{fig:action_segmentation}g-h, and with filtering see Extended Data Figure~\ref{fig:Extended_data_behavior}b) in 24 out of 30 videos, where we only miss one stride detection (either over or under, (Figure~\ref{fig:action_segmentation}h). We also computed the hoof-ground contacts and find generally good agreement between ground truth and predictions ($r=0.919$; Figure~\ref{fig:action_segmentation}i). The fraction of contacts within 1–5 frames of ground truth was 69.9\%-81.7\%, respectively, averaged across front and hind limbs across all videos. 
Collectively, this suggests our SuperAnimal models can be used in real-world tasks both in and outside the laboratory.

\vspace{-6pt}
\section*{Discussion}
 
We propose an approach to create robust, cross-lab neural network models that are applicable for rodents and many other quadrupeds (>45 species).  Our approach is general, and it will be an important future goal to expand the DeepLabCut Model Zoo to additional animals (e.g., insects, birds, or fish) and behavioral contexts. moreover, what keypoints are of relevance also depends on the experiment. For instance, in reaching experiments~\cite{lemon2008descending,mathis2018deeplabcut}, different keypoints are of interest than in open field studies, but many groups could still benefit from such collective model building efforts.
\medskip

Building a pretrained pose model via supervised learning benefits from the availability of the annotated pose datasets, and we show that our formulation removes the obstacles of leveraging inhomogenous pose datasets, which enables SuperAnimal models to benefit from learning pose prior from larger datasets. Alternatively, unsupervised keypoint discovery can be used~\cite{sun2022self,bala2023self}. While the unsupervised approach requires no pose annotations, the learned keypoints might lack interpretability and it is not clear yet whether it allows zero-shot inference on OOD data. Therefore, both approaches that create predictions based on the super-set of annotated keypoints from different studies and unsupervised keypoint discovery are promising, complementary directions.
\medskip

Moreover, labs may be more incentivized to share their data knowing their work can be leveraged by a global community effort to build more powerful models. The DeepLabCut Model Zoo web platform allows access to SuperAnimal pre-trained models, aids in collecting and labeling more data (Extended Data Fig~\ref{fig:Extended_data_behavior}d), and hosts other user-shared models at \url{http://modelzoo.deeplabcut.org}.
\medskip 

Taken together, we aimed to reduce the (human and computing) resources needed to create or adapt animal pose models in both lab and in-the-wild animal studies, thereby increasing access to critical tools in animal behavior quantification. We developed a new framework called panoptic pose estimation, where models can be used across various environments in a zero-shot manner and if fine-tuned, they require 10-100$\times$ less labeled data than previous models (Figures~\ref{fig:Fig1overviewASTVM}, ~\ref{fig:FigSAQ}). They also show increased performance compared to ImageNet transfer learning, plus we demonstrate their ability to be used in down-stream tasks such as behavioral classification (Figure~\ref{fig:action_segmentation}), suggesting they could become foundation models for animal pose estimation.
\end{multicols}

\section*{References}
\bibliography{refs}

\vspace{-4pt}
\subsection*{Acknowledgments} The authors thank Niels Poulsen, Mu Zhou, Lucas Stoffl and Shilong Zhang for code assistance, or discussions and feedback. We also thank Ole Kiehn, Jared Cregg, Carmelo Bellardita, Johannes Bohacek, Sam Golden and Nastacia Goodwin for generously sharing data on OpenBehavior or zenodo.org.
Funding was provided by the EPFL, CZI grant DAF2020-207363 from the Chan Zuckerberg Initiative DAF, an advised fund of Silicon Valley Community Foundation, The Vallee Foundation (MWM), SNSF grant no. 310030\_201057 (MWM), a Novartis Foundation for Medical-Biological Research Young Investigator Grant (MWM). MWM is the Bertarelli Foundation Chair of Integrative Neuroscience.

\subsection*{Author Contributions} 
Conceptualization: MWM, AM, SY;
Methodology: SY, AF, JL, StS, MWM, AM; 
Software: SY, AF, JL, MV, StS, AM, MWM; 
Investigation: SY, AF, JL, AM, MWM;
Dataset Curation: MWM, AM, SY, TQ;
MausHaus data collection and labeling: MWM; 
iNaturalist data curation and labeling: TQ;
WebApp: StS, MV;
Writing-Original Draft: MWM, SY; 
Writing-Editing: MWM, AM, SY, MV, StS, JL, AF.
\medskip

\textbf{Conflicts:} M.W. Mathis and S. Schneider are co-founders at Kinematik AI. The other authors declare no conflicts of interest. The funders had no role in the conceptualization, design, data collection, analysis, decision to publish, or preparation of the manuscript.

\subsection*{Data Availability}
The packaged datasets are detailed in the ``Datasheets''. SuperAnimal model weights are banked at HuggingFace: \url{https://huggingface.co/mwmathis/DeepLabCutModelZoo-SuperAnimal-Quadruped} and \url{https://huggingface.co/mwmathis/DeepLabCutModelZoo-SuperAnimal-TopViewMouse}, and see detailed ``Model Cards''.

\subsection*{Code Availability}
Code to use the DeepLabCut Model Zoo: \url{https://github.com/DeepLabCut/DeepLabCut}; it is available since version 2.3.1. Code and data to reproduce the figures: \url{https://github.com/AdaptiveMotorControlLab/modelzoo-figures}. All other requests should be made to the corresponding author.

\setlength{\parskip}{\medskipamount}

\newpage

\section*{Methods}

\section*{Datasets}

We collected publicly available datasets from the community, as well as provide two new datasets for showing how to build models with the SuperAnimal method, iRodent and MausHaus, as described below. Thereby, we sought to cover diverse individuals, backgrounds, scenarios, and postures. We did not modify the source data otherwise. In the following we detail the references for those datasets.

\subsection*{TopViewMouse-5k}\
%\url{ https://zenodo.org/record/3608658/export/json#.Y6IVU-zMJAd}
\textbf{3CSI, BM, EPM, LDB, OFT} See full details at~\cite{sturman2020deep} and in~\cite{lukas2020mouse}.
\textbf{BlackMice} See full details at~\cite{isaac_chang_2020_3955216}.
%\url{ https://zenodo.org/record/4556331#.Y6IVmOzMJAd}
\textbf{WhiteMice} See details in SIMBA~\cite{Nilsson2020.04.19.049452}. Courtesy of Prof. Sam Golden and Nastacia Goodwin.
\textbf{TriMouse} See full details at~\cite{Lauer2022multi}.
\textbf{DLC-Openfield} See full details at~\cite{mathis2018deeplabcut}.
\textbf{Kiehn-Lab-Openfield, Swimming, and treadmill} See details at~\cite{Cregg2020BrainstemNT}. Courtesy of Prof. Ole Kiehn, Dr. Jared Cregg, and Prof. Carmelo Bellardita.
\textbf{MausHaus}
 We collected video data from five single-housed C57BL/6J male and female mice in an extended home cage, carried out in the laboratory of Mackenzie Mathis at Harvard University and also EPFL (temperature of housing was 20-25C, humidity 20-50\%). Data were recorded at $30$Hz with $640 \times 480$ pixels resolution acquired with White Matter, LLC eV cameras. Annotators localized 26 keypoints across 322 frames sampled from within DeepLabCut using the k-means clustering approach~\cite{nath2019deeplabcut}. All experimental procedures for mice were in accordance with the National Institutes of Health Guide for the Care and Use of Laboratory Animals and approved by the Harvard Institutional Animal Care and Use Committee (IACUC) (n=1 mouse), and  by the Veterinary Office of the Canton of Geneva (Switzerland; license GE01) (n=4 mice).  MausHaus data is banked at: pending acceptance.

 Moreover, for ease of use, we packaged these datasets into one directory, which is banked at: pending acceptance

\subsection*{Quadruped-80K}

\textbf{AwA-Pose} Quadruped dataset, see full details at~\cite{Banik2021AND}.
\textbf{AnimalPose} See full details at~\cite{Cao2019CrossDomainAF}.
\textbf{AcinoSet} See full details at~\cite{Joska2021AcinoSetA3}.
\textbf{Horse-30} Horse-30 dataset, benchmark task is called Horse-10; See full details at~\cite{mathis2021pretraining}.
\textbf{StanfordDogs} See full details at~\cite{KhoslaYaoJayadevaprakashFeiFei_FGVC2011, biggs2018creatures}.
\textbf{AP-10K} See full details at~\cite{yu2021ap}.
\textbf{APT-36K} See full details at ~\cite{yang2022apt}
\textbf{iRodent} We utilized the iNaturalist API functions for scraping observations with the taxon ID of Suborder Myomorpha~\cite{iRodent}. The functions allowed us to filter the large amount of observations down to the ones with photos under the CC BY-NC creative license. The most common types of rodents from the collected observations are Muskrat (Ondatra zibethicus), Brown Rat (Rattus norvegicus), House Mouse (Mus musculus), Black Rat (Rattus rattus), Hispid Cotton Rat (Sigmodon hispidus), Meadow Vole (Microtus pennsylvanicus), Bank Vole (Clethrionomys glareolus), Deer Mouse (Peromyscus maniculatus), White-footed Mouse (Peromyscus leucopus), Striped Field Mouse (Apodemus agrarius). We then generated segmentation masks over target animals in the data by processing the media through an algorithm we designed that uses a Mask Region Based Convolutional Neural Networks(Mask R-CNN)~\cite{he2017mask} model with a ResNet-50-FPN backbone \cite{https://doi.org/10.48550/arxiv.1612.03144}, pretrained on the COCO datasets~\cite{cocodataset}. The processed 443 images were then manually labeled with pose annotations, and bounding boxes were generated by running Mega Detector~\cite{mega_detector} on the images, which were then manually verified. iRodent data is banked at \url{https://zenodo.org/record/8250392}.

Moreover, for ease of use, we packaged these datasets into one directory, which is banked at: pending acceptance

\subsection*{Additional OOD Videos}

In Figure~\ref{fig:FigVideoAdapt}, for video testing we additionally used the following data:
\textbf{Golden Lab mouse}: see details at~\cite{GoldenLabMouse}.
\textbf{Smear Lab Mouse}: see details at~\cite{SmearMouse}.
\textbf{Mathis Lab MausHaus}: New video conditions, but the same as MausHaus ethics approval as above. \textbf{BlackDog}: video from \url{https://www.pexels.com/video/unleashing-the-pet-dog-outdoors-4763071/}, \textbf{Elk} video from  \url{https://www.pexels.com/video/a-deer-looking-for-food-in-the-ground-covered-with-snow-3195531/}. \textbf{Horse-30 videos}: we used the ground truth annotations for 30 horse videos as described~\cite{mathis2021pretraining}.

\subsection*{Benchmarking: Data splits and training ratios}

\justify \textbf{Pre-training datasets} For every test of an OOD dataset we create a dataset that has all datasets that exclude the OOD dataset.
Within the pretraining datasets, we used 100\% of the images and annotations, and we use the OOD datasets for performance evaluation.

\textbf{OOD datasets} For AP-10K, we used the official training and validation set. For AnimalPose, iRodent, and DLC-Openfield, we create our own splits and shuffles. We use 80:20 train test ratio for AnimalPose and iRodent and we use 95:5 train test ratio for DLC-Openfield. 

We released the leave-one-out datasets here in order for other to benchmark their models in the future: 

\begin{itemize}
    \item SA-QminusDLC-OpenfieldandTriMouse: pending acceptance
    \item SA-QminusHorse30: pending acceptance
    \item SA-QminusiRodent: pending acceptance
    \item SA-QminusAP: pending acceptance
    \item SA-QminusAP-10K: pending acceptance
\end{itemize}

\section*{Panoptic pose estimation}

We cast animal pose estimation as panoptic segmentation~\cite{kirillov2019panoptic} on the animal body; i.e., every pixel on the body is potentially a semantically meaningful keypoint that has an individual identity. Ideally, an infinite collection of diverse pose datasets covers this and the union of keypoints that are defined across datasets make the label space of panoptic pose estimation.

\subsection*{Data conversion and panoptic vocabulary mapping (generalized data converter)}
\label{gdc}

Data came from multiple sources and in multiple formats. To homogenize different annotation formats (COCO-style, DeepLabCut format, etc.), we implemented a generalized data converter. We parsed more than 20 public datasets and re-formatted them into DeepLabCut projects. Besides data conversion, the generalized data converter also implements key steps for the panoptic animal pose estimation task formulation. These steps include:

\begin{enumerate}[noitemsep,nolistsep]
    \item \textbf{Hand-crafted conversion mapping.} The same anatomical keypoint might be named differently in different datasets, or different anatomical locations might correspond to different labels in different datasets. Thus, the generalized data converter used a hand-crafted conversion mapping (see Extended Data Figures~\ref{fig:Extended_data_fig1}a, ~\ref{fig:Extended_data_QuadCreate}) to enforce a shared vocabulary among datasets. We checked the visual appearance of keypoints to determine whether two keypoints (in different datasets) should be regarded as identical. In such cases, the model had to learn (possible) dataset-bias in a data-driven way. We can also think of it as a form of data augmentation that randomly shifts the coordinate of keypoints by a small magnitude, which is the case for keypoints which most dataset creators agree on (e.g., keypoints on the face). For keypoints on the body, the quality of the conversion table can be critical for the model to learn a stable representation of animal bodyparts.
    \item \textbf{Vocabulary projection.} After the conversion mapping is made, keypoints from various datasets were projected to a super-set keypoint space. Every keypoint became a one-hot vector in the union of keypoint spaces of all datasets. Thereby the animal pose vocabularies were unified.
    \item \textbf{Dataset merging.} After annotations were unified into the super-set annotation space, we merged annotations from datasets by concatenating them into a collection of annotation vectors. Note that if the images only displayed a single species, we essentially built a specialized dataset for that species in different cage and camera settings. If there were multiple species present, we essentially grouped them in a species-invariant way to encourage the model to learn species-agnostic keypoint representations, as is the case for our SuperAnimal-Quadruped model.
\end{enumerate}

\section*{The SuperAnimal algorithmic enhancements for training and inference}

\subsection*{Keypoint gradient masking}

First we manually verified a semantic mapping of the datasets with diverse naming (i.e., nose in dataset 1 and snout in dataset 2). Then, we defined a master keypoint space naming, where no one dataset needed to have all the named identified. This  yielded sparse keypoint annotations into the super-set keypoint space (Extended Data Figs.~\ref{fig:Extended_data_fig1}b, c).
Training naively on these projected annotations would harm the training stability, as the loss function penalizes undefined keypoints, as if they were not visible (i.e., occluded).

For stable training of our panoptic pose estimation model, we mask components of the loss function across keypoints.
The keypoint mask $n_k$ is set to $1$ if the keypoint $k$ is present in the annotation of the image and set to 0 if the keypoint is absent.
We denote the predicted probability for keypoint $k$ at pixel $(i,j)$ as $p_k(i,j) \in [0, 1)$ and the respective label as $t_k(i,j) \in \{0, 1\}$, and formulate the masked $L_k$ error loss function as 

\begin{equation}
\mathcal{L}_\mathrm{L_k} =   \sum_{k=1}^m \sum_{i,j} n_k \cdot \| p_k(i,j)  - t_k(i,j) \|^k,
\end{equation}
with $k=2$ for mean square error and $k=1$ for L1 loss (e.g. used for locref maps in DLCRNet~\cite{Lauer2022multi}) and the masked cross-entropy loss function as 
\begin{equation}
\mathcal{L}_\mathrm{CE} =   \sum_{k=1}^m \sum_{i,j} n_k t_k(i,j) \log p_k(i,j). 
\end{equation}
Note that we make distinct the difference between not annotated and not defined in the original dataset and we only mask undefined keypoints. This is important as, in the case of sideview animals, ``not annotated'' could also mean occluded/invisible. Adding masking to not annotated keypoints will encourage the model to assign high likelihood to occluded keypoints.

Also note that the network predictions $p_k(i,j)$ are generated by applying a softmax to the logits $l_k(i,j)$ across all possible keypoints, including masked ones:
\begin{equation}
    p_k(i,j) = \frac{\exp l_k(i,j)}{\sum_{j = 1}^{M} \exp l_j(i,j)}.
\end{equation}
The masking in the loss function then ensures that probability assigned to non-defined keypoints is neither penalized nor encouraged during training.

\subsection*{Automatic keypoint matching}

In cases where users want to apply our models to an existing, annotated pose dataset, we recommend to use our keypoint matching algorithm. This step is important because our models define their own vocabulary of keypoints that might differ from the novel pose dataset. To minimize the gap between the model and the dataset, we propose a matching algorithm to minimize the gap between the models' vocabulary and the dataset vocabulary. Thus, we use our model to perform zero-shot inference on the whole dataset. This gives pairs of prediction and ground-truth for every image. Then, we cast the matching between models' predictions (2D coordinates) and ground-truth as bipartite matching using the Euclidean distance as the cost between paired of keypoints. We then solve the matching using the Hungarian algorithm. Thus for every image, we end up getting a matching matrix where 1 counts for match and 0 counts for non-matching. Because the models' predictions can be noisy from image to image, we average the aforementioned matching matrix across all the images and perform another bipartite matching, resulting in the final keypoint conversion table between the model and the dataset (example affinity matrices are shown in Figure~\ref{fig:Extended_data_matching_ATP}a,b). 

Note that the quality of the matching will impact the performance of the model, especially for zero-shot. In the case where, e.g., the annotation \textit{nose} is mistakenly converted to keypoint \textit{tail} and vice versa, the model will have to unlearn the channel that corresponds to nose and tail (see also case study in Mathis et al.~\cite{Mathis2020APO}).
For evaluation metrics such as mAP where a per keypoint sigma is used, we sample the sigmas from the SuperAnimal sigmas (See Table~\ref{tab:mAP_sigma}).

\subsection*{Memory replay fine tuning}

Catastrophic forgetting~\cite{kirkpatrick2017overcoming} describes a classic problem in continual learning~\cite{van2020brain}. Indeed, a model gradually loses its ability to solve previous tasks after it learns to solve new ones. 

Fine-tuning a SuperAnimal models falls into the category of continual learning: the downstream dataset defines
potentially different keypoints than those learned by the models. Thus, the models might forget the keypoints they learned and only pick up those defined in the target dataset. Here, retraining with the original dataset and the new one, is not a feasible option as datasets cannot be easily shared and more computational resources would be required. 

To counter that, we treat zero-shot inference of the model as a memory buffer that stores knowledge from the original model. When we fine-tune a SuperAnimal model, we replace the model predicted keypoints with the ground-truth annotations, resulting in hybrid learning of old and new knowledge. The quality of the zero-shot predictions can vary and we use the confidence of prediction (0.7) as a threshold to filter out low confidence predictions. With the threshold set to 1, memory replay fine-tuning becomes naive-fine-tuning.

Memory replay pseudo-code: 
\begin{lstlisting}[language=Python]
def is_defined(keypoints):
	# check whether the original dataset defines each keypoint. We use a flag `-1` to denote that a given keypoint is not defined in the original dataset. Note this is different from not annotated, which use flag `0`
	return True if keypoints[2] >= 0 else False

def load_pseudo_keypoints(image_ids):
	# get the pseudo keypoints by image IDs. 
	# note, pseudo keypoints are loaded from disk and fixed throughout the process, so not drifting as is expected in typical online pseudo labeling
	return pseudo_keypoints

def get_confidence(keypoints):
	# get the model confidence of the predicted keypoints. Unlike ground truth data that have 3 discrete flags, predicted keypoints have confidence that can be used as likelihood readout for post-inference analysis
	return keypoints[2]

def memory_replay(model, superset_gt_data_loader, optimizer, threshold):

    # gt data is preprocessed such that annotations are now in superset keypoint space. 
    # every gt keypoint has 3 flags (-1: not defined, 0: not labeled, 1: annotated)

    For batch_data in superset_gt_data_loader:

    gt_keypoints = batch_data['keypoints']	
    image_ids = batch_data['image_ids']
    images = batch_data['images']
    # model() is a pytorch style forward function
    preds = model(images)
    pseudo_keypoints = load_pseudo_keypoints(image_ids)
    # 3 here is (x, y, flag)
    batch_size, num_kpts, 3 = gt_keypoints
    # iterate through batch
    For b_id in batch_size:
	# iterate through keypoints
        For kpt_id in range(num_kpts):
	# since this specific body part is not defined in the new dataset, we use saved pseudo labels (zero-shot prediction) as gt. This prevents catastrophic forgetting and drifting. We can also use confidence to filter the pseudo keypoints
	If not is_defined(gt_keypoints[b_id, kpt_id]) and get_confidence(pseudo_keypoints[b_id][kpt_id]) > threshold:
			# we assume a single animal scenario for simplicity. For multiple animals, matching between gt and pseudo keypoints need to be completed.
			gt_keypoints[b_id][kpt_id] = pseudo_keypoints[b_id][kpt_id]

    loss = criterion(preds, gt_keypoints)
    optimizer.zero_grad()
    loss.backward()
    optimizer.step()
\end{lstlisting}

\section*{Model architectures}

For SuperAnimal-TopViewMouse we used both a bottom-up model (DLCRNet) and top-down model (HRNet-w32), or transformer (AnimalTokenPose) (see below). Whereas for SuperAnimal-Quadruped we only use top-down based HRNet-w32. Please refer to the Extended Data Figure~\ref{fig:Extended_data_Spatial_Size} and Supplementary discussion for why we use only top-down models for quadruped.

\subsection*{Bottom-Up model}

\subsubsection*{DLCRNet}

The SuperAnimal-TopViewMouse used the bottom-up approach as described in DeepLabCut~\cite{mathis2018deeplabcut,Lauer2022multi}.
We use DLCRNet\_ms5~\cite{Lauer2022multi} as the baseline network architecture for its excellent performance on animal pose estimation.
A batch size of 8 was used and the SuperAnimal-TopViewMouse was trained for a total of 750k iterations. In the fine-tuning stage, a batch size of 8 was used for 70k iterations. The Adam optimizer \citep{kingma2014adam} was used for all training instances, and we otherwise used default parameters. We follow DeepLabCut's multi-step learning rate scheduler to drop learning rates three times from $1e-4$ to $1e-5$. Cross-entropy is used for learning heatmaps.
For fine-tuning experiments, we keep the same optimizer, batch size and learning rate scheduler. The total number of training steps is adjusted to 70k iterations.
During video adaptation, we keep the same optimizer and learning rate scheduler, but with batch size 1 and total training steps as 1000. We observe low computational budget as described is sufficient for the model to adapt.

\subsection*{Top-Down models}

\subsubsection*{Object detectors}

For the object detectors, we trained Faster R-CNN using ResNet-50 as the backbone \cite{ren2015faster} and incorporated Feature Pyramid Networks \cite{lin2017feature} for enhanced feature extraction. The training was conducted over $100$ epochs using the AdamW optimizer and LRListScheduler. We initiated the training with a learning rate of $0.0001$, which was decreased to $1e-05$ at the 90th epoch. The batch size was set to 4 for both the SuperAnimal-TopviewMice and SuperAnimal-Quadruped datasets.

We processed the Topview-5K and Quadruped-80K datasets to ensure that there is only one animal category, namely top-view mice or quadrupeds, in each dataset. This approach was adopted to train the model to detect generic animal types effectively. During training, image resizing to 1333$\times$800 pixels, random flipping, normalization, and padding were applied as part of the data augmentation process

\subsubsection*{HRNet-w32}

HRNet-w32~\cite{wang2020hrnet} is used for the top-down based SuperAnimal-Quadruped models. The training protocol follows that described in the AP-10K paper~\cite{yu2021ap}. Specifically, we employed the Adam optimizer~\cite{kingma2014adam} with an initial learning rate of $5e-4$. The total training duration was set to $210$ epochs, with a step decay applied to the learning rate at epochs $170$ and $200$. A batch size of $64$ was used. Consistent with the AP-10K protocol, random flip, half-body transformation, and random scale rotation were applied during training, along with flip testing during evaluation.

For fine-tuning models with a very small number of unique images (e.g., fewer than $64$ images in the training set), we omitted batch normalization and used an initial learning rate of $5e-5$. This setting provided stable training outcomes.

HRNet-w32 was also employed for the top-down based SuperAnimal-TopviewMouse models, adhering to the exact same training protocol as the SuperAnimal-Quadruped.

\subsubsection*{AnimalTokenPose}

Inspired by recent results of Vision Transformers~\cite{Dosovitskiy2021ViT} on human pose estimation tasks~\cite{xu2022vitpose} we assessed ViT's zero-shot performance.  
We conducted experiments with the original ViT architecture in three setups: with masked auto-encoder (MAE) \cite{he2022masked} initialization, DeiT~\cite{touvron2021training} initialization and truncated normal initialization with standard deviation $0.02$ and $0$ mean. Following the original setup~\cite{Dosovitskiy2021ViT}, we did not use a convolutional backbone. The input image of size $224\times224$ was split into patches of $16 \times 16$ pixels, the depth of the transformer encoder was equal to $12$ and each attention layer had $12$ heads with a feature dimension of $768$. It was crucial to use a pre-trained vision transformer; without pre-training, the model did not converge for either dataset (data not shown).

We also adapted the TokenPose model by Yang et al.~\cite{Yang_2021_ICCV}, which adds information about each keypoint in learnable queries called keypoint embeddings. The model was originally used for human pose estimation with a fixed number of keypoints. Combining TokenPose and panoptic animal pose estimation, we obtain AnimalTokenPose models that are able to achieve high zero-shot performance in OOD datasets we prepared (Fig.~\ref{fig:Fig1overviewASTVM} and Figure~\ref{fig:FigSAQ}).

For keypoint estimation, $12$ transformer encoder blocks with feature vector of size $192$ were stacked. While the ViT encoder received raw pixels as an input, in TokenPose~\cite{Yang_2021_ICCV} the images of size $256 \times 256$ are first processed by a convolutional backbone and captured abstract features are then split into patches of size $4 \times 4$. As in TokenPose~\cite{Yang_2021_ICCV}, we used the first $3$ stages of HRNet~\cite{wang2020deep} and $2$ stacked residual blocks from a ResNet~\cite{he2016deep}.

The training procedure for AnimalTokenPose is identical to HRNet-w32  detailed above.

\section*{Video inference methods and considerations}

\subsection*{Domain shifts and unsupervised adaptation}

These domain shifts~\cite{torralba2011unbiased} describe a classical vulnerability of neural networks, where a model takes inputs from a data domain that is dissimilar from the training data domain, which usually leads to large performance degradation. We empirically observe three types of domain shifts when applying our models in a zero-shot manner. These domain shifts range from pixel statistics shift~\cite{hoffman2018cycada}, to spatial shift~\cite{engstrom2019exploring}, to semantic shift~\cite{torralba2011unbiased, hoffman2018cycada}.
To mitigate those, we applied two methods, test time spatial-pyramid search and video adaptation.

\subsection*{Handling the train and test time resolution discrepancy for bottom-up models}

One notable challenge for our bottom-up models face at inference time is the discrepancy in the animal appearance sizes and image resolutions between train and test stages. Even though scale jitter augmentation is part of most pose estimation frameworks' data augmentation pipeline, including DeepLabCut's~\cite{nath2019deeplabcut,kane2020real,Lauer2022multi}, the model can still have trouble handling dramatic change in the image resolution or the animal appearance sizes.  To further deal with scale changes, we employ spatial-pyramid search at test-time (see below). 
The same challenge happens in fine-tuning stage. The downstream dataset (and the animals present in it) could have a very different animal sizes from the pre-training datasets, causing a distribution shift to the pre-trained models. We thus apply resizing (height 400 pixels and same aspect ratio) to downstream datasets if their sizes are drastically different from our training images.

\subsection*{Test time spatial-pyramid search for bottom-up models}

As bottom-up models do not enforce the standardization of the animal size seen by the pose estimator, the relative animal size (ratio between the animal's bounding box area and the area of the image) seen in the pre-training stage and testing stage can be very different. In other words, the bottom-up model performs best with the animal sizes seen in the training stage. The relative animal size in the test time is unknown and as a result, it can cause performance degradation due to distribution shift. We propose to apply multiple rescaling factors to the test image and aggregate the models' predictions

Therefore, during inference, we build a spatial-pyramid composed of model's predictions for multiple copies of the original image at different resolutions. We use model's confidence as the criterion to filter out the resolutions that give sub-optimal performance and aggregate (taking median) predictions from resolutions that have above-threshold confidence as our final prediction.  

The train-test resolution discrepancy~\cite{touvron2019fixing} has been studied actively and most approach it through multi-resolution fusion~\cite{wang2020deep, lin2017feature,Lauer2022multi}. Previous work mostly focuses on IID setting where the resolution of testing images did not vary considerably from the training images.
Moreover, prior work approaches multi-resolution fusion via deep features, requiring modifications of the architecture and adding more parameters. In contrast, the proposed spatial-pyramid search is designed to aid SuperAnimal models in zero-shot scenario where the resolutions of testing images are most likely out of distribution to our training images. We did not apply multi-resolution fusion via deep features for that requires fixing choice of architectures. On the other hand, commonly used multi-scale testing in IID setting does not need to carefully filter out very noisy predictions. This method can also be used for calibration to find the optimal scale. 

Spatial-pyramid pseudo-code:

\begin{lstlisting}[language=Python] 
def  spatial_pyramid_search(images, model, scale_list, confidence_threshold, cosine_threshold):
	# generate rescaled version of original images with multiple scaling factor
        rescaled_images = rescale_images(images, scale_list)
        preds_per_scale = []
        # gather predictions of the model, assuming the final pred_keypoints are projected to the original image space by the forward function
	for rescaled_image in rescaled_images:
		pred_keypoints = model(rescaled_image)
		preds_per_scale.append(pred_keypoints)

	# using median to get a good estimate of expected keypoint positions
        median_keypoint = get_median_keypoint(preds_per_scale)
        # If the rescaled image is not suitable for the model, we expect the model have a confidence less than a given threshold
        pred_keypoints = filter_by_confidence(pred_keypoints, confidence_threshold )
        # A median filter alone does not remove outliers. After confidence filtering, we compare the remained predictions to the median keypoint and drop the low quality predictions
        pred_keypoints = filter_by_cosine_similarity(pred_keypoints, median_keypoint, cosine_threshold)

        return get_median_keypoints(pred_keypoints)
\end{lstlisting}

\subsection*{Video adaptation} 

To aid SuperAnimal models to adapt to novel videos, we inference the model on the videos, and treat these predictions as the pseudo ground-truth~\cite{tarvainen2017mean} labels to train on. We remove the predictions that have low confidence to filter out unreliable predictions. We found that it is critical to fix the running stats of batch normalization layers during the adaptation training (See supp for more details).  Empirically, 1000 iterations with batch size 1 is sufficient to greatly reduce the jitter. The optimal number of iterations and the confidence threshold are hyperparameters for different videos.

Video adaptation pseudo-code: 

\begin{lstlisting}[language=Python]
def get_pseudo_predictions(frame_id):
	# return pseudo prediction by frame id
		
def video_adaptation(model, video_data_loader, optimizer, threshold):
	for data in video_data_loader:
            # fix the running stats of BN layers
            model.eval() 
		frame_id = data['frame_id']
		Image = data['image']
		pseudo_keypoints = get_pseudo_predictions(frame_id)
		preds = model(image)
		loss = criterion(preds, pseudo_keypoints, mask_by_threshold = threshold)
		optimizer.zero_grad()
		loss.backward()
		optimizer.step()
\end{lstlisting}

\section*{Evaluation metrics}

\subsection*{Supervised metrics for pose estimation}

\subsubsection*{RMSE} Root Mean Squared Error is a metric to measure the distance between prediction and ground truth annotations in pixel space~\cite{Mathis2020APO, mathis2018deeplabcut}. However for pose estimation, it does not take the scale of the image and individuals into consideration and the distance is thus non-normalized. As our data is highly variable, we also sometimes use normalized errors. We use RMSE for the DLC-Openfield benchmarking, as this was the original authors main reported metric. Note that during evaluating RMSE, we do not remove predictions that have low confidence due to occlusion. Therefore, all predictions including outliers are penalized by RMSE.

\subsubsection*{Normalized Error}  For Horse-10 experiments, we compute RMSE between ground-truth annotations and predictions with confidence cutoff 0 (to include all predictions to ensure low confidence predictions are also penalized). The resulted RMSE is then normalized by the eye-to-nose GT distance provided by ~\cite{mathis2021pretraining}).

\subsubsection*{mAP} Mean average precision (mAP) is the averaged precision of object keypoint similarity (OKS)~\cite{ronchi2017benchmarking}:
$$OKS = \frac{\sum\limits_{i = 1}^n\left[\exp \left(-d_i^2 / 2 s^2 k_i{ }^2\right) \delta\left(v_i>0\right)\right]}{\sum\limits_{i = 1}^n\left[\delta\left(v_i>0\right)\right]},$$ where $d_i$ is the Euclidean distances between each corresponding ground truth and detected keypoint and $v_i$ is the visibility flags of the ground truth, $s$ is the object scale and $k_i$ is a per keypoint constant that controls falloff (see full implementation details at \cite{cocodataset}). For lab mice, we used 0.1 for all keypoints following \cite{Lauer2022multi}. For quadruped, we used the sigmas of the 17 keypoints shared with AP-10K~\cite{yu2021ap} and used 0.067 for the rest of animal keypoints. See Table~\ref{tab:mAP_sigma}.
$s$ is the square root of the bounding box area (product of width $X$ height of the bounding box).

\begin{table*}[bh] 
    \begin{center}
    \begin{tabular}{llll}
        \toprule
        \textbf{Body Part} & \textbf{$k$, in pixels} & \textbf{Body Part} & \textbf{$k$, in pixels}\\
        \midrule
        nose & 0.026 & upper\_jaw & 0.067\\
        lower\_jaw & 0.067  & mouth\_end\_right & 0.067\\
        mouth\_end\_left & 0.067 & right\_eye & 0.025 \\
        right\_earbase & 0.067 & right\_earend & 0.067 \\
        right\_antler\_base & 0.067 & right\_antler\_end & 0.067\\
        left\_eye & 0.025 &  left\_earbase & 0.067\\
        left\_earend & 0.067 & left\_antler\_base & 0.067\\
        left\_antler\_end & 0.067 & neck\_base & 0.035\\
        neck\_end & 0.067 & throat\_base & 0.067\\
        throat\_end & 0.067 & back\_base & 0.067\\
        back\_end & 0.067 & back\_middle & 0.035\\
        tail\_base & 0.067 & tail\_end & 0.079\\
        front\_left\_thai & 0.072 & front\_left\_knee & 0.062\\
        front\_left\_paw & 0.079 & front\_right\_thigh & 0.072\\
        front\_right\_knee & 0.062 & front\_right\_paw & 0.089\\
        back\_left\_paw & 0.107 & back\_left\_thigh & 0.107\\
        back\_right\_thai & 0.087 & back\_left\_knee & 0.087\\
        back\_right\_knee & 0.089 & back\_right\_paw & 0.067\\
        belly\_bottom & 0.067 & body\_middle\_right & 0.067\\
        body\_middle\_left & 0.067 \\
        \bottomrule
    \end{tabular}
    \caption{We used the following $k$ values per bodypart for the SuperAnimal-Quadruped evaluation.}
    \label{tab:mAP_sigma}
    \end{center}
\end{table*}

\subsection*{Unsupervised metrics for video prediction smoothness}

\subsubsection*{Convex hull body area measurement}

To evaluate the smoothness of SuperAnimal model predictions in video, we utilize a simple unsupervised heuristic. It computes the area of a polygon encompassing all keypoints, the idea  being that the smoother the detections, the lower the variance of this polygon's area. This is formally noted by \(A_{\text{body}}\), to estimate the animal body area. \(A_{\text{body}}\) is calculated using the convex hull containing all keypoints over time. Let \(\mathcal{K}\) represent the set of all keypoints for the animal at each time step, and \(\text{H}(\mathcal{K})\) denote the convex hull containing all keypoints. The animal body area, \(A_{\text{body}}\), is then given by the area of the convex hull:
\[
A_{\text{body}} = \text{Area}(\text{H}(\mathcal{K}))
\]
where \(\text{Area}(\text{H}(\mathcal{K}))\) is the function that calculates the area of the convex hull \(\text{H}(\mathcal{K})\) containing all keypoints over time.

\subsubsection*{Jittering metric}

We define jittering, denoted by \(\text{J}\), as the average of the absolute values of centered, non-signed speeds across all examples and all keypoints. For a given keypoint \(k\) and example \(e\), the jittering value is computed as follows:
\[
\text{J}_{k, e} = \frac{1}{N_{k, e}} \sum_{i=1}^{N_{k, e}} \left| v_{k, e, i} \right|
\]
where:
\(\text{J}_{k, e}\) is the jittering value for keypoint \(k\) in example \(e\);
\(N_{k, e}\) is the total number of centered, non-signed speed measurements for keypoint \(k\) in example \(e\);
\(v_{k, e, i}\) is the \(i\)-th centered, non-signed speed measurement for keypoint \(k\) in example \(e\).

\subsubsection*{Keypoint dropping metric}

Keypoint drop is a count of the number of keypoints with predicted likelihood below a set threshold for every predicted frame (the cutoff was set to 0.1 for bottom-up models, and 0.05 for top-down models). In practice, low-confidence predictions are dropped to remove predictions that are not reliable or occluded. 

In this work, keypoint dropping is used to complement metrics such as RMSE to evaluate the jittery of predictions or catastrophic forgetting. Note keypoint dropping is only valid for top-view, openfield (almost no occlusion) videos where every keypoint is supposed to be predicted with relatively high confidence. For side-view poses, keypoint dropping is not suitable as many bodyparts are occluded.

Let \(K_{\text{total}}\) be the total number of keypoints in the video sequence, and \(K_{\text{dropped}}\) be the count of keypoints that are below a defined threshold \(T_{\text{threshold}}\) and considered for dropping in environments with little occlusion and a top view.
\[
K_{\text{dropped}}(t) = \sum_{i=1}^{K_{\text{total}}} \delta_{i}(t)
\]
where \(K_{\text{dropped}}(t)\) is the count of keypoints dropped at time \(t\), and \(\delta_{i}(t)\) is an indicator function that returns 1 if the \(i\)-th keypoint is below the threshold at time \(t\), and 0 otherwise:
\[
\delta_{i}(t) = \begin{cases}
1, & \text{if } \text{score}_{i}(t) < T_{\text{threshold}} \\
0, & \text{otherwise}
\end{cases}
\]
where \(\text{score}_{i}(t)\) is the confidence score or measurement of the \(i\)-th keypoint at time \(t\).

\subsubsection*{Adaptation gain (or loss) in mAP}

Every video in Horse-30 dataset is densely annotated. Thus we can calculate the mAP gain on the video after the model is adapted to it. We use the pre-adapted zero-shot mAP as the reference and calculate the difference between the post-adaptation mAP and pre-adaptation mAP.

\subsubsection*{Robustness gain (or loss) in mAP}

We use robustness gain to complement adaptation gain. We calculate the mAP for the adapted models on all 30 videos of Horse-30. This metric is to evaluate whether models obtain general robustness in the target domain vs. overfitting to the specific video they adapt to. A positive gain in robustness also suggests that the method can be used on one video and benefit all other videos in the same dataset.

\subsection*{Video adaptation compared to baselines using supervised metrics (mAP)}

\justify We use HRNet-w32 with the detector we trained to perform video adaptation to inference the videos to obtain pseudo-labels.

For video adaptation algorithm, the prediction confidence threshold is set to 0.5 and we perform video adaptation for 4 epochs for each video it adapts to. The learning rate scheduler and augmentations are identical to HRNet-w32's.

\subsubsection*{PPLO}
    
Progressive Pseudo-label-based Optimization~\cite{cao2019cross} implements iterative pseudo-labeling that follows a curriculum, namely, the pseudo-labeling starts with high confidence prediction, and then train with small confidence predictions, following a hard-to-easy curriculum. We initialize three confidence intervals as [0.9, 0.7, 0.5] and sequentially apply pseudo-labeling to the model for four epochs with each confidence level, making a total of 12 epochs training with PPLO.

The full algorithm of PPLO also requires training on both labeled source data and labeled target data, which the video adaptation does not do. For fairness reasons, we only performed the iterative pseudo-labeling step.

\subsubsection*{Kalman filtering}

We apply a constant-velocity Kalman filter (implemented in \texttt{filterpy} v1.4.5) as post-processing to our pre-adaptation zero-shot pose predictions. As Kalman filtering does not modify the model weights, we do not report the general robustness gain on it.

\subsection*{Statistical analysis}

Linear mixed-effects models were fitted in R~\cite{rcoreteam} using the \texttt{lme4} package (v1.1.31;~\cite{bates2014fitting}). Training data ratio (or, equivalently, the number of images) and fine-tuning methods were defined as fixed effects, whereas the various datasets and shuffles were treated as random effects; random intercepts and slopes were also added at the dataset level. The best models were selected based on the Akaike Information Criterion (AIC); adding complexity did not result in lower AIC, and even led to singular fits, indicative of overfitting. The weight of evidence for an effect was computed using likelihood ratio tests, as well as with \emph{p}-values provided by \texttt{lmerTest} (v.3.1.3). Pairwise contrasts and Cohen's \emph{d} standardized effect sizes were computed with the \texttt{emmeans} package (v.1.8.9), and degrees of freedom estimated with the Kenward-Roger method. Distributions of prediction errors with and without spatial-pyramid search were compared with the two-sample, one-sided (alternative hypothesis: "less") Kolmogorov-Smirnov test. The significance threshold was set at 0.05.

\vspace{-8pt}
\section*{Behavioral Action Segmentation, OFT}

As our benchmark dataset, we used the openfield test (OFT) task from Sturman et al.~\cite{sturman2020deep}. We calculated the same skeleton-based features by concatenating 10 distances between keypoints, six angles, four body areas and two additional boolean variables coding whether the nose and head center were inside the arena, resulting in a 22D vector at each time step. For the action classifier, we used an MLP neural network as the action decoder that acted as a sliding window across 31 time steps to perform action segmentation and used F1 score on supported and unsupported rears as evaluation metrics. As in the original paper, we performed leave-one-out cross-validation on 20 videos and across three annotators.
\medskip

Note that the original model for OFT task from Sturman et al. includes the center and four corners of the mouse cage, which is critical for their handcrafted features to determine the relative distance between the mouse and the walls. As our SuperAnimal models focus on animal bodyparts only, we take the corner coordinates from their released data for the sake of comparison. In practice, those static environmental keypoints can be provided by taking users' inputs via interactive GUI for videos.

For CEBRA~\cite{schneider2022cebra}, we used the model architecture `offset10-model'. The output dimension was set to 32, as found via a simple grid search over the following values: [4, 8, 16, 32]. We trained it for 5000 iterations with batch size 4096, the Adam optimizer, and learning rate 1e-4.

\section*{Behavioral Action Segmentation, MABe}

MABe has two rounds and since only round 2 released videos, we use videos from round 2 as the inputs for our pretrained SA-TVM model. Since our paper focuses on pretrained pose model, we use recommended baselines~\cite{pointnet_mabe,sun2023mabe22} from round 1 that build representation based on pose trajectories instead of RGB-based representation learning baselines (as RGB-based representation learning is known to be better than pose trajectory-based representation~\cite{shah2022pose}.
Videos from MABe round 2 have three mice in videos, therefore we used our top-down version SA-TVM.
The procedure is as follows: we inference our pretrained top-down SA-TVM on all 1,830 videos from round 2, converted the pose results into MABe keypoint file format and ran PointNet code to obtain embeddings.  Finally, we use the official evaluation code to compare the performance between using the official MABe poses obtained from fully supervised learning and poses that are obtained via our models’ zero-shot predictions.

\section*{Behavioral Action Segmentation, Horse Gait Analysis}

Our SA-Q model was run on the videos from Horse-30~\cite{mathis2021pretraining}. The start (2 s) and end (2 s) of each of the 30 videos were removed from the analysis, to ignore instants when the horse is only partially seen. Front and back hoof contacts and lifts were identified using respectively peak and valley detection from the 2D kinematic traces of the front and back hooves. Beforehand, these trajectories were smoothed using a 2nd-order, low-pass, zero-lag Butterworth filter (cutoff=3 Hz) and centered on a keypoint located on the animal's back; this effectively expresses keypoint coordinates in a reference frame stationary relative to the moving horse, facilitating event detection. We extracted fore and hind limb strides between consecutive ground contacts, and stance phases between a contact of one hoof until it is lifted off the ground. Stride lengths (in pixels), stances, and the number of identified hoof contacts were then computed, and qualitatively compared to those obtained using the densely annotated (ground truth) keypoints (Figure~\ref{fig:action_segmentation}g, h, i).

\section*{Code API}

High-level inference API (with optional spatial-pyramid search) for using SuperAnimal models in DeepLabCut:
\begin{lstlisting}[language=Python] 
video_path = 'demo-video.mp4'
superanimal_name = 'superanimal_topviewmouse'
scale_list = range(200, 600, 50)  # image height pixel size range and increment

deeplabcut.video_inference_superanimal(
    [video_path],
    superanimal_name,
    scale_list=scale_list,
    video_adapt=True,  # slow but likely most accurate
)
    
\end{lstlisting}

Low-level API that combines spatial-pyramid search and video adaptation:
\begin{lstlisting}[language=Python] 
from  deeplabcut.modelzoo.api import SpatiotemporalAdaptation
video_path = 'demo_video.mp4'
videotype = 'mp4'
superanimal_name = 'superanimal_topviewmouse'
scale_list = range(200, 600, 50)

adapter = SpatiotemporalAdaptation(
    video_path,
    superanimal_name,
    modelfolder="weight_directory",
    scale_list=scale_list,
)
adapter.before_adapt_inference()
adapter.adaptation_training()
adapter.after_adapt_inference()
\end{lstlisting}

\section*{Web App}

Many labs use DeepLabCut to define, annotate, and refine animal bodyparts, resulting in high quality, diverse keypoint annotations for animals in different contexts~\cite{nath2019deeplabcut,Lauer2022multi}. In order to enable a positive feedback loop to turn the collection of animal pose data and models into a community effort we developed a Web App.

The app is available at \href{https://contrib.deeplabcut.org/}{https://contrib.deeplabcut.org/}. This app allows anyone, within their browser, to a) upload their own image and label, b) annotate community images, c) run inference of available community models on their own data, d) share models to be hosted.
The website is written using JavaScript with the Svelte framework, and the models are run on cloud servers.

\subsubsection*{Data collection}

The website has an upload portal for groups to upload their models and labeled data in DeepLabCut format to help grow the pre-training datasets and allow researchers to build on top of varied models and data. 

\subsubsection*{Annotation}
Additionally, the website hosts a labeling web app that allows users to annotate curated images. The datasets currently available for annotation are from iNaturalist~\cite{inaturalist} and the OpenImage Datase~\cite{kuznetsova2020open}. After selecting which dataset to label, images are displayed successively with the target animal prominently shown in front of an opaque masked background (which can be toggled off). The keypoint set is selected taking into account the species morphology and keypoint value in subsequent analysis. Once the annotation is complete, the data is saved to the database and made available for use in further research.

\subsubsection*{Online inference}

To allow testing DeepLabCut models in the browser, the user selects a few images, which model to run, and receives predictions along with confidence scores for each keypoint. Users are then able to adjust or delete keypoints, as well as download the model weights from HuggingFace. This allows for a quick and hassle-free evaluation of DeepLabCut's capabilities and suitability for specific tasks, making it available to a wider range of users.

\newpage 
\clearpage
\beginsupplement

\section*{\huge Extended Data Figures}

\begin{figure*}[b]
\centering
\includegraphics[width=\textwidth]{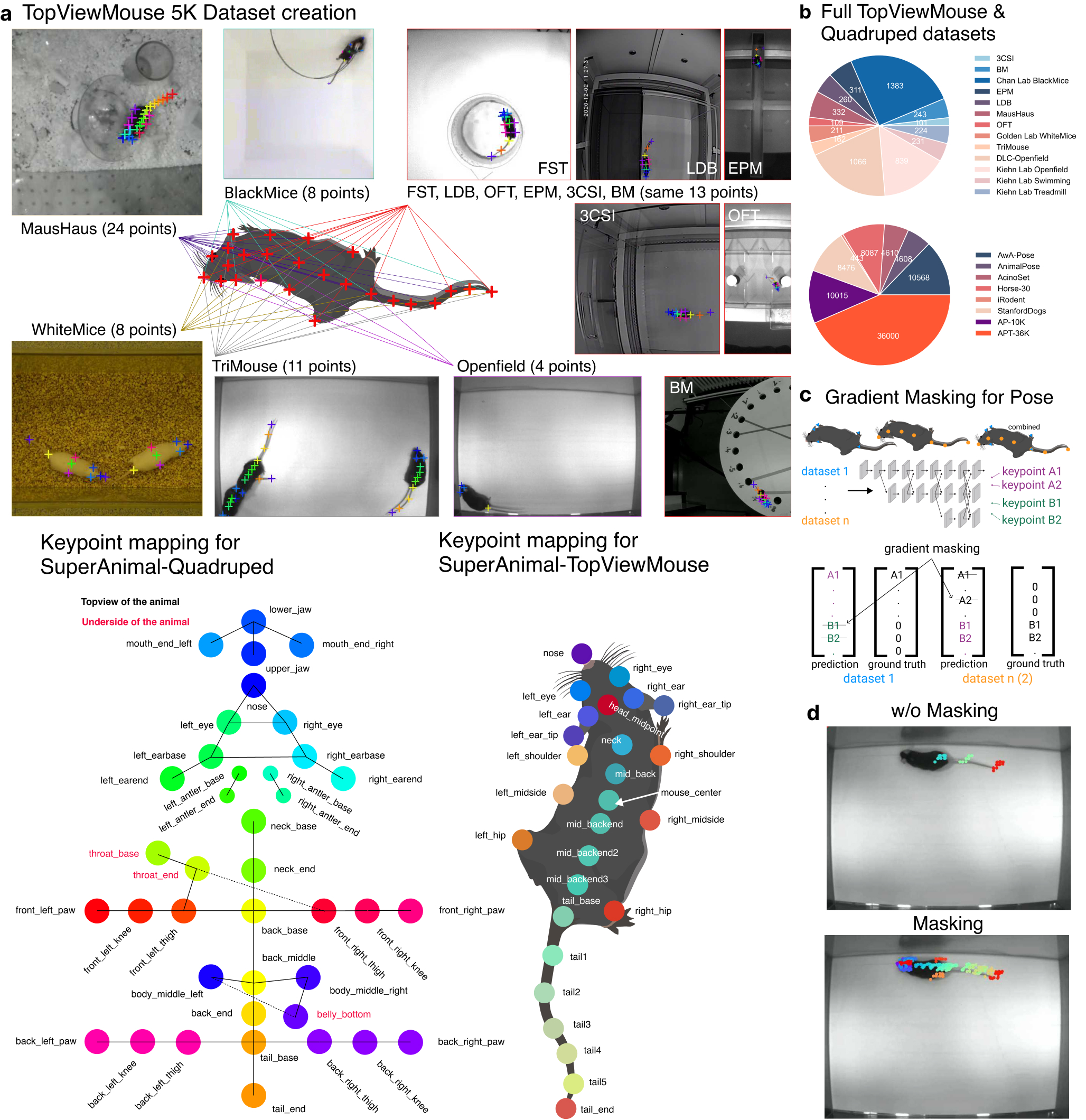}
\caption{\textbf{Constructing SuperAnimal models and keypoint gradient masking.} 
\textbf{a}: Demonstration of how multiple pose datasets are merged into a single dataset. We created a main keypoint names to cover all keypoints we observe from datasets. Then we built a conversion table to map keypoints from each dataset to the main keypoint names. We design a corresponding conversion table such that anatomically similar keypoints are mapped to the same keypoint. Below we add the keypoint naming map for both SuperAnimal-TopViewMouse and SuperAnimal-Quadruped models.
\textbf{b}: Composition of the SuperAnimal-Quadruped (left) and SuperAnimal-TopViewMouse (right) datasets.
\textbf{c}: Demonstration of keypoint gradient masking algorithm. Keypoints that were not defined in the original datasets introduce false penalties for the model training. Therefore, during back-propagation, the gradients of those undefined keypoints are artificially masked. 
\textbf{d}: With masking, the model is able to learn a pose representation that is the union of training datasets. Without masking, the model has severe degraded pose representation.
}
\label{fig:Extended_data_fig1}
\end{figure*}

\begin{figure*}
\includegraphics[width=\textwidth]{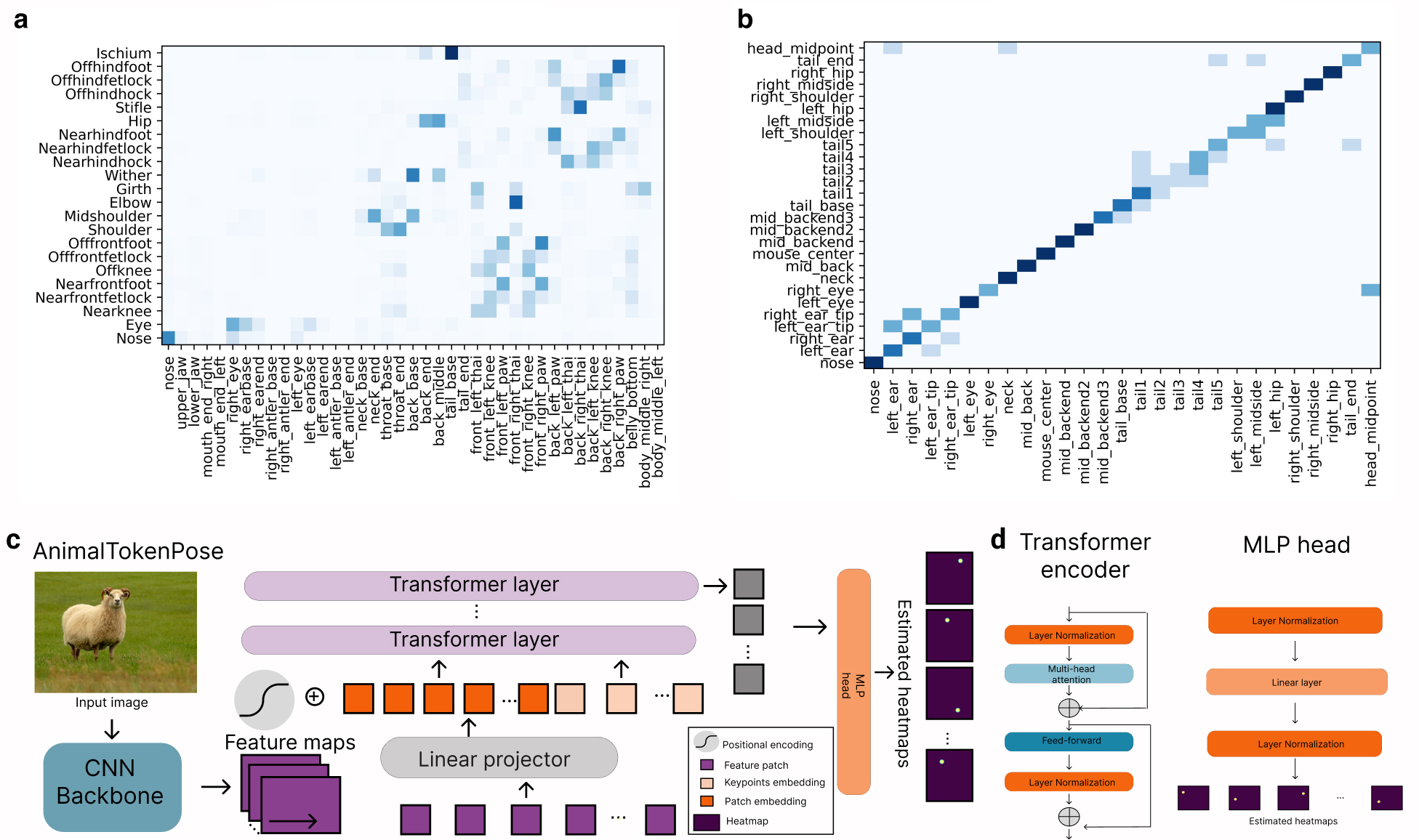}
\caption{\textbf{Keypoint Matching and AnimalTokenPose.} 
\textbf{a}: The affinity matrix represents the semantic similarity between keypoint defined by the model and keypoint defined by dataset annotations across images. The affinity matrix is obtained by hard voting. The voting per image is obtained via pairwise euclidean distance between SuperAnimal-Quadruped model's zero-shot predictions and Horse-30 dataset ground truth. 
\textbf{b}: Affinity matrix for Golden Lab Mouse (see Methods) video (bottom at Figure~\ref{fig:FigVideoAdapt}), where we deliberately tried to match the keypoint space to model's zero-shot prediction. The noise in the affinity matrix suggests annotator bias for hard keypoints (e.g., tail points along the tail where the exact position is not visually concretely defined, as say opposed to the nose). For this analysis we annotated 20 frames of the Golden Lab Mouse data to illustrate our matching process.
\textbf{c}: AnimalTokenPose architecture with additional MLP head for heatmap estimation. 
\textbf{d}: Transformer encoder architecture and MLP head architecture.
} 
\label{fig:Extended_data_matching_ATP}
\end{figure*}

\begin{figure*}
\centering
\includegraphics[width=.9\textwidth]{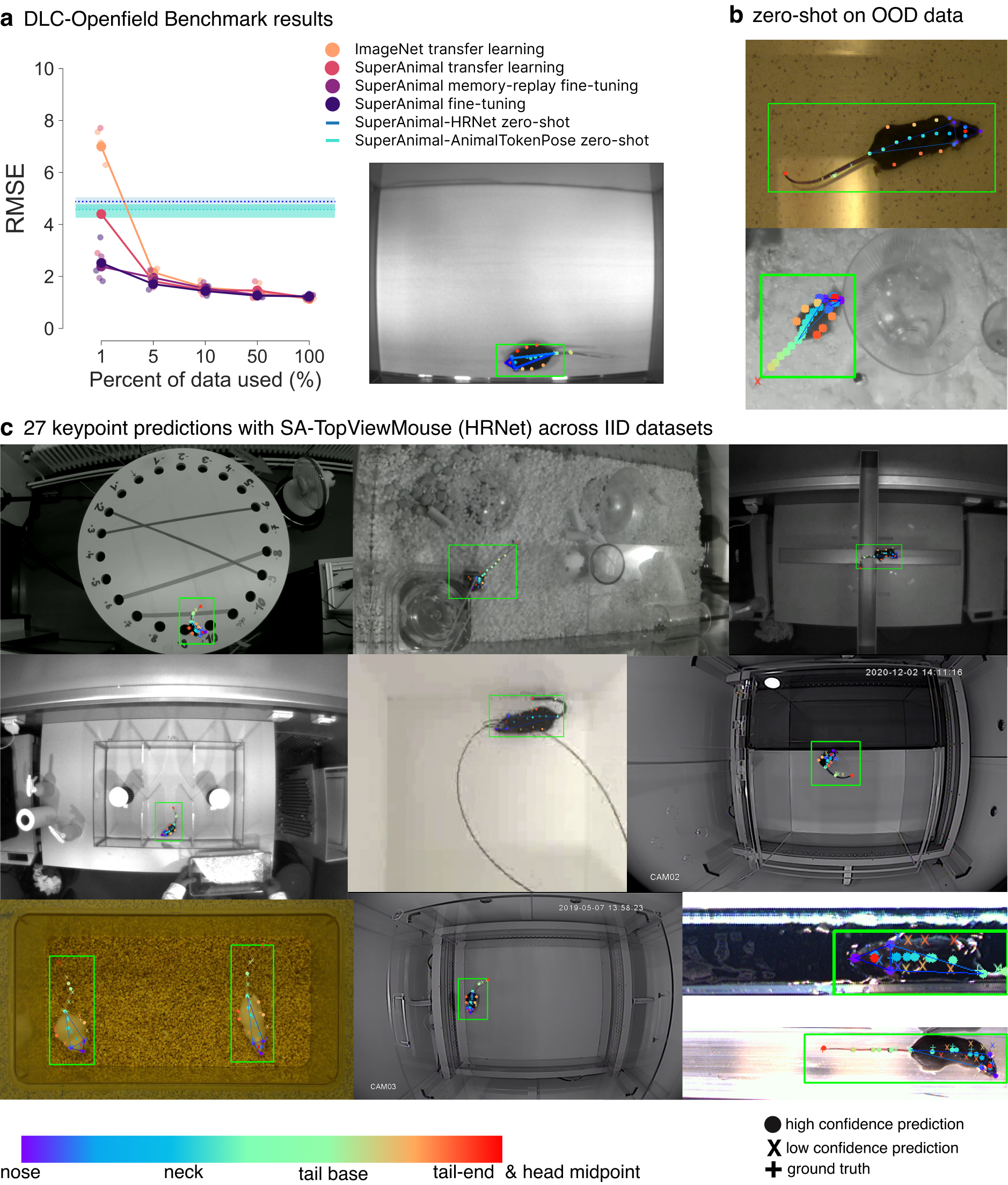}
\caption{\textbf{Top-down HRNet results}
\textbf{a}: SuperAnimal-TopViewMouse using HRNet-w32 on DLC Openfield benchmark. AnimalTokenPose is added as a zero-shot baseline. 1-100\% of the train data is 10, 50, 101, 506, 1012 frames respectively. Blue shadow represents minimum, maximum and blue dash is the mean for zero-shot performance across three shuffles. Large, connected dots represent mean results across three shuffles and smaller dots represent results for individual shuffles. Inset is the qualitative zero-shot performance of SA-TVM. 
\textbf{b}: Qualitative performance. SuperAnimal-TopViewMouse using HRNet on OOD videos (Top: Golden Lab; Bottom: Mathis MausHaus). Confidence cut off is set to be 0.6.
\textbf{c} Qualitative performance. SuperAnimal-TopViewMouse using HRNet on IID images. Confidence cut off is set to be 0.6.
}
\label{fig:Extended_data_TDmouse}
\end{figure*}

\begin{figure*}
\centering
\includegraphics[width=.9\textwidth]{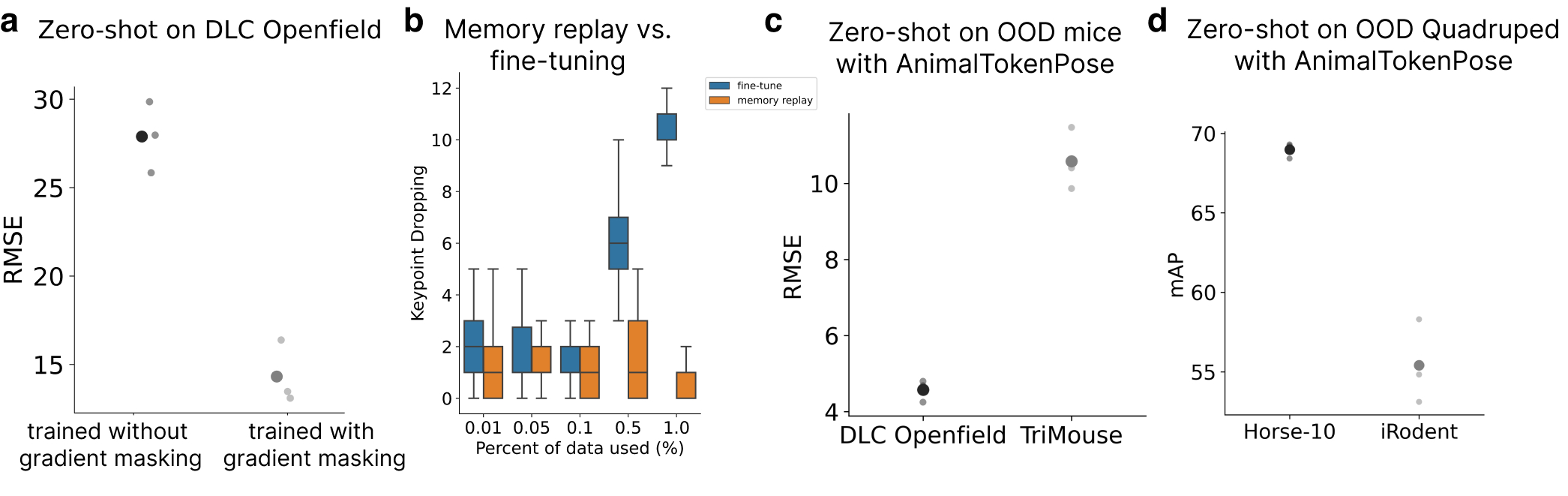}
\caption{
\textbf{a}: SA-TVM (DLCRNet) zero-shot performance on DLC Openfield. Comparison between SA-TVM trained with and without gradient masking.
\textbf{b}: SA-TVM (HRNet-w32) fine-tuning performance on DLC Openfield. Comparison between SA-TVM fine-tuned with memory replay and naive fine-tuning across different training ratios.
\textbf{c} SuperAnimal-TopViewMouse using AnimalTokenPose. Zero-shot performance on DLC Openfield  and TriMouse.
\textbf{d} SuperAnimal-TopViewMouse using AnimalTokenPose. Zero-shot performance on iRodent and Horse-10.
}
\label{fig:Extended_data_SATVMzeroshot}
\end{figure*}

\begin{figure*}
\centering
\includegraphics[width=.75\textwidth]{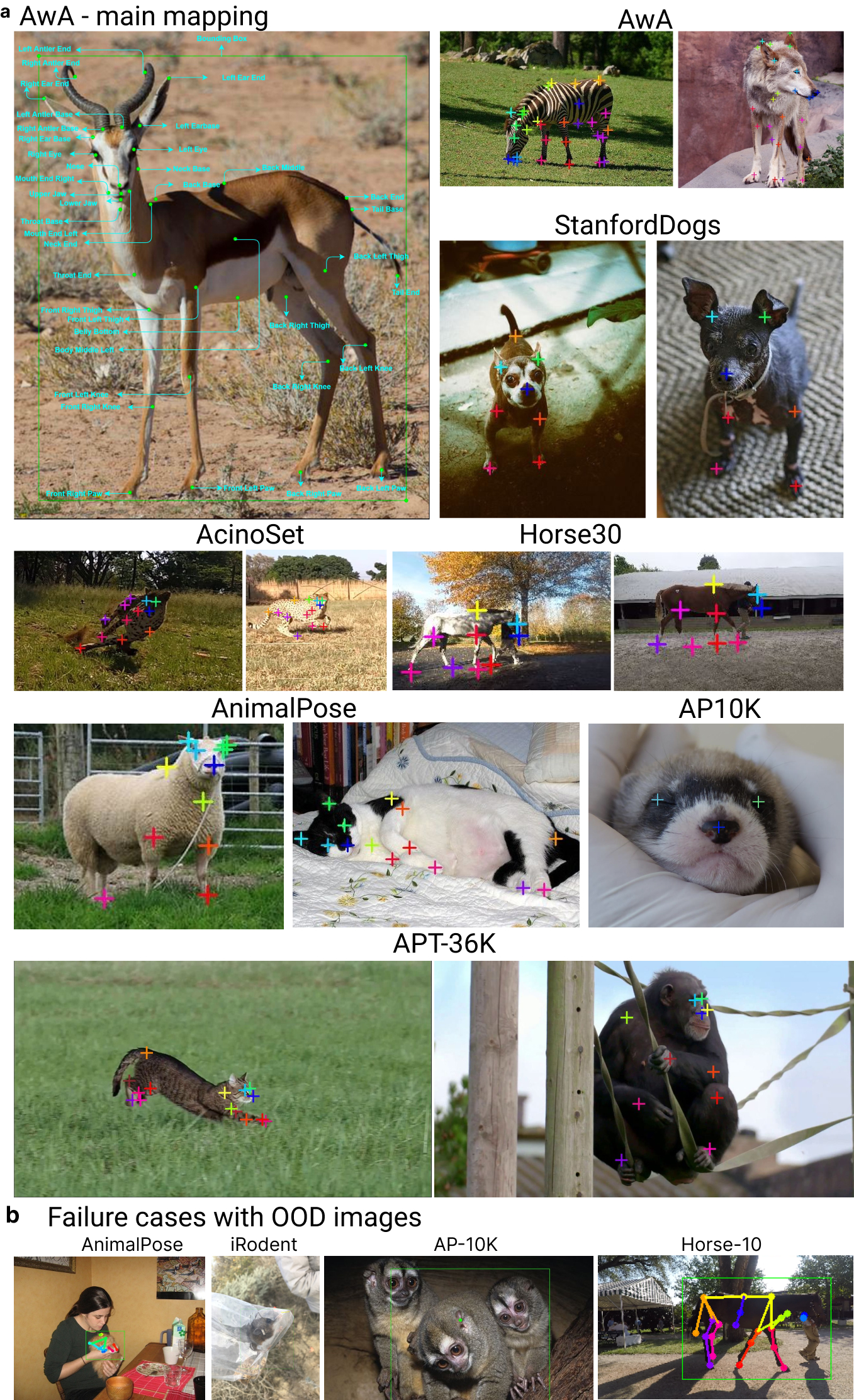}
\caption{\textbf{Quadruped datasets and model failures} 
\textbf{a}: AwA main mapping: Visual illustration of the datasets that compose Quadruped-80K (see Extended Data Figure~\ref{fig:Extended_data_fig1} for TopViewMouse). 
Rest of the examples are illustration of ground-truth annotations for each quadruped dataset in their original keypoint space
\textbf{b}: Examples of OOD failures from several datasets, as noted, from our SuperAnimal-Quadruped model.
} 
\label{fig:Extended_data_QuadCreate}
\end{figure*}

\begin{figure*}
\centering
\includegraphics[width=\textwidth]{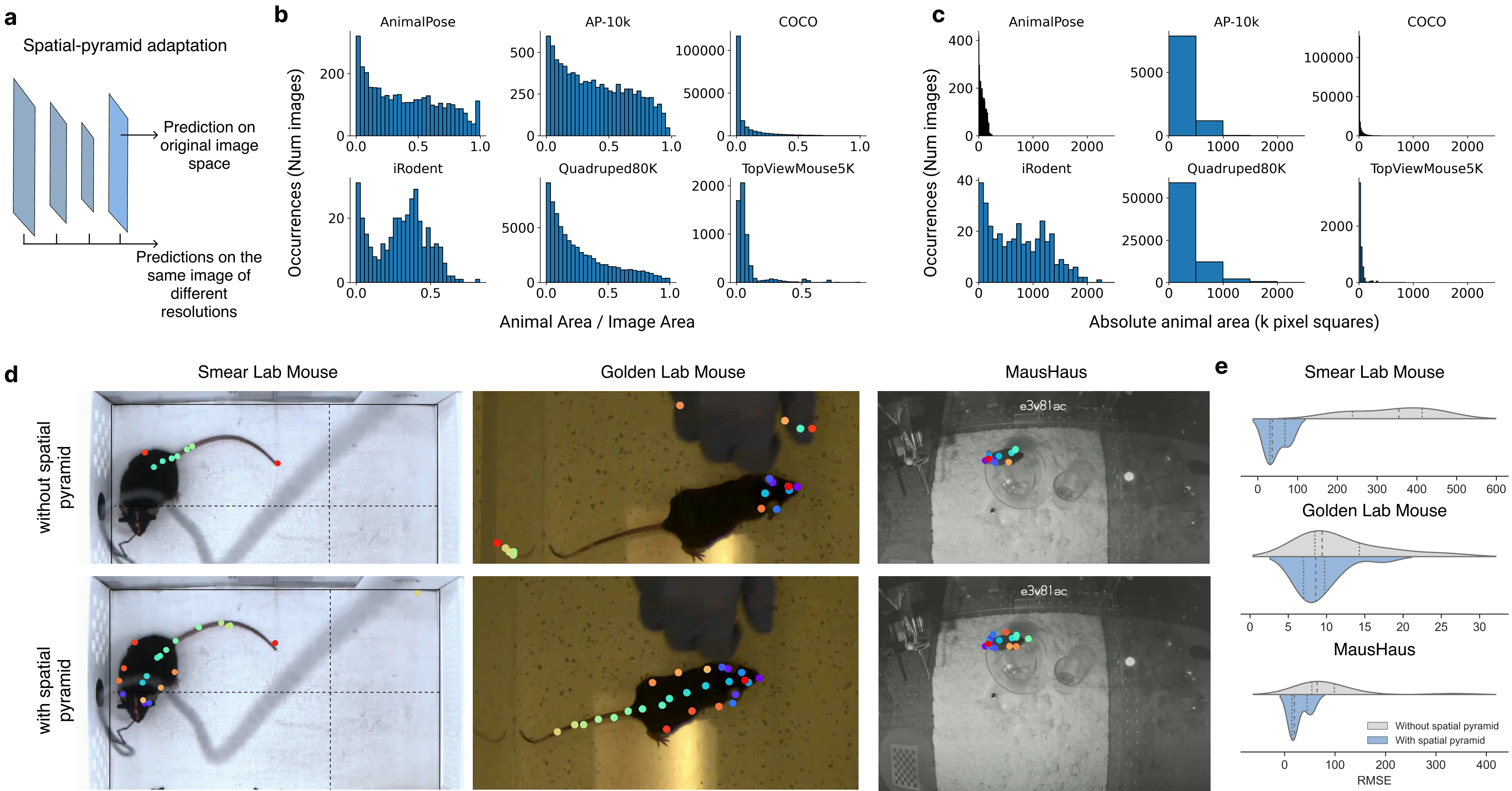}
\caption{\textbf{Challenges of animal appearance sizes}
\textbf{a}: Conceptual diagram to demonstrate that the spatial-pyramid search leverages prediction from multiple resolutions.
\textbf{b}: Relative animal size with respect to the image size in common benchmarks.
\textbf{c}: Absolute animal size (k pixel squares) in common benchmarks.
\textbf{d}: Bottom-up SuperAnimal-TopViewMouse model (i.e., DLCRNet) was used to infer poses on three OOD videos. Visual inspection shows zero-shot inference with vs. without the spatial-pyramid search. 
\textbf{e}: Quantitative results between with and without spatial-pyramid adaptation.
}
\label{fig:Extended_data_Spatial_Size}
\end{figure*}

\begin{figure*}
\centering
\includegraphics[width=.9\textwidth]{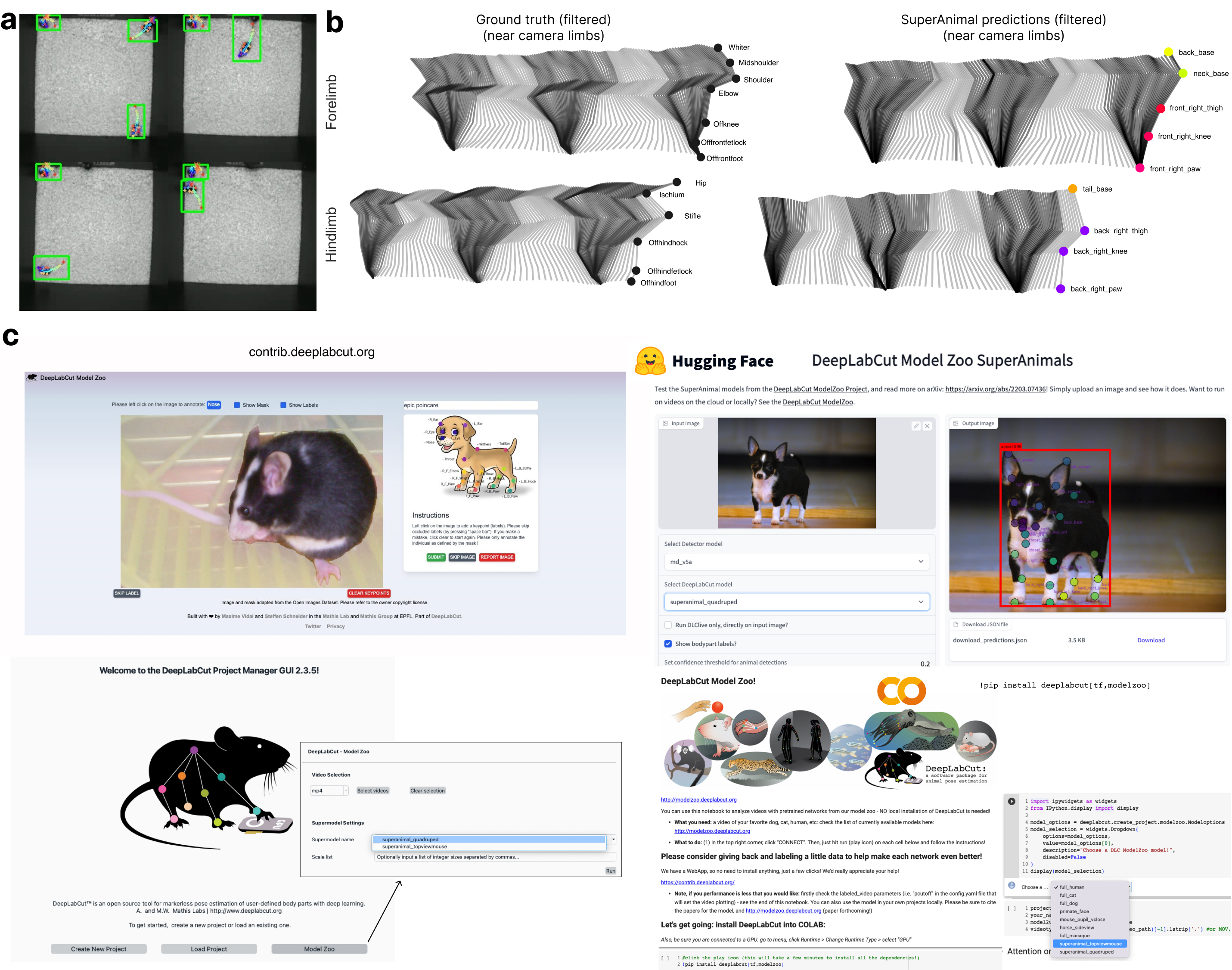}
\caption{
\textbf{a}: Visualization of top-down SuperAnimal-TopviewMouse on example MABe video frames, without trained on MABe videos.
\textbf{b} Same as Figure \ref{fig:action_segmentation}g, but smoothed with a 3-Hz zero-lag, low-pass, 2nd order Butterworth filter.
\textbf{c}: Top Left: An example of the current WebApp interface at \url{contrib.deeplabcut.org}. Users can add and edit the annotations from images we collect, following an anatomical figure that aids the expected location of bodyparts. Top Right: Example of current Gradio App on HuggingFace. Bottom Left: our current stand-alone GUI for local computer use showing a simple ModelZoo with SuperAnimal weights. Bottom Right: example of the Google Colaboratory interface with ModelZoo inference with SuperAnimal weights.
}
\label{fig:Extended_data_behavior}
\end{figure*}

\clearpage 
\newpage

\section*{\huge Supplementary Information}

\subsection*{Model Cards}

We provide Model Cards for the two major outputs, SA-TVM and SA-Q. These are also available on HuggingFace with the model weights, at \url{https://huggingface.co/mwmathis/DeepLabCutModelZoo-SuperAnimal-Quadruped} and \url{https://huggingface.co/mwmathis/DeepLabCutModelZoo-SuperAnimal-TopViewMouse}

\tcbset{colback=white!10!white}
\begin{tcolorbox}[title=\textbf{Model Card:  SuperAnimal-TopViewMouse (DLCRNet backbone and/or HRNet-w32)},
    breakable, sharp corners, boxrule=0.7pt]

\small{
\begin{mcsection}{Model Details}
    \item SuperAnimal-TopviewMouse model developed by the Mathis Lab in 2023, and trained to predict mouse 27 key points from a given top view image.
    \item DLCRNet \cite{Lauer2022multi} or HRNet-w32 was trained on the TopviewMouse-5K dataset.
     \item Models were trained within the \href{https://github.com/DeepLabCut}{DeepLabCut} framework or mmpose (HRNet-w32). You can use this model simply with our light-weight loading package called \href{https://github.com/DeepLabCut/DLClibrary}{DLCLibrary}.
     Full training details can be found in Ye et al. 2023. 
     Here is an example usage:
\begin{lstlisting}[language=Python]
from pathlib import Path
from dlclibrary import download_huggingface_model
# Creates a folder and downloads the model to it
model_dir = Path("./superanimal-topviewmouse_model_dlcrnet")
model_dir.mkdir()
download_huggingface_model("superanimal_topviewmouse_dlcrnet", model_dir)
\end{lstlisting}
\end{mcsection}

\begin{mcsection}{Intended Use}
    \item Intended to be used for pose tracking of lab mice videos filmed from an overhead view. The models can be used as a plug-and-play solution if extremely high precision is not required (we benchmark the zero-shot performance in the paper). Otherwise, it is recommended to also be used as the weights for transfer learning and fine-tuning.
        
    \item Intended for academic and research professionals working in fields related to animal behavior, neuroscience, biomechanics, and ecology.

    \item Not suitable for other species and other camera views. Also not suitable for videos that look dramatically different from those we show in the paper.
\end{mcsection}

\begin{mcsection}{Factors}

    \item Based on the known robustness issues of neural networks, the relevant factors include the lighting, contrast and resolution of the video frames. The presence of objects might also cause false detections of the mice and keypoints. When two or more animals are very close, it could cause the top-down detectors to only detect one animal, if used without further fine-tuning.
\end{mcsection}

\begin{mcsection}{Metrics}
     \item Mean Average Precision (mAP)
     \item Root Mean Square Error (RMSE)
\end{mcsection}

\begin{mcsection}{Evaluation Data}
    \item The test split of TopViewMouse-5K and in the paper on two benchmarks, DLC Openfield and TriMouse.

\end{mcsection}

\begin{mcsection}{Training Data}
\item \textbf{3CSI, BM, EPM, LDB, OFT} See full details at ~\cite{sturman2020deep} and~\cite{lukas2020mouse}.
\item \textbf{BlackMice} See full details at~\cite{isaac_chang_2020_3955216}.

\item \textbf{WhiteMice} Courtesy of Prof. Sam Golden and Nastacia Goodwin. See details in SIMBA~\cite{Nilsson2020.04.19.049452}.

\item \textbf{TriMouse} See full details at ~\cite{Lauer2022multi}..
\item \textbf{DLC-Openfield} See full details at ~\cite{mathis2018deeplabcut}.

\item \textbf{Kiehn-Lab-Openfield, Swimming, and Treadmill} Courtesy of Prof. Ole Kiehn, Dr. Jared Cregg, and Prof. Carmelo Bellardita; see details at ~\cite{Cregg2020BrainstemNT}.

\item \textbf{MausHaus} We collected video data from five single-housed C57BL/6J male and female mice in an extended home cage, carried out in the laboratory of Mackenzie Mathis at Harvard University and also EPFL (temperature of housing was 20-25C, humidity $20$-$50\%$). Data were recorded at 30Hz with $640$ × $480$ pixels resolution acquired with White Matter, LLC eV cameras. Annotators localized $26$ keypoints across 322 frames sampled from within DeepLabCut using the k-means clustering approach ~\cite{nath2019deeplabcut}. All experimental procedures for mice were in accordance with the National Institutes of Health Guide for the Care and Use of Laboratory Animals and approved by the Harvard Institutional Animal Care and Use Committee (IACUC) (n=$1$ mouse), and by the Veterinary Office of the Canton of Geneva (Switzerland; license GE01) (n=$4$ mice).
\end{mcsection}

%\pagebreak

\begin{mcsection}{Ethical Considerations}
\item Data was collected with IUCAC or other governmental approval. Each individual dataset used in training reports the ethics approval they obtained.
\end{mcsection}

\begin{mcsection}{Caveats and Recommendations}
\item The model may have reduced accuracy in scenarios with extremely varied lighting conditions or atypical mouse characteristics not well-represented in the training data. For example, this dataset only has one set of white mice, therefore it may not generalize well to diverse settings of white lab mice.

\item Please note that each training dataset was labeled by separate labs and different individuals, therefore while we map names to a unified pose vocabulary, there will be annotator bias in keypoint placement (See Ye et al. 2023 for our Supplementary Note on annotator bias). 
\item Note the dataset is primarily using C56Blk6/J mice and only some CD1 examples. 

\item We recommend if performance is not as good as you need it to be, first try video adaptation (see Ye et al. 2023), or fine-tune these weights with your own labeling.

\end{mcsection}

\begin{mcsection}{License}
    \item This software may not be used to harm any animal deliberately. Released under a modified MIT license. Please see details at \url{https://huggingface.co/mwmathis/DeepLabCutModelZoo-SuperAnimal-TopViewMouse}.
\end{mcsection}

\begin{mcsection}{Quantitative Analyses}
    \item See details at in Figure ~\ref{fig:Fig1overviewASTVM}.
\end{mcsection}
} 
\end{tcolorbox}

\tcbset{colback=white!10!white}
\begin{tcolorbox}[title=\textbf{Model Card:  SuperAnimal-Quadruped (HRNetw32)},
    breakable, sharp corners, boxrule=0.7pt]

\begin{mcsection}{Model Details}
    \item SuperAnimal-Quadruped model developed by the Mathis Lab in 2023, trained to predict quadruped pose from images.
    \item The main backbone model is an HRNet-w32 \cite{wang2020hrnet} trained on our Quadruped-80K dataset.
    \item We also release a top-down detector trained on the same data with Faster R-CNN~\cite{ren2015faster}.
\item Full training details can be found in Ye et al. 2023.
     You can use this model simply with our light-weight loading package called \href{https://github.com/DeepLabCut/DLClibrary}{DLCLibrary}.
     Here is an example usage:

\begin{lstlisting}[language=Python]
 from pathlib import Path
 from dlclibrary import download_huggingface_model
 # Creates a folder and downloads the model to it
 model_dir = Path("./superanimal_quadruped_hrnetw32")
 model_dir.mkdir()
 download_huggingface_model("superanimal_hrnetw32", model_dir)
 \end{lstlisting}

\end{mcsection}

\begin{mcsection}{Intended Use}
    \item Intended to be used for pose estimation of quadruped images taken from side-view. The model serves a better starting point than ImageNet weights in downstream datasets such as AP-10K.
        
    \item Intended for academic and research professionals working in fields related to animal behavior, such as neuroscience and ecology.

    \item Not suitable as a zeros-shot model for applications that require high keypiont precision, but can be fine-tuned with minimal data to reach human-level accuracy. Also not suitable for videos that look dramatically different from those we show in the paper.
\end{mcsection}

\begin{mcsection}{Factors}
    \item Based on the known robustness issues of neural networks, the relevant factors include the lighting, contrast and resolution of the video frames. The present of objects might also cause false detections and erroneous keypoints. When two or more animals are extremely close, it could cause the top-down detectors to only detect only one animal, if used without further fine-tuning or with a method such as BUCTD~\cite{zhou2023iccv}.
\end{mcsection}

\begin{mcsection}{Metrics}
     \item Mean Average Precision (mAP)
     \item Root Mean Square Error (RMSE)
     \item Normalized Error (NE)
\end{mcsection}

\begin{mcsection}{Evaluation Data}
    \item In the paper we benchmark on AP-10K, AnimalPose, Horse-10, and iRodent using a leave-one-out strategy. Here, we provide the model that has been trained on all datasets (see below), therefore it should be considered ``fine-tuned" on all animal training data listed below. This model is meant for production and evaluation in downstream scientific applications. 
    
\end{mcsection}

\begin{mcsection}{Training Data} 

\item \textbf{AwA-Pose} Quadruped dataset, see full details at~\cite{Banik2021AND}.
\item \textbf{AnimalPose} See full details at~\cite{Cao2019CrossDomainAF}.
\item \textbf{AcinoSet} See full details at~\cite{Joska2021AcinoSetA3}.
\item \textbf{Horse-30} Horse-30 dataset, benchmark task is called Horse-10; See full details at~\cite{mathis2021pretraining}.
\item \textbf{StanfordDogs} See full details at~\cite{KhoslaYaoJayadevaprakashFeiFei_FGVC2011, biggs2018creatures}.
\item \textbf{AP-10K} See full details at~\cite{yu2021ap}.
\item \textbf{APT-36K} See full details at ~\cite{yang2022apt}
\item \textbf{iRodent} We utilized the iNaturalist API functions for scraping observations with the taxon ID of Suborder Myomorpha~\cite{iRodent}. The functions allowed us to filter the large amount of observations down to the ones with photos under the CC BY-NC creative license. The most common types of rodents from the collected observations are Muskrat (Ondatra zibethicus), Brown Rat (Rattus norvegicus), House Mouse (Mus musculus), Black Rat (Rattus rattus), Hispid Cotton Rat (Sigmodon hispidus), Meadow Vole (Microtus pennsylvanicus), Bank Vole (Clethrionomys glareolus), Deer Mouse (Peromyscus maniculatus), White-footed Mouse (Peromyscus leucopus), Striped Field Mouse (Apodemus agrarius). We then generated segmentation masks over target animals in the data by processing the media through an algorithm we designed that uses a Mask Region Based Convolutional Neural Networks(Mask R-CNN)~\cite{he2017mask} model with a ResNet-50-FPN backbone \cite{https://doi.org/10.48550/arxiv.1612.03144}, pretrained on the COCO datasets~\cite{cocodataset}. The processed $443$ images were then manually labeled with pose annotations, and bounding boxes were generated by running Mega Detector~\cite{mega_detector} on the images, which were then manually verified. iRodent data is banked at \url{https://zenodo.org/record/8250392}.

\textbf{An image with the keypoint guide can be found in Extended Data Figure~\ref{fig:Extended_data_fig1}. }

\end{mcsection}

\begin{mcsection}{Ethical Considerations}
\item No experimental data was collected for this model; all datasets used are cited.
\end{mcsection}

\begin{mcsection}{Caveats and Recommendations}
\item The model may have reduced accuracy in scenarios with extremely varied lighting conditions or atypical animal characteristics not well-represented in the training data.
\item Please note that each dataset was labeled by separate labs and separate individuals, therefore while we map names
to a unified pose vocabulary (found here: https://github.com/AdaptiveMotorControlLab/modelzoo-figures), there will be annotator bias in keypoint placement (See Ye et al. 2023 for our Supplementary Note on annotator bias). 
\item Note the dataset is highly diverse across species, but collectively has more representation of domesticated animals like dogs, cats, horses, and cattle. 
\item We recommend if performance is not as good as you need it to be, first try video adaptation (see Ye et al. 2023), or fine-tune these weights with your own labeling.
\end{mcsection}

\begin{mcsection}{License}
    \item This software may not be used to harm any animal deliberately. Released under a modified MIT license. Please see details at \url{https://huggingface.co/mwmathis/DeepLabCutModelZoo-SuperAnimal-Quadruped}.
\end{mcsection}

\begin{mcsection}{Quantitative Analyses}
\item  See details at in Figure~\ref{fig:FigSAQ}.
\end{mcsection}
\end{tcolorbox}

\newpage

%%%%% TOPVIEWMOUSE
\section*{\huge Datasheet: TopViewMouse-5K dataset}
\begin{multicols}{2}
\bigskip

%%%%%%%%%%%%%%%%%%%%%%%%%%%%%%%%%%%%%%%%%%
\dssectionheader{Motivation}

\dsquestionex{For what purpose was the dataset created?}{Was there a specific task in mind? Was there a specific gap that needed to be filled? Please provide a description.}

\dsanswer{
We collected publicly available datasets from the community and additionally contribute MausHaus dataset. The purpose is to provide the community a unified vocabulary dataset for training pose models, and to help the community reproduce our findings. This dataset is used to train
models with the SuperAnimal method for mouse top-view pose estimation. The dataset was created intentionally with that task in mind, focusing on covering diverse lab settings of mice.
}

\dsquestion{Who created this dataset (e.g., which team, research group) and on behalf of which entity (e.g., company, institution, organization)?}

\dsanswer{
 
The merged dataset was created by Shaokai Ye, Ph.D. student at The Mathis Lab of Adaptive Intelligence, EPFL and checked by all co-authors. The merged dataset includes the following: 
\begin{enumerate}
    \item 3CSI, BM, EPM, LDB, OFT datasets, from the lab of Prof. Johannes Bohacek; see details at~\cite{sturman2020deep} and~\cite{lukas2020mouse}.
    \item BlackMice, from the lab of Prof. Chang; see details at~\cite{isaac_chang_2020_3955216}.
    \item WhiteMice, courtesy of Prof. Sam Golden and Nastacia Goodwin; see details in SIMBA~\cite{Nilsson2020.04.19.049452}.
    \item TriMouse benchmark dataset, see details at~\cite{Lauer2022multi}.
    \item DLC-Openfield, see details at~\cite{mathis2018deeplabcut}.
    \item Kiehn-Lab-Openfield, Swimming, and Treadmill, courtesy of Prof. Ole Kiehn, Dr. Jared Cregg, and Prof. Carmelo Bellardita; see details at~\cite{Cregg2020BrainstemNT}.
    \item MausHaus dataset, collected in the lab of Prof. Mackenzie Mathis at Harvard University and EPFL.
\end{enumerate}

}
\dsquestionex{Who funded the creation of the dataset?}{If there is an associated grant, please provide the name of the grantor and the grant name and number.}

\dsanswer{
Each individual paper denotes the funding for the work, therefore check the references. For the newly created MausHaus data, it was funded by start-up funds to Prof. Mackenzie Mathis at the Rowland Institute of Harvard and at EPFL.
}

\dsquestion{Any other comments?}

\dsanswer{
None.
}

%%%%%%%%%%%%%%%%%%%%%%%%%%%%%%%%%%%%%%%%%%
\bigskip
\dssectionheader{Composition}

\dsquestionex{What do the instances that comprise the dataset represent (e.g., documents, photos, people, countries)?}{ Are there multiple types of instances (e.g., movies, users, and ratings; people and interactions between them; nodes and edges)? Please provide a description.}

\dsanswer{The instances are images of mice extracted from the top-view video coupled with the human annotated keypoints. Videos have different resolutions, number of animals per frame, number of annotated keypoints as well as frame frequencies. To our best knowledge, frames were only annotated once per instance.
}

\dsquestion{How many instances are there in total (of each type, if appropriate)?}

\dsanswer{
The merged dataset consists of approximately 5,000 frames. For more information see Extended Data Figure 1.
}

\dsquestionex{Does the dataset contain all possible instances or is it a sample (not necessarily random) of instances from a larger set?}{ If the dataset is a sample, then what is the larger set? Is the sample representative of the larger set (e.g., geographic coverage)? If so, please describe how this representativeness was validated/verified. If it is not representative of the larger set, please describe why not (e.g., to cover a more diverse range of instances, because instances were withheld or unavailable).}

\dsanswer{
The merged dataset contains all possible instances from each individual source. For MausHaus, the frames were extracted from multiple different mice and videos using kmeans clustering then labeled within the DeepLabCut software package (versions 2.0.7-2.2 were used).
}

\dsquestionex{What data does each instance consist of? “Raw” data (e.g., unprocessed text or images) or features?}{In either case, please provide a description.}

\dsanswer{
Each instance in the dataset comprises a top-view image featuring one or more mice. Accompanying these images are human annotated keypoints for each individual mouse, which detail specific points of interest or markers on the animal’s body. These keypoints provide valuable information for pose estimation and behavioral analysis.
}

\dsquestionex{Is there a label or target associated with each instance?}{If so, please provide a description.}

\dsanswer{
The labels are the 2D coordinates (x, y in pixel space) and visibility flag (unlabeled if occluded) per each keypoint for each dataset.
}

\dsquestionex{Is any information missing from individual instances?}{If so, please provide a description, explaining why this information is missing (e.g., because it was unavailable). This does not include intentionally removed information, but might include, e.g., redacted text.}

\dsanswer{
Unknown to the authors of the merged dataset.
}

\dsquestionex{Are relationships between individual instances made explicit (e.g., users’ movie ratings, social network links)?}{If so, please describe how these relationships are made explicit.}

\dsanswer{
In the dataset of frames extracted from top-view videos of mice, the relationships between individual instances (frames) are not explicitly defined in terms of behavioral interactions or social links. Instead, the dataset primarily focuses on isolated frames as individual instances. Any temporal or behavioral relationships between the frames would be implicit, derived from the sequence in which they appear in the videos. 
}

\dsquestionex{Are there recommended data splits (e.g., training, development/validation, testing)?}{If so, please provide a description of these splits, explaining the rationale behind them.}

\dsanswer{
The dataset is partitioned into a train-test split with a ratio of 95:5. This distribution is established to rigorously evaluate the model's efficacy on a set of data distinct from those used during its training phase.
}

\dsquestionex{Are there any errors, sources of noise, or redundancies in the dataset?}{If so, please provide a description.}

\dsanswer{
There are two primary sources of error in our dataset: firstly, annotation errors from the annotators of individual datasets may exist - we did not correct any original data source; and secondly, imperfections in the projection of keypoints from the original keypoint space to the target keypoint space cannot be guaranteed to not have occured, although the authors did their best efforts to avoid such errors. Please see the pre-processing Methods section for more details and for the conversion table that the authors created.
}

\dsquestionex{Is the dataset self-contained, or does it link to or otherwise rely on external resources (e.g., websites, tweets, other datasets)?}{If it links to or relies on external resources, a) are there guarantees that they will exist, and remain constant, over time; b) are there official archival versions of the complete dataset (i.e., including the external resources as they existed at the time the dataset was created); c) are there any restrictions (e.g., licenses, fees) associated with any of the external resources that might apply to a future user? Please provide descriptions of all external resources and any restrictions associated with them, as well as links or other access points, as appropriate.}

\dsanswer{
The merged single source dataset is self-contained and does not rely on external link that might change over time. Individual dataset links could be modified.
}

\dsquestionex{Does the dataset contain data that might be considered confidential (e.g., data that is protected by legal privilege or by doctor-patient confidentiality, data that includes the content of individuals non-public communications)?}{If so, please provide a description.}

\dsanswer{To our best knowledge, no such data is included, and all data was collected under ethics approval for animal research.
}

\dsquestionex{Does the dataset contain data that, if viewed directly, might be offensive, insulting, threatening, or might otherwise cause anxiety?}{If so, please describe why.}

\dsanswer{
Unknown to the authors of the datasheet, but the images are of uninjured animals in freely moving settings in laboratories, therefore we do not anticipate they cause alarm for humans.
}

\dsquestionex{Does the dataset relate to people?}{If not, you may skip the remaining questions in this section.}

\dsanswer{
No
}

\dsquestion{Any other comments?}

\dsanswer{
None.
}

%%%%%%%%%%%%%%%%%%%%%%%%%%%%%%%%%%%%%%%%%%
%\bigskip
\dssectionheader{Collection Process}

\dsquestionex{How was the data associated with each instance acquired?}{Was the data directly observable (e.g., raw text, movie ratings), reported by subjects (e.g., survey responses), or indirectly inferred/derived from other data (e.g., part-of-speech tags, model-based guesses for age or language)? If data was reported by subjects or indirectly inferred/derived from other data, was the data validated/verified? If so, please describe how.}

\dsanswer{
Individual datasets before merging were acquired from published papers or annotated by authors of the paper.

Datasets are validated and verified by the original dataset creators and later verified by authors of this paper.
}

\dsquestionex{What mechanisms or procedures were used to collect the data (e.g., hardware apparatus or sensor, manual human curation, software program, software API)?}{How were these mechanisms or procedures validated?}

\dsanswer{
For MausHaus dataset we collected video data from five single-housed C57BL/6J male and female mice in an extended home cage, carried out in the laboratory of Mackenzie Mathis at Harvard University and also EPFL. Data were recorded with White Matter, LLC eV cameras. Annotators localized 27 keypoints across 322 frames sampled from within DeepLabCut using the k-means clustering approach~\cite{nath2019deeplabcut}. All experimental procedures for mice were in accordance with the National Institutes of Health Guide for the Care and Use of Laboratory Animals and approved by the Harvard Institutional Animal Care and Use Committee (IACUC) (n=1 mouse), and  by the Veterinary Office of the Canton of Geneva (Switzerland; license GE01) (n=4 mice).
}

\dsquestion{If the dataset is a sample from a larger set, what was the sampling strategy (e.g., deterministic, probabilistic with specific sampling probabilities)?}

\dsanswer{
For publicly available data, please see their methods. For MausHaus, it was sampled via k-means clustering of videos.
}

\dsquestion{Who was involved in the data collection process (e.g., students, crowdworkers, contractors) and how were they compensated (e.g., how much were crowdworkers paid)?}

\dsanswer{
We do not have information on the publicly available datasets. MausHaus was annotated by Prof. Mackenzie Mathis as part of her employment at either Harvard University or EPFL.
}

\dsquestionex{Over what time frame was the data collected? Does this time frame match the creation time frame of the data associated with the instances (e.g., recent crawl of old news articles)?}{If not, please describe the time frame in which the data associated with the instances was created.}

\dsanswer{
Data was collected from 2020-2023.
}

\dsquestionex{Were any ethical review processes conducted (e.g., by an institutional review board)?}{If so, please provide a description of these review processes, including the outcomes, as well as a link or other access point to any supporting documentation.}

\dsanswer{
Yes, every individual paper we sourced data from included a relevant ethical approval to collect data from mice.
}

\dsquestionex{Does the dataset relate to people?}{If not, you may skip the remaining questions in this section.}

\dsanswer{
No.
}

%%%%%%%%%%%%%%%%%%%%%%%%%%%%%%%%%%%%%%%%%%
%\bigskip
\dssectionheader{Preprocessing/cleaning/labeling}

\dsquestionex{Was any preprocessing/cleaning/labeling of the data done (e.g., discretization or bucketing, tokenization, part-of-speech tagging, SIFT feature extraction, removal of instances, processing of missing values)?}{If so, please provide a description. If not, you may skip the remainder of the questions in this section.}

\dsanswer{
Data came from multiple sources
and in multiple formats. To homogenize different annotation formats (COCO-style, DeepLabCut format, etc.), we implemented
a generalized data converter. We parsed public datasets and reformatted them into DeepLabCut projects. Besides data conversion, the generalized data converter also implements key steps for the panoptic animal pose estimation task formulation, but no individual keypoints were changed. 
}

\dsquestionex{Was the “raw” data saved in addition to the preprocessed/cleaned/labeled data (e.g., to support unanticipated future uses)?}{If so, please provide a link or other access point to the “raw” data.}

\dsanswer{
The raw data was used, and can be extracted from the corresponding original source.
}

\dsquestionex{Is the software used to preprocess/clean/label the instances available?}{If so, please provide a link or other access point.}
\dsanswer{
Yes, the conversion table is available at \url{https://github.com/AdaptiveMotorControlLab/modelzoo-figures}.
}

\dsquestion{Any other comments?}

\dsanswer{
None.
}

%%%%%%%%%%%%%%%%%%%%%%%%%%%%%%%%%%%%%%%%%%
%\bigskip
\dssectionheader{Uses}

\dsquestionex{Has the dataset been used for any tasks already?}{If so, please provide a description.}

\dsanswer{
At the time of publication, only the original paper has used the dataset.
}

\dsquestionex{Is there a repository that links to any or all papers or systems that use the dataset?}{If so, please provide a link or other access point.}

\dsanswer{
We suggest to check the citations of original paper sources.
}

\dsquestion{What (other) tasks could the dataset be used for?}

\dsanswer{
The dataset could be used for anything related to mouse pose estimation.
}

\dsquestionex{Is there anything about the composition of the dataset or the way it was collected and preprocessed/cleaned/labeled that might impact future uses?}{For example, is there anything that a future user might need to know to avoid uses that could result in unfair treatment of individuals or groups (e.g., stereotyping, quality of service issues) or other undesirable harms (e.g., financial harms, legal risks) If so, please provide a description. Is there anything a future user could do to mitigate these undesirable harms?}

\dsanswer{
Keypoint annotations from individual datasets were projected to a super-set keypoint space which represents this dataset. The model that is trained over this dataset might have bias on keypoints that are more common in the individual datasets and might have larger errors on keypoints that are not under-represented in the source datasets.
}

\dsquestionex{Are there tasks for which the dataset should not be used?}{If so, please provide a description.}

\dsanswer{
The dataset cannot be used to harm any animal.
The dataset should not be used to train a model that is expected to be directly used (i.e., without further fine-tuning) for applications that require extremely high-precision, as there were annotator bias from 
the source datasets.
}

\dsquestion{Any other comments?}

\dsanswer{None.}

%%%%%%%%%%%%%%%%%%%%%%%%%%%%%%%%%%%%%%%%%%
%\bigskip
\dssectionheader{Distribution}

\dsquestionex{Will the dataset be distributed to third parties outside of the entity (e.g., company, institution, organization) on behalf of which the dataset was created?}{If so, please provide a description.}

\dsanswer{
Yes, the dataset will be publicly available with a license for use.
}

\dsquestionex{How will the dataset will be distributed (e.g., tarball on website, API, GitHub)}{Does the dataset have a digital object identifier (DOI)?}

\dsanswer{
The dataset is distributed on zenodo (pending acceptance of this paper).
}

\dsquestion{When will the dataset be distributed?}

\dsanswer{
The merged dataset will be released pending acceptance of this paper).
}

\dsquestionex{Will the dataset be distributed under a copyright or other intellectual property (IP) license, and/or under applicable terms of use (ToU)?}{If so, please describe this license and/or ToU, and provide a link or other access point to, or otherwise reproduce, any relevant licensing terms or ToU, as well as any fees associated with these restrictions.}

\dsanswer{
The data copyright belongs to the authors of the original datasets. 
}

\dsquestionex{Have any third parties imposed IP-based or other restrictions on the data associated with the instances?}{If so, please describe these restrictions, and provide a link or other access point to, or otherwise reproduce, any relevant licensing terms, as well as any fees associated with these restrictions.}

\dsanswer{
Yes, please check the original sources. We assume no liability or guarantees on this model's use.
}

\dsquestionex{Do any export controls or other regulatory restrictions apply to the dataset or to individual instances?}{If so, please describe these restrictions, and provide a link or other access point to, or otherwise reproduce, any supporting documentation.}

\dsanswer{
Unknown.
}

\dsquestion{Any other comments?}

\dsanswer{
None.
}

%%%%%%%%%%%%%%%%%%%%%%%%%%%%%%%%%%%%%%%%%%
\dssectionheader{Maintenance}

\dsquestion{Who will be supporting/hosting/maintaining the dataset?}

\dsanswer{
The dataset will be hosted on zenodo.
}

\dsquestion{How can the owner/curator/manager of the dataset be contacted (e.g., email address)?}

\dsanswer{
The head of The Mathis Lab of Adaptive Intelligence, Mackenzie Mathis, can be contacted at \href{mackenzie.mathis@epfl.ch}{mackenzie.mathis@epfl.ch}.
}

\dsquestionex{Is there an erratum?}{If so, please provide a link or other access point.}

\dsanswer{
None.
}

\dsquestionex{Will the dataset be updated (e.g., to correct labeling errors, add new instances, delete instances)?}{If so, please describe how often, by whom, and how updates will be communicated to users (e.g., mailing list, GitHub)?}

\dsanswer{
Not at this time.
}

\dsquestionex{If the dataset relates to people, are there applicable limits on the retention of the data associated with the instances (e.g., were individuals in question told that their data would be retained for a fixed period of time and then deleted)?}{If so, please describe these limits and explain how they will be enforced.}

\dsanswer{
No.
}

\dsquestionex{Will older versions of the dataset continue to be supported/hosted/maintained?}{If so, please describe how. If not, please describe how its obsolescence will be communicated to users.}

\dsanswer{
The data will be banked with zenodo.
}

\dsquestionex{If others want to extend/augment/build on/contribute to the dataset, is there a mechanism for them to do so?}{If so, please provide a description. Will these contributions be validated/verified? If so, please describe how. If not, why not? Is there a process for communicating/distributing these contributions to other users? If so, please provide a description.}

\dsanswer{
Others may do so and should contact the original authors about incorporating fixes/extensions.
}

\dsquestion{Any other comments?}

\dsanswer{None.
}

\end{multicols}

\newpage

%%%%% QUAD 
\section*{\huge Datasheet: Quadruped-80K dataset}
\begin{multicols}{2}
%%%%%%%%%%%%%%%%%%%%%%%%%%%%%%%%%%%%%%%%%%
\dssectionheader{Motivation}

\dsquestionex{For what purpose was the dataset created?}{Was there a specific task in mind? Was there a specific gap that needed to be filled? Please provide a description.}

\dsanswer{
We collected publicly available datasets from the community and additionally contribute iRodent dataset. The purpose is to provide the community a unified vocabulary dataset for training pose models, and to help the community reproduce our findings. This dataset is used to train
models with the SuperAnimal method for quadruped pose estimation. The dataset was created intentionally with that task in mind, focusing on covering animals in the wild.
}

\dsquestion{Who created this dataset (e.g., which team, research group) and on behalf of which entity (e.g., company, institution, organization)?}

\dsanswer{
 
The merged dataset was created by Shaokai Ye, Ph.D. student at The Mathis Lab of Adaptive Intelligence, EPFL and checked by all co-authors. The merged dataset includes the following: 
\begin{enumerate}
    \item \textbf{AwA-Pose} Quadruped dataset, see full details at~\cite{Banik2021AND}.
    \item \textbf{AnimalPose} See full details at~\cite{Cao2019CrossDomainAF}.
    \item \textbf{AcinoSet} See full details at~\cite{Joska2021AcinoSetA3}.
    \item \textbf{Horse-30} Horse-30 dataset, benchmark task is called Horse-10; See full details at~\cite{mathis2021pretraining}.
    \item \textbf{StanfordDogs} See full details at~\cite{KhoslaYaoJayadevaprakashFeiFei_FGVC2011, biggs2018creatures}.
    \item \textbf{AP-10K} See full details at~\cite{yu2021ap}.
    \item \textbf{APT-36K} See full details at ~\cite{yang2022apt}
    \item \textbf{iRodent} We utilized the iNaturalist API functions for scraping observations with the taxon ID of Suborder Myomorpha~\cite{iRodent}. The functions allowed us to filter the large amount of observations down to the ones with photos under the CC BY-NC creative license. The most common types of rodents from the collected observations are Muskrat (Ondatra zibethicus), Brown Rat (Rattus norvegicus), House Mouse (Mus musculus), Black Rat (Rattus rattus), Hispid Cotton Rat (Sigmodon hispidus), Meadow Vole (Microtus pennsylvanicus), Bank Vole (Clethrionomys glareolus), Deer Mouse (Peromyscus maniculatus), White-footed Mouse (Peromyscus leucopus), Striped Field Mouse (Apodemus agrarius). We then generated segmentation masks over target animals in the data by processing the media through an algorithm we designed that uses a Mask Region Based Convolutional Neural Networks(Mask R-CNN)~\cite{he2017mask} model with a ResNet-50-FPN backbone \cite{https://doi.org/10.48550/arxiv.1612.03144}, pretrained on the COCO datasets~\cite{cocodataset}. The processed 443 images were then manually labeled with pose annotations, and bounding boxes were generated by running Mega Detector~\cite{mega_detector} on the images, which were then manually verified. iRodent data is banked at \url{https://zenodo.org/record/8250392}.
\end{enumerate}

}
\dsquestionex{Who funded the creation of the dataset?}{If there is an associated grant, please provide the name of the grantor and the grant name and number.}

\dsanswer{
Each individual paper denotes the funding for the work, therefore check the references. For the newly created iRodent data, it was funded by start-up funds to Prof. Mackenzie Mathis at EPFL.
}

\dsquestion{Any other comments?}

\dsanswer{
None.
}

%%%%%%%%%%%%%%%%%%%%%%%%%%%%%%%%%%%%%%%%%%
\bigskip
\dssectionheader{Composition}

\dsquestionex{What do the instances that comprise the dataset represent (e.g., documents, photos, people, countries)?}{ Are there multiple types of instances (e.g., movies, users, and ratings; people and interactions between them; nodes and edges)? Please provide a description.}

\dsanswer{The instances are images of animals extracted from the side-view images coupled with the human annotated keypoints. Videos/images have different resolutions, number of animals per frame, number of annotated keypoints as well as frame frequencies. To our best knowledge, frames were only annotated once per instance.
}

\dsquestion{How many instances are there in total (of each type, if appropriate)?}

\dsanswer{
The merged dataset consists of approximately 85,000 frames. For more information see Extended Data Figure 1.
}

\dsquestionex{Does the dataset contain all possible instances or is it a sample (not necessarily random) of instances from a larger set?}{ If the dataset is a sample, then what is the larger set? Is the sample representative of the larger set (e.g., geographic coverage)? If so, please describe how this representativeness was validated/verified. If it is not representative of the larger set, please describe why not (e.g., to cover a more diverse range of instances, because instances were withheld or unavailable).}

\dsanswer{
The merged dataset contains all possible instances from each individual source.
}

\dsquestionex{What data does each instance consist of? “Raw” data (e.g., unprocessed text or images) or features?}{In either case, please provide a description.}

\dsanswer{
Each instance in the dataset comprises a side-view image featuring one or more animals. Accompanying these images are human annotated keypoints for each individual, which detail specific points of interest or markers on the animal’s body. These keypoints provide valuable information for pose estimation and behavioral analysis.
}

\dsquestionex{Is there a label or target associated with each instance?}{If so, please provide a description.}

\dsanswer{
The labels are the 2D coordinates (x, y in pixel space) and visibility flag (unlabeled if occluded) per each keypoint for each dataset.
}

\dsquestionex{Is any information missing from individual instances?}{If so, please provide a description, explaining why this information is missing (e.g., because it was unavailable). This does not include intentionally removed information, but might include, e.g., redacted text.}

\dsanswer{
Unknown to the authors of the merged dataset.
}

\dsquestionex{Are relationships between individual instances made explicit (e.g., users’ movie ratings, social network links)?}{If so, please describe how these relationships are made explicit.}

\dsanswer{
In the dataset of pictures or frames extracted from side-view videos of animals, the relationships between individual instances (frames) are not explicitly defined in terms of behavioral interactions or social links. Instead, the dataset primarily focuses on isolated frames as individual instances. Any temporal or behavioral relationships between the frames would be implicit, derived from the sequence in which they appear in the videos. 
}

\dsquestionex{Are there recommended data splits (e.g., training, development/validation, testing)?}{If so, please provide a description of these splits, explaining the rationale behind them.}

\dsanswer{
The dataset is partitioned into a train set, where individual datasets can be dropped to test OOD performance.
}

\dsquestionex{Are there any errors, sources of noise, or redundancies in the dataset?}{If so, please provide a description.}

\dsanswer{
There are two primary sources of error in our dataset: firstly, annotation errors from the annotators of individual datasets may exist - we did not correct any original data source; and secondly, imperfections in the projection of keypoints from the original keypoint space to the target keypoint space cannot be guaranteed to not have occurred, although the authors did their best efforts to avoid such errors. Please see the pre-processing Methods section for more details and for the conversion table that the authors created.
}

\dsquestionex{Is the dataset self-contained, or does it link to or otherwise rely on external resources (e.g., websites, tweets, other datasets)?}{If it links to or relies on external resources, a) are there guarantees that they will exist, and remain constant, over time; b) are there official archival versions of the complete dataset (i.e., including the external resources as they existed at the time the dataset was created); c) are there any restrictions (e.g., licenses, fees) associated with any of the external resources that might apply to a future user? Please provide descriptions of all external resources and any restrictions associated with them, as well as links or other access points, as appropriate.}

\dsanswer{
The merged single source dataset is self-contained and does not rely on external link that might change over time. Individual dataset links could be modified.
}

\dsquestionex{Does the dataset contain data that might be considered confidential (e.g., data that is protected by legal privilege or by doctor-patient confidentiality, data that includes the content of individuals non-public communications)?}{If so, please provide a description.}

\dsanswer{To our best knowledge, no such data is included.
}

\dsquestionex{Does the dataset contain data that, if viewed directly, might be offensive, insulting, threatening, or might otherwise cause anxiety?}{If so, please describe why.}

\dsanswer{
Some images from iNaturalist contain dead rodents that could cause anxiety.
}

\dsquestionex{Does the dataset relate to people?}{If not, you may skip the remaining questions in this section.}

\dsanswer{
No
}

\dsquestion{Any other comments?}

\dsanswer{
None.
}

%%%%%%%%%%%%%%%%%%%%%%%%%%%%%%%%%%%%%%%%%%
%\bigskip
\dssectionheader{Collection Process}

\dsquestionex{How was the data associated with each instance acquired?}{Was the data directly observable (e.g., raw text, movie ratings), reported by subjects (e.g., survey responses), or indirectly inferred/derived from other data (e.g., part-of-speech tags, model-based guesses for age or language)? If data was reported by subjects or indirectly inferred/derived from other data, was the data validated/verified? If so, please describe how.}

\dsanswer{
Individual datasets before merging were acquired from published papers or annotated by authors of the paper.

Datasets are validated and verified by the original dataset creators and later verified by authors of this paper.
}

\dsquestionex{What mechanisms or procedures were used to collect the data (e.g., hardware apparatus or sensor, manual human curation, software program, software API)?}{How were these mechanisms or procedures validated?}

\dsanswer{
No new data was collected for this merged dataset.
}

\dsquestion{If the dataset is a sample from a larger set, what was the sampling strategy (e.g., deterministic, probabilistic with specific sampling probabilities)?}

\dsanswer{
For publicly available data, please see their methods.
}

\dsquestion{Who was involved in the data collection process (e.g., students, crowdworkers, contractors) and how were they compensated (e.g., how much were crowdworkers paid)?}

\dsanswer{
We do not have information on the publicly available datasets. iRodent was annotated by Prof. Mackenzie Mathis and Tian Qiu at Harvard University and/or EPFL.
}

\dsquestionex{Over what time frame was the data collected? Does this time frame match the creation time frame of the data associated with the instances (e.g., recent crawl of old news articles)?}{If not, please describe the time frame in which the data associated with the instances was created.}

\dsanswer{
Data was collected from 2020-2023.
}

\dsquestionex{Were any ethical review processes conducted (e.g., by an institutional review board)?}{If so, please provide a description of these review processes, including the outcomes, as well as a link or other access point to any supporting documentation.}

\dsanswer{
Unknown.
}

\dsquestionex{Does the dataset relate to people?}{If not, you may skip the remaining questions in this section.}

\dsanswer{
No.
}

%%%%%%%%%%%%%%%%%%%%%%%%%%%%%%%%%%%%%%%%%%
\bigskip
\dssectionheader{Preprocessing/cleaning/labeling}

\dsquestionex{Was any preprocessing/cleaning/labeling of the data done (e.g., discretization or bucketing, tokenization, part-of-speech tagging, SIFT feature extraction, removal of instances, processing of missing values)?}{If so, please provide a description. If not, you may skip the remainder of the questions in this section.}

\dsanswer{
Data came from multiple sources
and in multiple formats. To homogenize different annotation formats (COCO-style, DeepLabCut format, etc.), we implemented
a generalized data converter. We parsed public datasets and reformatted them into DeepLabCut projects. Besides data conversion, the generalized data converter also implements key steps for the panoptic animal pose estimation task formulation, but no individual keypoints were changed. 
}

\dsquestionex{Was the “raw” data saved in addition to the preprocessed/cleaned/labeled data (e.g., to support unanticipated future uses)?}{If so, please provide a link or other access point to the “raw” data.}

\dsanswer{
The raw data was used, and can be extracted from the corresponding original source.
}

\dsquestionex{Is the software used to preprocess/clean/label the instances available?}{If so, please provide a link or other access point.}
\dsanswer{
Yes, the conversion table is available at \url{https://github.com/AdaptiveMotorControlLab/modelzoo-figures}.
}

\dsquestion{Any other comments?}

\dsanswer{
None.
}

%%%%%%%%%%%%%%%%%%%%%%%%%%%%%%%%%%%%%%%%%%
\bigskip
\dssectionheader{Uses}

\dsquestionex{Has the dataset been used for any tasks already?}{If so, please provide a description.}

\dsanswer{
At the time of publication, only the original paper has used the dataset.
}

\dsquestionex{Is there a repository that links to any or all papers or systems that use the dataset?}{If so, please provide a link or other access point.}

\dsanswer{
We suggest to check the citations of original paper sources.
}

\dsquestion{What (other) tasks could the dataset be used for?}

\dsanswer{
The dataset could be used for anything related to animal pose estimation.
}

\dsquestionex{Is there anything about the composition of the dataset or the way it was collected and preprocessed/cleaned/labeled that might impact future uses?}{For example, is there anything that a future user might need to know to avoid uses that could result in unfair treatment of individuals or groups (e.g., stereotyping, quality of service issues) or other undesirable harms (e.g., financial harms, legal risks) If so, please provide a description. Is there anything a future user could do to mitigate these undesirable harms?}

\dsanswer{
Keypoint annotations from individual datasets were projected to a super-set keypoint space which represents this dataset. The model that is trained over this dataset might have bias on keypoints that are more common in the individual datasets and might have larger errors on keypoints that are not under-represented in the source datasets.
}

\dsquestionex{Are there tasks for which the dataset should not be used?}{If so, please provide a description.}

\dsanswer{
The dataset cannot be used to harm any animal.
The dataset should not be used to train a model that is expected to be directly used (i.e., without further fine-tuning) for applications that require extremely high-precision, as there were annotator bias from 
the source datasets.
}

\dsquestion{Any other comments?}

\dsanswer{None.}

%%%%%%%%%%%%%%%%%%%%%%%%%%%%%%%%%%%%%%%%%%
\bigskip
\dssectionheader{Distribution}

\dsquestionex{Will the dataset be distributed to third parties outside of the entity (e.g., company, institution, organization) on behalf of which the dataset was created?}{If so, please provide a description.}

\dsanswer{
Yes, the dataset will be publicly available with a license for use.
}

\dsquestionex{How will the dataset will be distributed (e.g., tarball on website, API, GitHub)}{Does the dataset have a digital object identifier (DOI)?}

\dsanswer{
The dataset is distributed on zenodo (pending acceptance of this paper).
}

\dsquestion{When will the dataset be distributed?}

\dsanswer{
The merged dataset will be released pending acceptance of this paper).
}

\dsquestionex{Will the dataset be distributed under a copyright or other intellectual property (IP) license, and/or under applicable terms of use (ToU)?}{If so, please describe this license and/or ToU, and provide a link or other access point to, or otherwise reproduce, any relevant licensing terms or ToU, as well as any fees associated with these restrictions.}

\dsanswer{
The data copyright belongs to the authors of the original datasets. Horse-30 is non-commercial user only.
}

\dsquestionex{Have any third parties imposed IP-based or other restrictions on the data associated with the instances?}{If so, please describe these restrictions, and provide a link or other access point to, or otherwise reproduce, any relevant licensing terms, as well as any fees associated with these restrictions.}

\dsanswer{
Yes, please check the original sources. We assume no liability or guarantees on this model's use.
}

\dsquestionex{Do any export controls or other regulatory restrictions apply to the dataset or to individual instances?}{If so, please describe these restrictions, and provide a link or other access point to, or otherwise reproduce, any supporting documentation.}

\dsanswer{
Unknown.
}

\dsquestion{Any other comments?}

\dsanswer{
None.
}

%%%%%%%%%%%%%%%%%%%%%%%%%%%%%%%%%%%%%%%%%%
\bigskip
\dssectionheader{Maintenance}

\dsquestion{Who will be supporting/hosting/maintaining the dataset?}

\dsanswer{
The dataset will be hosted on zenodo.
}

\dsquestion{How can the owner/curator/manager of the dataset be contacted (e.g., email address)?}

\dsanswer{
The head of The Mathis Lab of Adaptive Intelligence, Mackenzie Mathis, can be contacted at \href{mackenzie.mathis@epfl.ch}{mackenzie.mathis@epfl.ch}.
}

\dsquestionex{Is there an erratum?}{If so, please provide a link or other access point.}

\dsanswer{
None.
}

\dsquestionex{Will the dataset be updated (e.g., to correct labeling errors, add new instances, delete instances)?}{If so, please describe how often, by whom, and how updates will be communicated to users (e.g., mailing list, GitHub)?}

\dsanswer{
Not at this time.
}

\dsquestionex{If the dataset relates to people, are there applicable limits on the retention of the data associated with the instances (e.g., were individuals in question told that their data would be retained for a fixed period of time and then deleted)?}{If so, please describe these limits and explain how they will be enforced.}

\dsanswer{
No.
}

\dsquestionex{Will older versions of the dataset continue to be supported/hosted/maintained?}{If so, please describe how. If not, please describe how its obsolescence will be communicated to users.}

\dsanswer{
The data will be banked with zenodo.
}

\dsquestionex{If others want to extend/augment/build on/contribute to the dataset, is there a mechanism for them to do so?}{If so, please provide a description. Will these contributions be validated/verified? If so, please describe how. If not, why not? Is there a process for communicating/distributing these contributions to other users? If so, please provide a description.}

\dsanswer{
Others may do so and should contact the original authors about incorporating fixes/extensions.
}

\dsquestion{Any other comments?}

\dsanswer{None.
}

%%%%%
%\newpage 
%\bibliography{refs}
\end{multicols}
%\end{singlespace}

%%%%%%%%%%%
\section*{Considerations on building general datasets for pretraining}

To build generalizable pose models, a large-scale pre-training dataset is the key. It has been shown in both computer vision and natural language processing that pre-trained models significantly improve the generalization of models and data efficiency in the downstream datasets~\cite{he2022masked,devlin2018bert}. However, data of lab animals are not ubiquitous on the internet. To get large scale animal pose data, it is critical to gather the data directly from the research community in a responsible and transparent way. A platform that actively interacts with the community is thus required to build such a pre-training dataset.
 As such a vocabulary is built on top of a wide range of pose datasets, it can be used across different research needs and it is also key to for useful zero-shot inference (see Methods).

We acknowledge that these SuperAnimal models would not have been be possible without the accumulated data from the community. 
In the future, feedback from the community for models' efficacy and failure modes (Extended Data Fig.~\ref{fig:Extended_data_matching_ATP}) in different downstream data will be critical for updated model releases and algorithmic updates. As publicly available data increase, we expect the performance will improve.

\subsection*{Annotator bias in labeled data}

Unlike previous works that require labeling data to create a working model, our models can be used as they are. For the purpose of evaluation, we could use the ground-truth of the target dataset or label frames of a novel video. We note when it comes to evaluating the performance of zero-shot inference, there will always be a systematic errors between the model and the annotator of the target dataset. We refer this type of error to be caused annotator bias, meaning annotators of different datasets try to place keypoints in slightly different places due to the bias of annotators. Therefore, the supervised metrics will tend to be an over-estimation of the error.

Reversely, SuperAnimal models can be used to monitor annotator bias as the model's predictions are consistent across frames while in many cases human annotators annotate keypoints in a inconsistent way.

\section*{Supervised metrics do not capture the richness of SuperAnimal-models}

In pose estimation literature, work mostly report supervised metrics (RMSE, Normalized Error, and mAP). What is shared in the supervised metrics is that the metrics do not penalize keypoints that are not annotated in the dataset. In contrast to other pose models, our SuperAnimal models can predict keypoints that are not annotated in the labeled dataset. For instance, if we apply only supervised metrics to evaluate SuperAnimal models, catastrophic forgetting is not detected as metrics do not penalize keypoint predictions that are not annotated.

\section*{Why we used a top-down approach for quadruped pre-trained pose models}

Compared to the COCO keypoint benchmark~\cite{cocodataset}, animal in the wild shows long tail distribution of subject sizes in both relative and absolute terms (Figure ~\ref{fig:Extended_data_Spatial_Size}a,b). As convolutional neural networks are not built to be scale-invariant, this can make it challenging for the models. Even though spatial-pyramid adaptation we proposed can mitigate it (Figure~\ref{fig:Extended_data_Spatial_Size} c,d,e), early attempts show that bottom-up models give inferior performance compared to top-down models especially for quadruped data. Therefore, we chose top-down for quadruped as it standardized the animal the pose estimator sees, making the pre-training and test-time tasks easier.

\section*{How to use the DeepLabCut Model Zoo}
% the workflow, how to use it
The DeepLabCut Model Zoo consists of two parts. The first is a web-based platform that accepts pose data contributions, ranging from a DeepLabCut project, labeled images from our WebApp, and public animal pose datasets (See Figure~\ref{fig:Extended_data_matching_ATP}d). As these data come in different formats, we implement a software-based data layer dubbed ``generalized data converter" (see Methods) that convert data of various forms to DeepLabCut pose format. We call models we provide Super-Animal models for their generalization powers. After users download these super models from our website or via DeepLabCut APIs, they can either use the models as a plug-and-play solution or alternatively choose to adapt or fine-tune these models from videos or pose datasets. 
\medskip

\section*{Video captions}

\justify \textbf{Suppl\_video1.mp4} Video prediction results by comparison model trained with and without gradient masking.
%gradient-masking.mov

\textbf{Suppl\_video2.mp4} Video prediction results by SuperAnimal model fine-tuned with naive-fine-tuning and SuperAnimal model fine-tuned with memory replay.
%memory-replay.mov

\textbf{Suppl\_video3.mp4} Video prediction results by SuperAnimal-TopViewMouse model with and without spatial pyramid inference. Note that because we use a detector for the SuperAnimal-Quadruped, this is not needed.
%spatial-pyramid.mov

\textbf{Suppl\_video4.mp4} Video prediction results by SuperAnimal models with and without video adaptation.
%video-adaptation.mov

\textbf{Suppl\_video5.mov} Example video from Sturman et al~\cite{sturman2020deep} vs. SuperAnimal-TopViewMouse without any training.

\textbf{Suppl\_video6.mp4} Top-down based SuperAnimal-TopviewMouse's video prediction from one example MABe video, without being trained on any MABe videos.

\clearpage
\newpage
\section*{\huge Supplementary Tables}

\section*{Results Table Summary}

\setcounter{table}{0}

To summarize the results in Figure 1 and Figure 2 on OOD Benchmarks, we provide two Tables:\\

%summary Table for Mouse TS1
\begin{table}[h]
\centering
\caption{\textbf{Main Results on Mouse Benchmarks}. The mAP on multiple architectures, CNN (HRNet, DLCRNet) and Transformer based models (TokenPose models) on SuperAnimal-TopViewMouse. As a reminder, transfer learning means using a randomly initialized decoder that is also trained. Memory replay involves fine-tuning the encoder and decoder.}
\begin{tabular}{l|l|l|l|l|l|l}
\toprule
           Method & Pre-trained Weights &  Data Ratio &     mAP &   RMSE &       Dataset & Architecture \\
\midrule
        zero-shot &    SuperAnimal &   - &  50.397 & 14.32  & DLC\_Openfield &    DLCRNet \\
        zero-shot &    SuperAnimal &   - &  95.219 &  4.881 & DLC\_Openfield &    HRNetw32 \\
        zero-shot &    SuperAnimal &   - &  96.348 &  4.572 & DLC\_Openfield &    AnimalTokenPose \\ 
        \hline 
transfer learning &    ImageNet &   0.01 &  62.226 & 18.136  & DLC\_Openfield &    DLCRNet \\        
transfer learning &       ImageNet &   0.01 &  91.458 &  7.001 & DLC\_Openfield &    HRNetw32 \\
transfer learning &    ImageNet &   1.00 & 99.23  & 2.340   & DLC\_Openfield &    DLCRNet \\
transfer learning &       ImageNet &   1.00 & 100 &  1.131 & DLC\_Openfield &    HRNetw32 \\
    memory replay &    SuperAnimal &   0.01 & 74.225  &  7.688  & DLC\_Openfield &    DLCRNet \\
    memory replay &    SuperAnimal &   0.01 &  99.599 &  2.381 & DLC\_Openfield &    HRNetw32 \\    
    memory replay &    SuperAnimal &   1.00 &  97.946 & 3.071  & DLC\_Openfield &    DLCRNet \\
    memory replay &    SuperAnimal &   1.00 &  99.868 &  1.210 & DLC\_Openfield &    HRNetw32 \\
    \hline 
    \hline 
        zero-shot &    SuperAnimal &   - &  76.139 &  9.013 &      TriMouse &    HRNetw32 \\
        zero-shot &    SuperAnimal &   - &  70.372 &  10.580 &      TriMouse &    AnimalTokenPose \\
        \hline 
transfer learning &       ImageNet &   0.01 &  26.116 & 31.562 &      TriMouse &    HRNetw32 \\
transfer learning &       ImageNet &   1.00 &  97.730 &  2.276 &      TriMouse &    HRNetw32 \\
    memory replay &    SuperAnimal &   0.01 &  90.320 &  5.850 &      TriMouse &    HRNetw32 \\
    memory replay &    SuperAnimal &   1.00 &  98.547 &  2.103 &      TriMouse &    HRNetw32 \\
\bottomrule
\end{tabular}
\label{tab:overall_mouse}
\end{table}

%summary Table for Quad TS2
\begin{table}
\footnotesize
\centering
\caption{\textbf{Main Results on Quadruped Benchmarks}. Here, the base SuperAnimal-Quadruped model had none of the held-out datasets. Full results can be found in Figure~\ref{fig:FigSAQ} for fine-tuning with different amounts of data, but the best fine-tuning performance is shown, which matches the top-performance of the SuperAnimal (SA) variant as shown in Figure~\ref{fig:FigSAQ}. *NOTE: Cao et al.\cite{cao2019cross} do not report a unified single mAP, rather per animal, therefore we trained a model using their dataset to estimate top-line performance if only trained on AP. **Number as reported in~\cite{Xu2022ViTPoseVT} using the data from~\cite{yu2021ap}.}
\begin{tabular}{c|c|c|c|c|c|c|c|c}
\toprule
           Method & Pre-trained Weights &  Data Ratio &    mAP &    RMSE &    Dataset & NE\_IID & NE\_OOD & Architecture \\
\midrule
        zero-shot &    SuperAnimal &   - & 68.038 &  12.971 &      AP-10K &      - &      - &    HRNetw32 \\
        zero-shot &    SuperAnimal &   - & 66.110 &  12.849 &      AP-10K &      - &      - &    AnimalTokenPose \\
        \hline
transfer learning &       ImageNet &   1.00 & 70.548 &  11.228 &      AP-10K &      - &      - &    HRNetw32 \\
    memory replay &    SuperAnimal &   1.00 & 80.113 &  11.296 &      AP-10K &      - &      - &    HRNetw32 \\
    \hline
    \hline
        zero-shot &         AP-10K &   - & 79.447 &   5.774 & AnimalPose &      - &      - &    HRNetw32 \\
        zero-shot &    SuperAnimal &   - & 84.639 &   4.884 & AnimalPose &      - &      - &    HRNetw32 \\
        zero-shot &    SuperAnimal &   - & 83.043 &   5.154 & AnimalPose &      - &      - &    AnimalTokenPose \\
\bottomrule
\end{tabular}
\label{tab:quad}
\end{table}

\newpage
\section*{Extended Full Results on Animal Benchmarks \& Statistical Analysis}

% keypoint dropping TS3
\begin{table}[h]
\centering
\caption{\label{memreplay_lme}Type-III Analysis of Variance Table for the mixed model relative to the quantification of memory replay in terms of keypoint dropping.}
\begin{tabular}{lrrrrrr}
  \hline
 & Sum Sq & Mean Sq & NumDF & DenDF & F value & Pr($>$F) \\ 
  \hline
method & 4096.16 & 4096.16 & 1.00 & 1608.00 & 2636.32 & 0.0000 \\ 
  train data\_ratio & 4839.77 & 1209.94 & 4.00 & 1608.00 & 778.73 & 0.0000 \\ 
  method: train data\_ratio & 5040.49 & 1260.12 & 4.00 & 1608.00 & 811.02 & 0.0000 \\ 
   \hline
\end{tabular}
\end{table}

\begin{table}[ht]
\centering
\caption{\label{contrasts_memreplay}Pairwise contrasts adjusted with Tukey's method for the mixed model relative to the quantification of memory replay in terms of keypoint dropping}
\begin{tabular}{lrrrrlr}
  \hline
contrast & estimate & SE & df & t.ratio & p.value & eff.size \\ 
  \hline
\multicolumn{7}{l}{train data\_ratio = 0.01}\\
fine-tune - (memory replay) & 1.0309 & 0.1385 & 1608 & 7.443 & $<$.0001 & 0.8270 \\ 
   \hline
\multicolumn{7}{l}{train data\_ratio = 0.05}\\
fine-tune - (memory replay) & 0.3025 & 0.1385 & 1608 & 2.184 & 0.0291 & 0.2427 \\ 
   \hline
\multicolumn{7}{l}{train data\_ratio = 0.1}\\
fine-tune - (memory replay) & 0.4444 & 0.1385 & 1608 & 3.209 & 0.0014 & 0.3566 \\ 
   \hline
\multicolumn{7}{l}{train data\_ratio = 0.5}\\
fine-tune - (memory replay) & 4.6420 & 0.1385 & 1608 & 33.516 & $<$.0001 & 3.7240 \\ 
   \hline
\multicolumn{7}{l}{train data\_ratio = 1}\\
fine-tune - (memory replay) & 9.4815 & 0.1385 & 1608 & 68.459 & $<$.0001 & 7.6065 \\ 
   \hline
\multicolumn{7}{l}{{\footnotesize Degrees-of-freedom method: kenward-roger}}\\
\end{tabular}
\end{table}

%TD DLC openfield results TS4
\begin{table}[h]
\centering
\caption{\label{TD_openfield_results} HRNet-w32 TopViewMouse-5k \textbf{DLC Openfield}}
\begin{tabular}{c|c|c|c|c}
\toprule
           method & pretrain\_model & train data\_ratio &     mAP &  RMSE \\
\midrule
      fine-tuning &    SuperAnimal &           0.01 &  98.813 & 2.518 \\
      fine-tuning &    SuperAnimal &           0.05 &  99.802 & 1.706 \\
      fine-tuning &    SuperAnimal &            0.1 &  99.892 & 1.439 \\
      fine-tuning &    SuperAnimal &            0.5 &  99.878 & 1.261 \\
      fine-tuning &    SuperAnimal &            1.0 &  99.925 & 1.234 \\
    memory replay &    SuperAnimal &           0.01 &  99.599 & 2.381 \\
    memory replay &    SuperAnimal &           0.05 &  99.765 & 1.954 \\
    memory replay &    SuperAnimal &            0.1 &  99.929 & 1.538 \\
    memory replay &    SuperAnimal &            0.5 &  99.778 & 1.293 \\
    memory replay &    SuperAnimal &            1.0 &  99.868 & 1.210 \\
transfer learning &       ImageNet &           0.01 &  91.458 & 7.001 \\
transfer learning &       ImageNet &           0.05 &  98.930 & 2.162 \\
transfer learning &       ImageNet &            0.1 &  99.273 & 1.565 \\
transfer learning &       ImageNet &            0.5 &  99.179 & 1.424 \\
transfer learning &       ImageNet &            1.0 & 100.000 & 1.131 \\
transfer learning &    SuperAnimal &           0.01 &  96.612 & 4.400 \\
transfer learning &    SuperAnimal &           0.05 &  99.605 & 1.818 \\
transfer learning &    SuperAnimal &            0.1 &  99.753 & 1.468 \\
transfer learning &    SuperAnimal &            0.5 &  99.252 & 1.463 \\
transfer learning &    SuperAnimal &            1.0 &  99.798 & 1.184 \\
        zero-shot &    SuperAnimal &            1.0 &  95.219 & 4.881 \\
\bottomrule
\end{tabular}
\end{table}

% TD TriMouse results TS5
\begin{table}
\centering
\caption{\label{TD_trimouse_result} HRNet-w32 TopViewMouse-5k \textbf{TriMouse}}
\begin{tabular}{c|c|c|c|c}
\toprule
           method & pretrain\_model & train data\_ratio &    mAP &   RMSE \\
\midrule
      fine-tuning &    SuperAnimal &           0.01 & 88.516 &  9.196 \\
      fine-tuning &    SuperAnimal &           0.05 & 92.695 &  4.314 \\
      fine-tuning &    SuperAnimal &            0.1 & 97.543 &  2.865 \\
      fine-tuning &    SuperAnimal &            0.5 & 98.650 &  2.136 \\
      fine-tuning &    SuperAnimal &            1.0 & 99.021 &  2.020 \\
    memory replay &    SuperAnimal &           0.01 & 90.320 &  5.850 \\
    memory replay &    SuperAnimal &           0.05 & 93.569 &  4.188 \\
    memory replay &    SuperAnimal &            0.1 & 97.744 &  2.864 \\
    memory replay &    SuperAnimal &            0.5 & 98.618 &  2.184 \\
    memory replay &    SuperAnimal &            1.0 & 98.547 &  2.103 \\
transfer learning &       ImageNet &           0.01 & 26.116 & 31.562 \\
transfer learning &       ImageNet &           0.05 & 83.369 &  6.927 \\
transfer learning &       ImageNet &            0.1 & 92.747 &  4.206 \\
transfer learning &       ImageNet &            0.5 & 98.525 &  2.205 \\
transfer learning &       ImageNet &            1.0 & 97.730 &  2.276 \\
transfer learning &    SuperAnimal &           0.01 & 79.292 &  8.740 \\
transfer learning &    SuperAnimal &           0.05 & 89.499 &  4.868 \\
transfer learning &    SuperAnimal &            0.1 & 95.266 &  3.416 \\
transfer learning &    SuperAnimal &            0.5 & 97.838 &  2.246 \\
transfer learning &    SuperAnimal &            1.0 & 98.825 &  2.052 \\
        zero-shot &    SuperAnimal &            1.0 & 76.139 &  9.013 \\
\bottomrule
\end{tabular}
\end{table}

\newpage

% stats TD trimouse TS6
\begin{table}
\centering
\caption{\label{topview_trimouse_lme}Type-III Analysis of Variance Table for the top-down SuperAnimal-TopViewMouse \textbf{TriMouse} benchmark mixed model.}
\begin{tabular}{lrrrrrr}
  \hline
 & Sum Sq & Mean Sq & NumDF & DenDF & F value & Pr($>$F) \\ 
  \hline
method & 510.45 & 127.61 & 4.00 & 48.00 & 23.33 & 0.0000 \\ 
  data\_ratio & 927.34 & 231.83 & 4.00 & 48.00 & 42.39 & 0.0000 \\ 
  method:data\_ratio & 1180.53 & 73.78 & 16.00 & 48.00 & 13.49 & 0.0000 \\ 
   \hline
\end{tabular}
\end{table}

\begin{longtable}{lrrrrlr}
\caption{\label{contrasts_topview_trimouse}Pairwise contrasts adjusted with Tukey's method for the top-down SuperAnimal-TopViewMouse \textbf{TriMouse} benchmark mixed model.
} \\
\hline
contrast & estimate & SE & df & t.ratio & p.value & eff.size \\ 
  \hline
\multicolumn{7}{l}{train data\_ratio = 0.01}\\
(ImageNet transfer learning) - SuperAnimal fine-tune & 22.3653 & 1.9095 & 48 & 11.713 & $<$.0001 & 9.5633 \\ 
  (ImageNet transfer learning) - (SuperAnimal memory replay) & 25.7114 & 1.9095 & 48 & 13.465 & $<$.0001 & 10.9940 \\ 
  (ImageNet transfer learning) - (SuperAnimal transfer learning) & 22.8212 & 1.9095 & 48 & 11.951 & $<$.0001 & 9.7582 \\ 
  (ImageNet transfer learning) - SuperAnimal zero-shot & 22.5486 & 1.9095 & 48 & 11.809 & $<$.0001 & 9.6416 \\ 
  SuperAnimal fine-tune - (SuperAnimal memory replay) & 3.3461 & 1.9095 & 48 & 1.752 & 0.4128 & 1.4308 \\
\end{longtable}

%stats BU dlc openfield TS9
\begin{table}[ht]
\centering
\caption{\label{BU_openfield_lme}Type-III Analysis of Variance Table for bottom-up \textbf{DLC-Openfield} mixed model.}
\begin{tabular}{lrrrrrr}
  \hline
 & Sum Sq & Mean Sq & NumDF & DenDF & F value & Pr($>$F) \\ 
  \hline
method & 845.14 & 211.29 & 4.00 & 48.00 & 46.06 & 0.0000 \\ 
  train data\_ratio & 839.85 & 209.96 & 4.00 & 48.00 & 45.78 & 0.0000 \\ 
  method:train data\_ratio & 407.57 & 25.47 & 16.00 & 48.00 & 5.55 & 0.0000 \\ 
   \hline
\end{tabular}
\end{table}

\begin{table}
\centering
\caption{\label{BU_contrasts_openfield}Pairwise contrasts adjusted with Tukey's method for the bottom-up \textbf{DLC-Openfield} mixed model.
}
\begin{tabular}{lrrrrlr}
  \hline
contrast & estimate & SE & df & t.ratio & p.value & eff.size \\ 
  \hline
\multicolumn{7}{l}{train data\_ratio = 0.01}\\
ImageNet transfer learning - (SuperAnimal memory replay) & 10.4477 & 1.7487 & 48 & 5.975 & $<$.0001 & 4.8783 \\ 
  ImageNet transfer learning - SuperAnimal fine-tune & 6.5661 & 1.7487 & 48 & 3.755 & 0.0041 & 3.0659 \\ 
  ImageNet transfer learning - SuperAnimal transfer learning & 0.7263 & 1.7487 & 48 & 0.415 & 0.9936 & 0.3391 \\ 
  ImageNet transfer learning - (zero-shot) & 3.8184 & 1.7487 & 48 & 2.184 & 0.2034 & 1.7829 \\ 
  (SuperAnimal memory replay) - SuperAnimal fine-tune & -3.8816 & 1.7487 & 48 & -2.220 & 0.1900 & -1.8124 \\ 
  (SuperAnimal memory replay) - SuperAnimal transfer learning & -9.7214 & 1.7487 & 48 & -5.559 & $<$.0001 & -4.5391 \\
\end{tabular}
\end{table}

\newpage

%stats TD dlc openfield TS8
\begin{table}
\centering
\caption{\label{TD_topview_openfield_lme}Type-III Analysis of Variance Table for the top-down SuperAnimal-TopViewMouse DLC-Openfield benchmark mixed model.}
\begin{tabular}{lrrrrrr}
  \hline
 & Sum Sq & Mean Sq & NumDF & DenDF & F value & Pr($>$F) \\ 
  \hline
method & 109.24 & 27.31 & 4.00 & 48.00 & 74.19 & 0.0000 \\ 
  data\_ratio & 54.02 & 13.51 & 4.00 & 48.00 & 36.69 & 0.0000 \\ 
  method:data\_ratio & 45.72 & 2.86 & 16.00 & 48.00 & 7.76 & 0.0000 \\ 
   \hline
\end{tabular}
\end{table}

\begin{longtable}{lrrrrlr}
\caption{\label{contrasts_topview_openfield}Pairwise contrasts adjusted with Tukey's method for the top-down SuperAnimal-TopViewMouse DLC-Openfield benchmark mixed model.
} \\
  \hline
contrast & estimate & SE & df & t.ratio & p.value & eff.size \\ 
  \hline
\multicolumn{7}{l}{train data\_ratio = 0.01}\\
(ImageNet transfer learning) - SuperAnimal fine-tune & 4.4830 & 0.4954 & 48 & 9.050 & $<$.0001 & 7.3890 \\ 
  (ImageNet transfer learning) - (SuperAnimal memory replay) & 4.6197 & 0.4954 & 48 & 9.326 & $<$.0001 & 7.6143 \\ 
  (ImageNet transfer learning) - (SuperAnimal transfer learning) & 2.6006 & 0.4954 & 48 & 5.250 & $<$.0001 & 4.2864 \\ 
  (ImageNet transfer learning) - SuperAnimal zero-shot & 2.1198 & 0.4954 & 48 & 4.279 & 0.0008 & 3.4940 \\ 
  SuperAnimal fine-tune - (SuperAnimal memory replay) & 0.1367 & 0.4954 & 48 & 0.276 & 0.9987 & 0.2254 \\
\end{longtable}

%% QUAD
%Horse-10
\begin{table}
\centering
\caption{\label{horse10_results}HRNet-w32 Quadruped80K \textbf{Horse-10}}
\begin{tabular}{c|c|c|c|c|c|c}
\toprule
           method & pretrain\_model & train data\_ratio &    mAP &  NE\_IID &  NE\_OOD &   RMSE \\
\midrule
      fine-tuning &         AP-10K &           0.01 & 66.284 &   0.286 &   0.285 &  5.029 \\
      fine-tuning &         AP-10K &           0.05 & 80.265 &   0.187 &   0.187 &  2.950 \\
      fine-tuning &         AP-10K &            0.1 & 81.987 &   0.199 &   0.175 &  2.661 \\
      fine-tuning &         AP-10K &            0.5 & 91.369 &   0.070 &   0.101 &  1.557 \\
      fine-tuning &         AP-10K &            1.0 & 93.973 &   0.036 &   0.083 &  1.220 \\
      fine-tuning &    SuperAnimal &           0.01 & 71.684 &   0.219 &   0.213 &  3.855 \\
      fine-tuning &    SuperAnimal &           0.05 & 85.444 &   0.131 &   0.136 &  2.162 \\
\bottomrule
\end{tabular}
\end{table}
%iRodent
\begin{table}
\centering
\caption{\label{rodent_results}HRNet-w32 Quadruped80K \textbf{iRodent}}
\begin{tabular}{c|c|c|c|c}
\toprule
           method & pretrain\_model & train data\_ratio &    mAP &    RMSE \\
\midrule
      fine-tuning &         AP-10K &           0.01 & 43.144 &  37.704 \\
      fine-tuning &         AP-10K &           0.05 & 49.605 &  34.235 \\
      fine-tuning &         AP-10K &            0.1 & 50.019 &  36.970 \\
      fine-tuning &         AP-10K &            0.5 & 57.858 &  29.547 \\
      fine-tuning &         AP-10K &            1.0 & 61.635 &  26.758 \\
      fine-tuning &    SuperAnimal &           0.01 & 59.194 &  32.599 \\
      fine-tuning &    SuperAnimal &           0.05 & 61.255 &  30.897 \\
\bottomrule
\end{tabular}
\end{table}
%AnimalPose
\begin{table}
\centering
\caption{\label{animalpose_results}HRNetw32 Quadruped80K \textbf{AnimalPose}}
\begin{tabular}{c|c|c|c|c}
\toprule
           method & pretrain\_model & train data\_ratio &    mAP &  RMSE \\
\midrule
      fine-tuning &         AP-10K &            1.0 & 86.794 & 4.860 \\
      fine-tuning &    SuperAnimal &            1.0 & 86.851 & 4.706 \\
    memory replay &    SuperAnimal &            1.0 & 87.034 & 4.636 \\
transfer learning &         AP-10K &            1.0 & 89.402 & 5.275 \\
transfer learning &       ImageNet &            1.0 & 86.864 & 5.757 \\
transfer learning &    SuperAnimal &            1.0 & 89.612 & 5.185 \\
        zero-shot &         AP-10K &            1.0 & 79.447 & 5.774 \\
        zero-shot &    SuperAnimal &            1.0 & 84.639 & 4.884 \\
\bottomrule
\end{tabular}
\end{table}
%AP-10K
\begin{table}
\centering
\caption{\label{ap10k_results}HRNet-w32 Quadruped80K \textbf{AP-10K}}
\begin{tabular}{c|c|c|c|c}
\toprule
           method & pretrain\_model & train data\_ratio &    mAP &   RMSE \\
\midrule
      fine-tuning &    SuperAnimal &            1.0 & 79.511 & 11.021 \\
    memory replay &    SuperAnimal &            1.0 & 80.113 & 11.296 \\
transfer learning &       ImageNet &            1.0 & 70.548 & 11.228 \\
transfer learning &    SuperAnimal &            1.0 & 74.379 & 10.748 \\
        zero-shot &    SuperAnimal &            1.0 & 68.038 & 12.971 \\
\bottomrule
\end{tabular}
\end{table}

\newpage 

% horse-10 model stats
\begin{table}
\centering
\caption{\label{horse_ood_lme}Type-III Analysis of Variance Table for Horse-10 OOD mixed model.}
\begin{tabular}{lrrrrrr}
  \hline
 & Sum Sq & Mean Sq & NumDF & DenDF & F value & Pr($>$F) \\ 
  \hline
cond & 2.10 & 0.30 & 7.00 & 78.00 & 183.97 & 0.0000 \\ 
  data\_ratio & 2.88 & 0.72 & 4.00 & 78.00 & 442.97 & 0.0000 \\ 
  cond:data\_ratio & 2.73 & 0.10 & 28.00 & 78.00 & 59.87 & 0.0000 \\ 
   \hline
\end{tabular}
\end{table}

\begin{longtable}{lrrrrlr}
\caption{\label{contrasts_horse_ood}Pairwise contrasts adjusted with Tukey's method for the Horse-10 OOD mixed model.
} \\
  \hline
contrast & estimate & SE & df & t.ratio & p.value & eff.size \\ 
  \hline
\multicolumn{7}{l}{train data\_ratio = 0.01}\\
AP10k fine-tune - (AP10k transfer learning) & 0.6528 & 0.0329 & 78 & 19.819 & $<$.0001 & 16.1825 \\ 
  AP10k fine-tune - AP10k zero-shot & 0.0055 & 0.0329 & 78 & 0.168 & 1.0000 & 0.1375 \\ 
  AP10k fine-tune - (ImageNet transfer learning) & 0.6535 & 0.0329 & 78 & 19.841 & $<$.0001 & 16.2000 \\ 
  AP10k fine-tune - SuperAnimal fine-tune & -0.0540 & 0.0329 & 78 & -1.640 & 0.7249 & -1.3388 \\ 
  AP10k fine-tune - (SuperAnimal memory replay) & -0.0708 & 0.0329 & 78 & -2.150 & 0.3929 & -1.7558 \\
\end{longtable}

% irodent 2
\begin{table}
\centering
\caption{\label{rodent_lme}Type-III Analysis of Variance Table for iRodent benchmark mixed model.}
\begin{tabular}{lrrrrrr}
  \hline
 & Sum Sq & Mean Sq & NumDF & DenDF & F value & Pr($>$F) \\ 
  \hline
cond & 1.44 & 0.21 & 7.00 & 78.00 & 572.88 & 0.0000 \\ 
  data\_ratio & 1.08 & 0.27 & 4.00 & 78.00 & 750.34 & 0.0000 \\ 
  cond:data\_ratio & 0.90 & 0.03 & 28.00 & 78.00 & 89.47 & 0.0000 \\ 
   \hline
\end{tabular}
\end{table}

\begin{longtable}{lrrrrlr}
\caption{\label{contrasts_rodent}Pairwise contrasts adjusted with Tukey's method for the iRodent benchmark mixed model.
} \\
  \hline
contrast & estimate & SE & df & t.ratio & p.value & eff.size \\ 
  \hline
\multicolumn{7}{l}{train data\_ratio = 0.01}\\
AP10k fine-tune - (AP10k transfer learning) & 0.3023 & 0.0155 & 78 & 19.513 & $<$.0001 & 15.9326 \\ 
  AP10k fine-tune - AP10k zero-shot & 0.0275 & 0.0155 & 78 & 1.778 & 0.6365 & 1.4517 \\ 
  AP10k fine-tune - (ImageNet transfer learning) & 0.4236 & 0.0155 & 78 & 27.339 & $<$.0001 & 22.3221 \\ 
  AP10k fine-tune - SuperAnimal fine-tune & -0.1605 & 0.0155 & 78 & -10.359 & $<$.0001 & -8.4579 \\ 
  AP10k fine-tune - (SuperAnimal memory replay) & -0.1771 & 0.0155 & 78 & -11.429 & $<$.0001 & -9.3318 \\
\end{longtable}

%stats for SP
\begin{table}[ht]
\centering
\caption{\label{ks}Spatial Pyramid. Exact two-sample one-sided Kolmogorov-Smirnov test.}
\begin{tabular}{lrr}
  \hline
 & D & \emph{p} \\ 
  \hline
Smear Lab mouse & 1 & <.0001 \\ 
  ood\_ Mathis MausHaus & 0.33 & .27 \\ 
  Golden Lab Mouse & 0.75 & <.0001 \\ 
   \hline
\end{tabular}
\end{table}

% keypoint jittering
\begin{table}[ht]
\centering
\caption{\label{jitter_lme}Type-III Analysis of Variance Table for the mixed model relative to the quantification of video adaptation in terms of keypoint jittering.}
\begin{tabular}{lrrrrrr}
  \hline
 & Sum Sq & Mean Sq & NumDF & DenDF & F value & Pr($>$F) \\ 
  \hline
video & 807.99 & 134.67 & 6.00 & 2.00 & 2.94 & 0.2753 \\ 
  cond & 8699.68 & 8699.68 & 1.00 & 23286.00 & 190.03 & 0.0000 \\ 
  video:cond & 11630.29 & 1938.38 & 6.00 & 23286.00 & 42.34 & 0.0000 \\ 
   \hline
\end{tabular}
\end{table}

% keypoint jittering 2
\begin{table}
\centering
\caption{\label{contrasts_jitter}Pairwise contrasts adjusted with Tukey's method for the mixed model relative to the quantification of video adaptation in terms of keypoint jittering.
}
\begin{tabular}{rlllrrrrrr}
  \hline
contrast & estimate & SE & df & z.ratio & p.value & eff.size \\ 
  \hline
\multicolumn{7}{l}{video = DLC-Openfield}\\
after\_adapt - before\_adapt & -6.3923 & 0.3791 & Inf & -16.860 & $<$.0001 & -0.9447 \\ 
   \hline
\multicolumn{7}{l}{video = Dog}\\
after\_adapt - before\_adapt & -0.1828 & 0.1983 & Inf & -0.922 & 0.3566 & -0.0270 \\ 
   \hline
\multicolumn{7}{l}{video = Elk}\\
after\_adapt - before\_adapt & -1.6065 & 0.1674 & Inf & -9.596 & $<$.0001 & -0.2374 \\ 
   \hline
\multicolumn{7}{l}{video = Golden Lab}\\
after\_adapt - before\_adapt & -0.3877 & 0.7839 & Inf & -0.495 & 0.6209 & -0.0573 \\ 
   \hline
\multicolumn{7}{l}{video = Horse}\\
after\_adapt - before\_adapt & -6.6062 & 0.7974 & Inf & -8.285 & $<$.0001 & -0.9764 \\ 
   \hline
\multicolumn{7}{l}{video = MausHaus}\\
after\_adapt - before\_adapt & -1.5839 & 0.5878 & Inf & -2.695 & 0.0070 & -0.2341 \\ 
   \hline
\multicolumn{7}{l}{video = Smear Lab}\\
after\_adapt - before\_adapt & -1.8673 & 0.1373 & Inf & -13.603 & $<$.0001 & -0.2760 \\ 
   \hline
\multicolumn{7}{l}{{\footnotesize Degrees-of-freedom method: asymptotic}}\\
   \hline
\end{tabular}
\end{table}

% video adaptation 2
\begin{table}
\caption{\label{adapt_gain_aov} One-way repeated measures ANOVA table testing for differences in adaptation gain between smoothing methods.}
\begin{tabular}{llrrrrrrrrrr}
\toprule
 & Source & ddof1 & ddof2 & F & p-unc & p-GG-corr & ng2 & eps & sphericity & W-spher & p-spher \\
\midrule
0 & method & 2 & 58 & 25.078469 & 0.000000 & 0.000018 & 0.363796 & 0.520776 & False & 0.079789 & 0.000000 \\
\bottomrule
\end{tabular}
\end{table}

\begin{table}
\centering
\caption{\label{adapt_gain_contrasts} Post-hoc pairwise contrasts for the adaptation gain ANOVA.}
\begin{tabular}{lllrrlrr}
\toprule
 A & B & T & dof & alternative & p-unc & cohen \\
\midrule
Kalman filter & Self-pacing & -4.725 & 29.000 & two-sided & 0.000 & -1.226 \\
Kalman filter & Video adaptation & -5.319 & 29.000 & two-sided & 0.000 & -1.358 \\
Self-pacing & Video adaptation & -3.261 & 29.000 & two-sided & 0.003 & -0.785 \\
\bottomrule
\end{tabular}
\end{table}

% video adaptation 3
\begin{table}
\centering
\caption{\label{robust_gain_tt} Paired t-test testing for differences in robustness gain between self-pacing and video adaptation.}
\begin{tabular}{lrrlrlr}
\toprule
 & T & dof & alternative & p-val & CI95\% & cohen-d \\
\midrule
T-test & -15.473 & 29 & two-sided & 0.000 & [-4.36 -3.34] & 3.124 \\
\bottomrule
\end{tabular}
\end{table}

% OFT behavior figure 4
\begin{table}
\centering
\caption{\label{contrasts_lme_oft}Type-III Analysis of Variance Table for OFT linear mixed effect model.}
\begin{tabular}{lrrrrrr}
  \hline
 & Sum Sq & Mean Sq & NumDF & DenDF & F value & Pr($>$F) \\ 
  \hline
method & 0.1371 & 0.0457 & 3 & 453 & 0.9988 & 0.3932 \\ 
  action & 12.1056 & 12.1056 & 1 & 453 & 264.6151 & <0.0001 \\ 
  method:action & 0.0202 & 0.0067 & 3 & 453 & 0.1474 & 0.9313 \\ 
   \hline
\end{tabular}
\end{table}

\newpage

\begin{table}
\begin{center}
\caption{\label{mabe_results}\textbf{MABe Results with SA-TVM zero-shot vs. Official MABe pose data.} We show that with SuperAnimal keypoints, we get same performance (independent t-test; \textit{t}=-.02, \textit{p}=.99) in downstream action segmentation as the official pose does in all 13 considered tasks~\cite{sun2023mabe22}, even though our model is never trained on MABe videos. This demonstrates the effectiveness of our models in downstream action segmentation tasks. To qualitatively support our results see Suppl. Video 6.}
\begin{tabular}{lll}
\hline
Task No. & Official MABe pose & SuperAnimal zero-shot \\ \hline
T0    & 0.095018           & 0.095018             \\
T1    & 0.096345           & 0.096350   \\
T2    & 0.657165           & 0.657245    \\
T3    & 0.020959           & 0.020963    \\
T4    & 0.34015            & 0.34020    \\
T5    & 0.718520           & 0.718519             \\
T6    & 0.565967           & 0.565954             \\
T7    & 0.261730           & 0.261697             \\
T8    & 0.005427           & 0.005427             \\
T9    & 0.025384           & 0.025381             \\
T10   & 0.021717           & 0.021703             \\
T11   & 0.107985           & 0.107988    \\
T12   & 0.610986           & 0.610956             \\ \hline
\end{tabular}
\end{center}
\end{table}

\end{document}